\documentclass{article}

\usepackage[final,eandd]{neurips_2026}

\usepackage[utf8]{inputenc}
\usepackage[T1]{fontenc}
\usepackage{hyperref}
\hypersetup{hidelinks}
\usepackage{url}
\usepackage{booktabs}
\usepackage{amsfonts}
\usepackage{nicefrac}
\usepackage{microtype}
\usepackage{xcolor}
\usepackage[accsupp]{axessibility}
\usepackage{graphicx}
\graphicspath{{figs/}}
\usepackage{caption}
\usepackage{makecell}
\usepackage{floatrow}
\newfloatcommand{capbtabbox}{table}[][\FBwidth]
\usepackage{wrapfig}
\usepackage{multirow}
\usepackage{amssymb}
\usepackage{subcaption}
\usepackage{pifont}
\usepackage{longtable}
\usepackage{fvextra}
\usepackage[most]{tcolorbox}
\usepackage{placeins}
\usepackage{pgfplots}
\pgfplotsset{compat=1.18}
\usetikzlibrary{arrows.meta,positioning}
\usepackage[nameinlink,noabbrev]{cleveref}

\crefname{section}{Sec.}{Secs.}
\Crefname{section}{Section}{Sections}
\crefname{table}{Tab.}{Tabs.}
\Crefname{table}{Table}{Tables}
\crefname{figure}{Fig.}{Figs.}
\Crefname{figure}{Figure}{Figures}

\newcommand{\ReefNetTitle}{ReefNet: A Large-Scale Dataset and Benchmark for Fine-Grained Coral Reef Recognition}

\newcommand{\ReefNetAuthors}{%
\normalfont Abdulwahab Felemban$^{1,*}$, Yahia Battach$^{1,*}$, Faizan Farooq Khan$^1$\\
\normalfont Yuqian Fu$^1$, Xuhui Liu$^1$, Yesmeen M.~Khattab$^1$, Yousef A.~Radwan$^1$\\
\normalfont Xiang Li$^1$, Natalie Dunn$^1$, Susanne B\"ahr$^1$, Tullia Isotta Terraneo$^1$\\
\normalfont Fabio Marchese$^1$, Sara Beery$^2$, Luis Silva$^3$, Rhonda Suka$^3$\\
\normalfont Karla Gonzalez Martinez$^1$, Marta Ezeta Watts$^1$, Federica Barreca$^1$\\
\normalfont Nicolas Oury$^1$, Burton H.~Jones$^1$, Francesca Benzoni$^1$\\
\normalfont Mohamed Elhoseiny$^1$\\
\normalfont $^1$King Abdullah University of Science and Technology (KAUST), Saudi Arabia\\
\normalfont $^2$Massachusetts Institute of Technology (MIT), USA\\
\normalfont $^3$Red Sea Global\\
\normalfont $^*$Equal contribution%
}

\title{\ReefNetTitle}
\author{\ReefNetAuthors}

\begin{document}

\maketitle

\begin{abstract}
Coral reef monitoring requires scalable recognition of visually similar taxa across heterogeneous underwater surveys. We introduce ReefNet, a public dataset that aligns approximately 925K hard-coral point annotations from 76 CoralNet sources and an independently collected Al-Wajh dataset with the World Register of Marine Species taxonomy. A stratified expert audit of 8,962 patches guides filtering to a 479,027-annotation benchmark with 39 labels: 38 genera and the family Fungiidae. Expert agreement is 92\% on the retained audit subset, rather than an exhaustive accuracy estimate. We evaluate pretrained models through zero-shot recognition, few-shot adaptation, and full fine-tuning, each on within-source ReefNet and held-out-source Al-Wajh tests. Biological vision--language pretraining provides the strongest zero-shot results; supervised adaptation substantially improves recognition, with DINOv2 reaching 65.79\% macro recall on Al-Wajh. Matched text-description experiments improve all four evaluated MLLMs on Al-Wajh, while a GPT-5.4 reasoning ablation does not improve over direct classification. Long-tail recognition, label uncertainty, and geographic imbalance remain challenges. ReefNet provides a taxonomically grounded resource for studying these challenges; the Al-Wajh test is a regional stress test rather than evidence of global robustness. Public data and resource links are documented in the supplement.
\end{abstract}

\section{Introduction}
\label{sec:intro}

Coral reefs are among the most biodiverse ecosystems on Earth, providing immense ecological and economic value~\cite{barbier2011economic, spalding2017global}. However, these vibrant habitats are increasingly threatened by anthropogenic pressures, including climate change, overfishing, pollution, and ocean acidification~\cite{cooley2022coral, bellwood2004coral, hughes2017coral}. As these stressors intensify, coral cover and overall reef resilience continue to decline, underscoring the urgent need for effective monitoring, conservation, and restoration efforts~\cite{Brandl2019, bellwood2004coral}. A key component of reef conservation is habitat mapping and coverage analysis, which typically relies on expert taxonomists to annotate \textit{in situ} underwater images~\cite{hill2004methods, beijbom2015coralnet}. However, this manual annotation process is labor-intensive and difficult to scale, limiting the spatial and temporal coverage of reef monitoring programs.

\begin{figure}[!t]
    \centering
    \includegraphics[width=\linewidth]{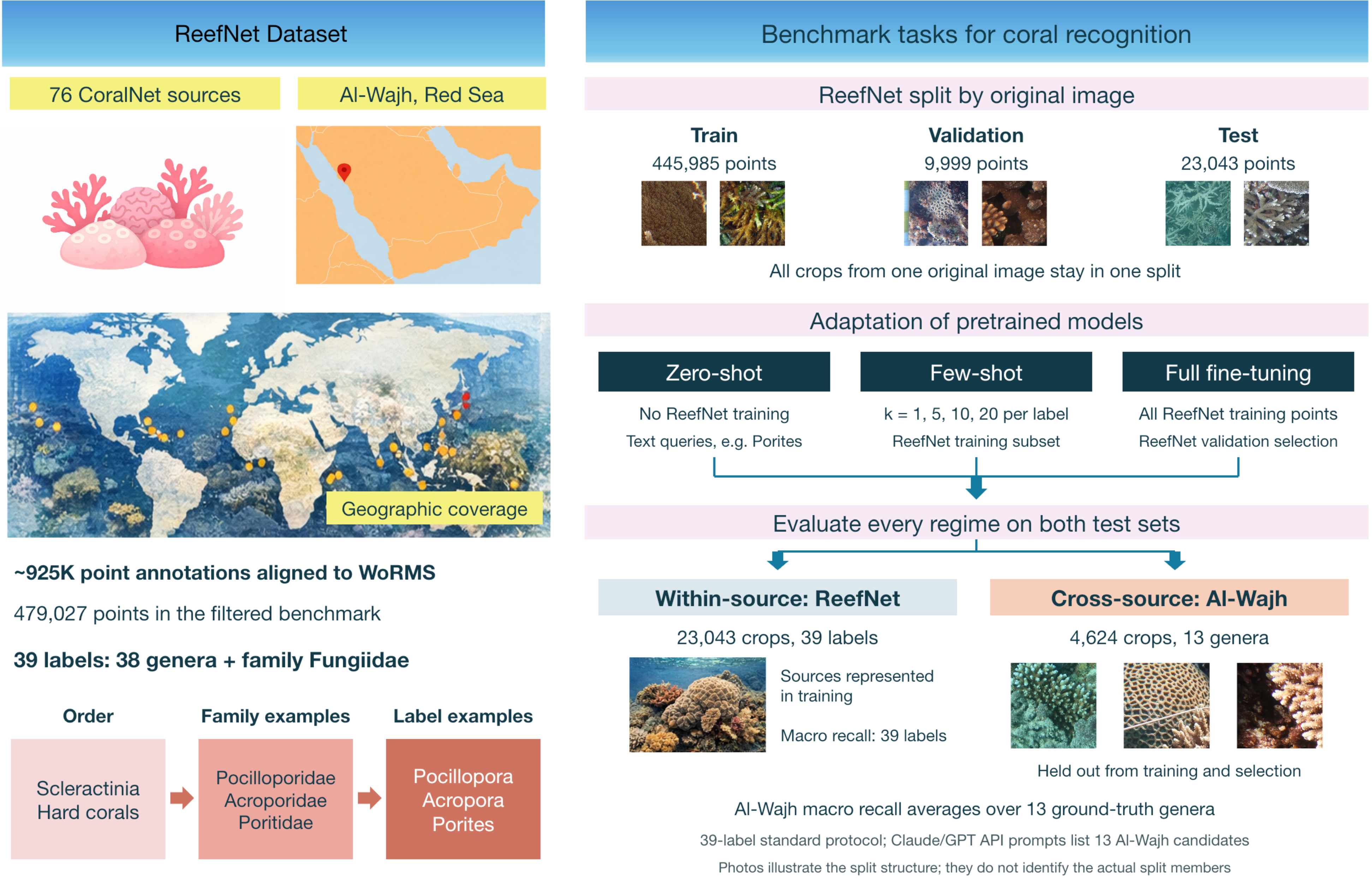}
    \caption{\textbf{ReefNet benchmark overview.} All crops from one original image share a split. Adaptation uses ReefNet training points and validation-based selection; zero-shot models use no ReefNet training examples. Every regime is evaluated on both test sets. Al-Wajh never enters training or selection; its metric averages over 13 present genera. The standard protocol retains 39 output labels; Claude/GPT API prompts instead list 13 Al-Wajh candidates. Photos illustrate the protocol and do not identify actual split membership.}
    \label{fig:teaser}
\end{figure}

Recent advances in deep learning (DL) have accelerated automation across many vision applications~\cite{pan2025locate, ren2023masked, zheng2023deep, zheng2023both, ren2025manifold,scenegraphloc, miao2026langhopslanguagegroundedhierarchical, yu2023point, fu2024cross, fu2025objectrelator}. Yet progress in coral reef recognition remains hindered by two major challenges. First, the availability of large-scale, DL-ready coral datasets with consistent taxonomy and fine-grained annotations is still limited. Second, domain shift across reef sites, caused by variations in imaging conditions, local species assemblages, and annotation conventions, often leads to substantial performance degradation when models are deployed in new locations~\cite{Belcher2023, chen2021coral}. These challenges highlight the need for standardized datasets and evaluation benchmarks that enable systematic investigation of model generalization across reef environments.

Regarding source data, CoralNet~\cite{beijbom2015coralnet, chen2021coral}, one of the most widely used coral annotation platforms, addresses part of this challenge by allowing users to fine-tune classifiers on individual “sources.” However, these source-specific models often perform poorly on unseen reef sites because of differences in taxonomy, imaging protocols, and ecological composition~\cite{williams2019coralnet}. ReefCloud.ai~\cite{reefcloud} further improves label standardization but remains inaccessible to the broader research community. As a result, the lack of large-scale, publicly available datasets with standardized taxonomy and diverse reef coverage continues to limit the development and evaluation of modern machine-learning methods.

To address these limitations, we introduce \textbf{ReefNet}, a large-scale public coral reef image dataset with point-level annotations aligned with the taxonomy of the \emph{World Register of Marine Species} (WoRMS)~\cite{worms}. As illustrated in Figure~\ref{fig:teaser}, ReefNet unifies imagery from 76 carefully curated CoralNet sources and further includes 1.3K newly collected underwater images from Al-Wajh Lagoon in the Red Sea, containing 4,624 expert annotations of hard coral genera from an understudied biogeographic region. In total, ReefNet contains approximately 925K hard-coral point annotations of scleractinian corals (Table~\ref{table1}), making it one of the largest publicly available coral reef datasets for machine-learning research.
To ensure annotation reliability, 8,962 annotations across multiple sources were manually verified by marine biologists, producing a high-confidence repository of coral labels spanning diverse reef ecosystems. The standardized alignment to WoRMS further enables consistent labeling across heterogeneous data sources, facilitating large-scale machine-learning research and cross-dataset comparisons.
Beyond point annotations, our experiments use textual descriptions of the retained coral labels. Appendix~\ref{sup_text_descriptions} reproduces the descriptions and prompts used in the GPT-5.4 experiments and states the available provenance. These descriptions enable language-grounded classification and support emerging research on vision-language models (VLMs) and multimodal large language models (MLLMs) for fine-grained coral reef understanding.

Beyond dataset construction, ReefNet establishes a comprehensive benchmark for fine-grained coral reef recognition. As illustrated in Figure~\ref{fig:teaser}, we evaluate models under multiple realistic settings, including zero-shot recognition, few-shot adaptation of pretrained models, and full fine-tuning, each evaluated within-source and cross-source, enabling systematic analysis of model generalization across reef environments. Extensive experiments reveal that state-of-the-art models struggle with rare classes and morphologically similar coral taxa, highlighting persistent challenges related to adaptation from pretraining, cross-source transfer, and fine-grained visual discrimination.
Through the ReefNet dataset and benchmark protocols, we aim to provide a standardized platform for advancing deep learning research in coral reef recognition. Moreover, the clear performance gains observed under both within-source and cross-source evaluation demonstrate the value of ReefNet not only as a benchmark dataset but also as a large-scale training resource for ecological computer vision, with coral reef understanding serving as a representative real-world application.

The main contributions of this work are as follows. \textbf{ReefNet dataset:} we introduce ReefNet, a large-scale coral reef dataset with WoRMS-aligned standardized point annotations aggregated from 76 CoralNet sources together with newly collected imagery from Al-Wajh Lagoon in the Red Sea, totaling approximately 925K hard-coral point annotations. \textbf{Comprehensive benchmark:} we establish a benchmark for fine-grained coral recognition under zero-shot recognition, few-shot adaptation of pretrained models, and full fine-tuning, each evaluated within-source and cross-source. \textbf{Systematic evaluation and insights:} we conduct extensive experiments with fine-tuned, few-shot, and zero-shot models, revealing persistent challenges in long-tailed coral distributions, held-out-source transfer, and fine-grained visual discrimination.

\section{Related Work}

\paragraph{\textbf{Large-Scale Biodiversity and Marine Datasets.}}
Several large-scale datasets have recently emerged to facilitate computer vision in biodiversity classification. For instance, TreeOfLife-10M~\cite{treeoflife_10m} consolidates over 10 million images of terrestrial and marine organisms from iNat2021~\cite{iNat2021}, BIOSCAN-1M~\cite{BIOSCAN}, and the Encyclopedia of Life (eol.org). However, underwater taxa remain underrepresented in such general repositories due to the logistical challenges of \textit{in situ} marine data acquisition. To bridge this gap in marine ecosystems, BenthicNet~\cite{BenthicNet} aggregates seafloor imagery from multiple surveys, including the Seaview Survey Photo-quadrat~\cite{gonzalez2019seaview}, Reef Life Survey~\cite{Edgar2020RLS}, and others~\cite{SQUIDLE}. BenthicNet constitutes a major step forward, supporting the WoRMS~\cite{worms} taxonomy with 887,533 annotations, of which 287K are hard coral (\emph{Scleractinia}) annotations (Table~\ref{table1}). Nonetheless, it primarily supports broad-scale benthic habitat labeling using the CATAMI classification scheme~\cite{Althaus2015CATAMI}, rather than the fine-grained taxonomic granularity required for specialized coral ecological studies. In contrast, {ReefNet} is explicitly designed to provide the fine-grained taxonomic granularity required for specialized coral ecological studies.

\paragraph{\textbf{Coral-specific Datasets.}}
Existing coral-specific datasets typically face a stark trade-off between taxonomic resolution, geographic diversity, and task specificity. Datasets tailored for dense prediction, such as CoralVOS and CoralSCOP~\cite{coralvos, coralscop}, offer valuable pixel-level masks but categorize corals only at the coarse order level (\emph{Scleractinia}), limiting their utility for fine-grained ecological studies. Similarly, the recent Coralscapes dataset~\cite{sauder2025coralscapesdatasetsemanticscene} extends benthic classification to a sizeable Red Sea dataset with over 170K polygon annotations, but focuses on broader habitat cover types rather than genus- or species-level identification. In contrast, MosaicsUCSD~\cite{edwards2017} and Eilat~\cite{Beijbom2016, Alonso2019} offer highly detailed genus- or species-level annotations but are geographically constrained to a single reef site. As a result, the localized nature heavily restricts the ability of models to generalize across different regions and imaging conditions. To overcome this geographic bias, {ReefNet} aggregates imagery across a diverse, global set of reef sites, enabling the study of spatial generalization while retaining substantial geographic imbalance.

\paragraph{\textbf{Standardizing Coral Reef Recognition.}}
At a broader scale, CoralNet~\cite{beijbom2015coralnet} hosts vast volumes of benthic imagery with both human- and machine-generated labels, providing a rich resource for reef analysis. However, its user-generated labels are heavily source-dependent and lack a unified taxonomic standard like WoRMS, making cross-source learning and systematic evaluation exceedingly difficult. To address this critical inconsistency, {ReefNet} meticulously aligns approximately 925K source annotations, audited through a stratified subset of 8,962 patches, to the universally recognized WoRMS taxonomy, providing a common label vocabulary. Consequently, {ReefNet} provides one of the most comprehensive benchmarks available, supporting a broad spectrum of computer vision tasks for point-based, fine-grained taxon recognition. We provide further clarification of our advantages over CoralNet in the Appendix.

\begin{table*}[!t]
\caption{\textbf{Comparison with existing coral classification datasets.} *Image counts reflect availability as of May 13, 2025. Hard coral/genus-level annotations are not specified for CoralNet because its labels are not standardized. $^{\dagger}$MosaicsUCSD includes 16 annotated orthomosaics, each from $\sim$1.5K images. $^{\ddagger}$CoralSCOP~\cite{coralscop} contains 330,144 binary coral/non-coral masks.}
\label{table1}
\scriptsize
\setlength{\tabcolsep}{2pt}
\renewcommand{\arraystretch}{0.88}
\resizebox{0.94\textwidth}{!}{
\begin{tabular}{lccccccc}
\toprule
 & & & & & & \multicolumn{2}{c}{\textbf{Number of annotations}} \\
\cmidrule{7-8}
\textbf{Dataset} & \makecell{\textbf{Geographic} \\ \textbf{coverage}} & \makecell{\textbf{Annotation} \\ \textbf{type}} & \makecell{\textbf{Lowest} \\ \textbf{taxonomic level}} & \makecell{\textbf{Mapped to} \\ \textbf{WoRMS}} & \makecell{\textbf{Number of} \\ \textbf{images}} & \makecell{\textbf{Hard corals}} & \makecell{\textbf{Genus-level} \\ \textbf{annotations}} \\
\midrule
CoralNet~\cite{beijbom2015coralnet} & \textbf{World} & Sparse points & \textbf{Species} & No & 4,524,792* & N/A & No \\
CoralSCOP~\cite{coralscop} & \textbf{World} & \textbf{Masks} & Order & No & 41,297 & 330,144$^{\ddagger}$ & No \\
CoralVOS~\cite{coralvos}  & 17 sites (South China Sea) & \textbf{Masks} & Order & No & 60,456 & N/A & No \\
MosaicsUCSD~\cite{edwards2017} & Palmyra & \textbf{Masks} & \textbf{Species} & No & 16$^{\dagger}$ & 44,008 & \textbf{Yes} \\
Eilat~\cite{Raphael2020eilat} & Eilat & Sparse points & Genus & No & 212 & $\sim$12,000 & \textbf{Yes} \\
BenthicNet~\cite{BenthicNet} & \textbf{World} & \makecell{Sparse points /\\ image label} & \textbf{Species} & \textbf{Yes} & \textbf{11,408,887} & 287,181 & \textbf{Yes} \\
Coralscapes~\cite{sauder2025coralscapesdatasetsemanticscene} & Red Sea (5 countries) & \textbf{Masks} & N/A & No & 2,075 & N/A & No \\
\textbf{ReefNet (ours)} & \textbf{World} & Sparse points & Genus/family & \textbf{Yes} & 181,223 & \textbf{924,626} & \textbf{Yes} \\
\bottomrule
\end{tabular}
}
\end{table*}

\section{ReefNet Dataset}

\begin{figure}[!t]
    \centering
    \includegraphics[width=0.69\linewidth]{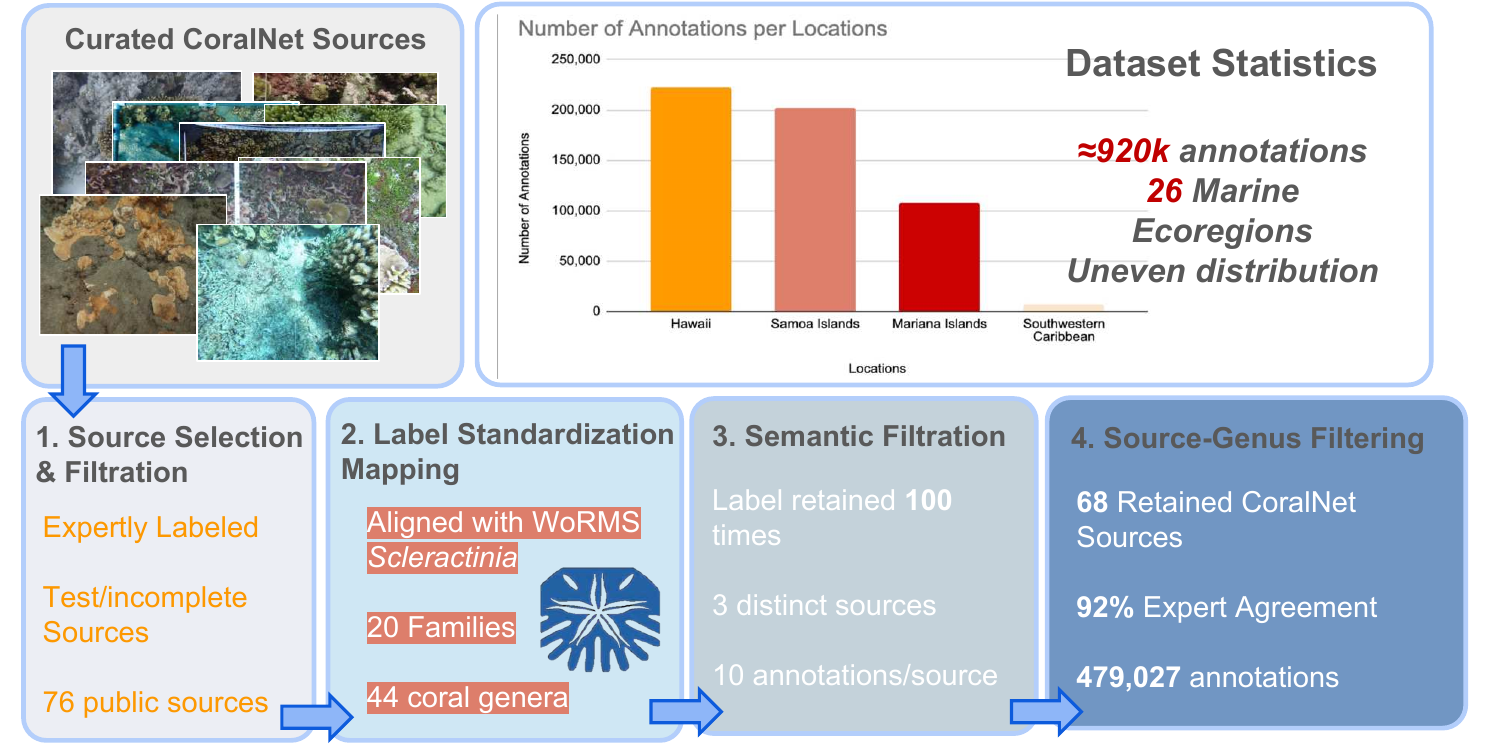}
    \caption{\textbf{ReefNet data construction pipeline.} We apply source filtering, WoRMS taxonomy alignment, semantic label consolidation, and expert verification to construct ReefNet, yielding \(\sim\)925K point annotations across 39 hard-coral label classes together with a new Red Sea test set from Al-Wajh Lagoon.}
    \label{fig_annotation_spread}
\end{figure}

ReefNet is constructed through a multi-stage, expert-guided curation pipeline applied to publicly available benthic imagery and annotations from CoralNet, as shown in Figure~\ref{fig_annotation_spread}. Starting from a large collection of public sources, we perform systematic source filtering, taxonomy alignment, semantic label consolidation, and expert verification to build a high-quality coral reef dataset. This section describes the dataset construction process, including data collection, label standardization, and quality control procedures that ensure the taxonomic reliability and ecological validity of ReefNet.

\subsection{Data Collection}
We started with {1,366} publicly available CoralNet sources and applied a series of semantic, ecological, and technical filters to identify high-quality reef datasets with taxonomically meaningful annotations. This involved removing test or incomplete sources, selecting labels marked as human-confirmed in the source data, and retaining sources with sufficient image and annotation counts. We then applied domain-specific criteria, such as the presence of reef-building corals and shallow, \textit{in situ} imagery, yielding {76} public sources with approximately $\sim$\textbf{920K} genus-level coral annotations. The full source selection pipeline is detailed in the appendix.

\textbf{Al-Wajh Lagoon Data.} We additionally contribute a new dataset from the Al-Wajh Lagoon in the Red Sea (25.6°N, 36.8°E), comprising {1.3K} high-resolution \textit{in situ} images and {4,624} expert annotations of hard corals. This dataset serves as the test set for the cross-source benchmark, enabling evaluation of fine-grained classification and cross-region generalization. Data collection details are provided in the appendix.

\subsection{Taxonomy Mapping and Label Standardization}
To ensure taxonomic traceability and biological consistency, we mapped annotation labels to the {World Register of Marine Species (WoRMS~\cite{worms})}. This step was applied, where relevant, to annotations from the 76 curated CoralNet sources and the Al-Wajh Lagoon dataset. Hard coral labels were manually aligned with canonical scientific names and corresponding {AphiaIDs}\footnote{An AphiaID is a unique identifier from the WoRMS database, linking each label to its corresponding entry in the hierarchical taxonomy. As coral taxonomy evolves, AphiaIDs allow users to verify the current validity of labels.}. This mapping enabled consistent aggregation of biological entities across sources and ensured compatibility with other biodiversity databases. The initial mapping revealed \(\sim\)920K annotations spanning multiple taxonomic levels, which we then consolidated into a unified hard-coral label space. The final retained label set comprises 38 genera plus one family-level class (\textit{Fungiidae}), for a total of 39 hard-coral labels spanning 20 families. However, the initial label space exhibited inconsistencies in naming, taxonomic granularity, and semantic intent, necessitating further filtering and restructuring.

\noindent\textbf{Label Filtering and Semantic Consolidation.} To construct a benchmark dataset suitable for training AI models, we applied a biologically informed filtering strategy targeting syntactic variation (e.g., “Staghorn coral” vs.\ \textit{Acropora cervicornis}) and taxonomic inconsistency (e.g., mixing species-, genus-, and family-level labels). This step was motivated by the inherent difficulty of fine-grained coral identification from imagery alone~\cite{chen2021coral,BenthicNet}, which often results in community datasets with inconsistent or imprecise taxonomic labeling.

Labels were retained only if they referred to hard corals at the genus level, except for \textit{Fungiidae}, which was kept at the family level because several fungiid genera cannot be reliably distinguished in photo-quadrat imagery: fine skeletal details may be unresolved and morphological plasticity blurs genus boundaries. Expert-recommended family-level aggregation preserves an ecologically meaningful group without asserting unsupported visual precision. Retained labels also had to meet the following criteria: appear at least 100 times in the dataset, be present in at least three distinct CoralNet sources with a minimum of 10 annotations per source, exhibit consistent and distinguishable visual patterns, and be taxonomically valid according to WoRMS. After filtering and consolidation, the ReefNet dataset used in our experiments includes 39 unique labels, grouped primarily at the genus level. This curated, taxonomy-aware structure enables hierarchical subsetting, scalable annotation, and custom label aggregation for ecological and machine-learning tasks.

\begin{figure}[!t]
    \centering
    \begin{subfigure}{0.62\linewidth}
        \vspace{0pt}
        \centering
        \includegraphics[width=\linewidth]{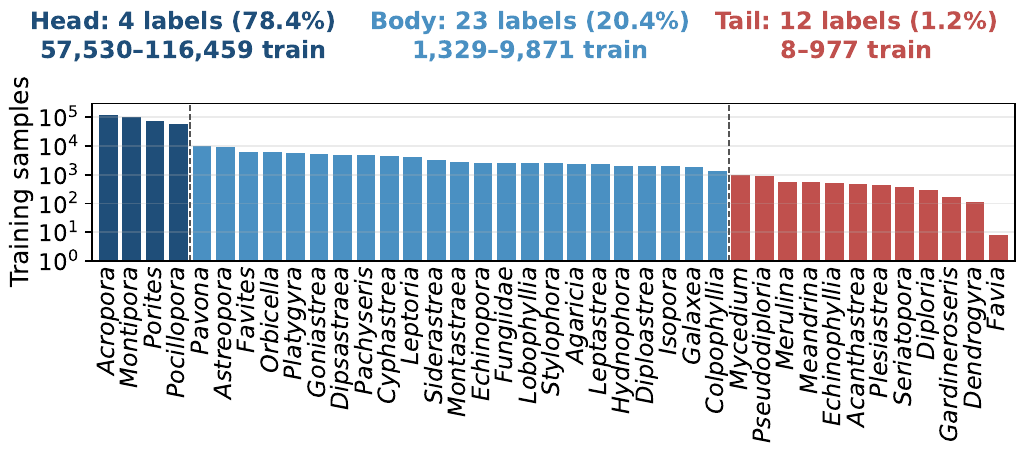}
        \caption{Benchmark training frequencies}
        \label{fig_distribution}
    \end{subfigure}
    \hfill
    \begin{subfigure}{0.34\linewidth}
        \vspace{0pt}
        \centering
        \includegraphics[width=\linewidth]{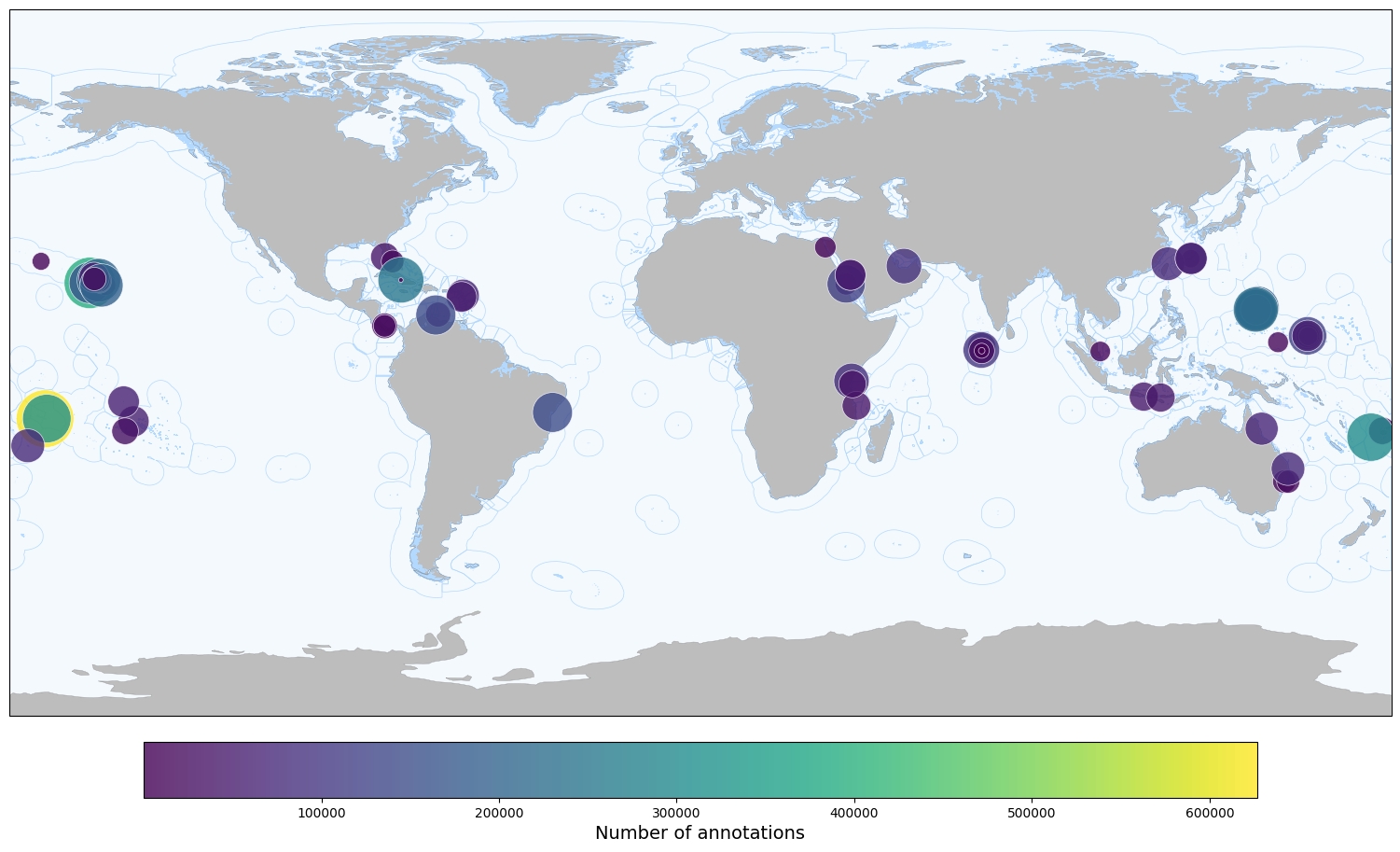}
        \caption{Geographic Distribution}
        \label{geographic_distribution}
    \end{subfigure}

    \caption{
        \textbf{Distributions of Dataset Annotations.}
        {(a) The 445,985 benchmark training samples across 39 labels (38 genera and Fungiidae), on a log scale. Dashed separators mark the fixed head/body/tail groups; annotations give observed training-count ranges.}
        {(b) Full-release geographic distribution of annotations; nearby sources are grouped, and marine ecoregions are shown in light blue~\cite{spalding_ecoregions}.}
    }
    \label{fig:combined_dataset_overview}
\end{figure}

\subsection{Dataset Statistics}

\subsubsection{Annotation Distribution.} \label{sec:annotation_distribution}

 The ReefNet dataset comprises \textbf{334,162} images, of which \textbf{181,223} contain hard coral annotations, totaling \textbf{924,626} point annotations. These annotations span {26 marine ecoregions} across the tropical belt \textit{sensu}~\cite{spalding_ecoregions} (see the appendix for details). The distribution is uneven: the Hawaiian Ecoregion alone contributes over 221K annotations, followed by the Samoa Islands (202K) and the Mariana Islands (108K). In contrast, regions such as the Southwestern Caribbean, Floridian, and Eastern Brazil are sparsely represented, each with fewer than 8K samples. This geographic imbalance reflects broader disparities in coral monitoring efforts and may influence model generalization to underrepresented areas. Among all annotations, 921,351 are assigned to one of 39 labels, capturing a wide range of coral diversity. Figure~\ref{fig_distribution} separately shows the filtered benchmark training split: \textit{Acropora}, \textit{Montipora}, \textit{Porites}, and \textit{Pocillopora} account for 78.4\% of its samples, while 12 labels each have fewer than 1,000 training samples.

\subsubsection{Covariate Diversity.}
In addition to the biological breadth of ReefNet, it exhibits substantial covariate diversity across environmental conditions, camera systems, and imaging protocols (Fig.~\ref{geographic_distribution}), leading to significant variations in resolution, lighting, and water quality across the data. For specific target deployments, it may be advantageous to train only on images from similar environmental conditions or acquisition parameters. We provide detailed metadata on geographic context and camera setups for each CoralNet source in the Appendix, along with the metadata that will be publicly released, thereby enabling the exploration of optimal subsets.

\subsection{Annotation Quality Control}
\label{subsec:annotation-qc}
To assess annotation reliability, we conducted an expert review on a stratified subset of 8,962 hard coral samples. Each review involved evaluating images tied to specific \textit{source–genus} pairs, with labels categorized as \textit{Correct}, \textit{Incorrect}, \textit{Low Quality Image}, or \textit{Hard to Decide}. A micro-averaged expert agreement percentage was calculated as the proportion of \textit{Correct} labels among all reviewed samples.

To support this process, we built a custom web-based application that enabled structured review of annotations across sources. Further details on the interface and the full verification pipeline are provided in the Appendix. Based on this expert feedback, which showed an overall agreement rate of 73\%, we performed targeted quality control, removing low-confidence genera, sources, or genus--source pairs. The quality-control input comprised 920,017 hard-coral annotations. This stage-specific count is distinct from the full-release total of 924,626 annotations and the 921,351 annotations mapped to the 39-label set reported above. These filters improve agreement in the reviewed subset; they do not constitute exhaustive re-verification of the full dataset. Further details on the quality-control procedure are provided in the Appendix.

\noindent\textbf{Source and Genus Filtering.} We first excluded any source where fewer than 50\% of reviewed samples were deemed correct. Similarly, coral genera with less than 50\% expert agreement across all verified sources were removed. This step retained 70 sources, with a post-filter expert agreement of 78\%. After filtering, 860,463 annotations remained across 39 unique hard-coral labels.

\noindent\textbf{Source--Genus Filtering.} We then applied a stricter filter at the source--genus level, retaining only pairs with at least 70\% expert agreement in their verified samples. This resulted in a high-confidence subset with \textbf{92\%} expert agreement on the retained audit subset. After this step, \textbf{479,027} annotations were retained from 68 sources while preserving all 39 label classes. A summary of the filtering process and benchmark splits is presented in Table~\ref{tab:dataset-verification-summary}.

The audit sampled approximately ten patches per source--taxon group rather than sampling in proportion to group size. The 73\%, 78\%, and 92\% agreement rates refer to reviewed patches before filtering, after source/taxon filtering, and after source--taxon filtering, respectively. They are not unbiased accuracy estimates or accuracy floors for the full release. The residual 8\% includes incorrect, low-quality, and hard-to-decide audit cases; it is not a known error list covering 8\% of all benchmark points.

Filtering removes point annotations. Rejected sources lose all their points, rejected taxa lose their points across sources, and rejected source--taxon pairs lose only matching points. The resulting benchmark contains 114,715 contributing images, of which 39,557 (34.5\%) retain only a subset of their original annotations. A further 60,498 images lose all points at this filtering stage and contribute no benchmark samples. These stage-specific image counts are distinct from the full-release image totals above. There is no negative class: an image without a retained hard-coral point supplies no negative example. Appendix~\ref{sup_crop_analysis} quantifies crop composition and spatial overlap.

\begin{table}[!t]
\centering
\scriptsize
\setlength{\tabcolsep}{3pt}
\renewcommand{\arraystretch}{0.9}
\begin{tabular}{lccc|ccc}
\toprule
\textbf{Split}
& \multicolumn{3}{c}{\textbf{ReefNet}}
& \multicolumn{3}{c}{\textbf{Al-Wajh}} \\

\cmidrule(lr){2-4} \cmidrule(lr){5-7}

& \textbf{Annotations} & \textbf{Classes} & \textbf{\# Sources}
& \textbf{Annotations} & \textbf{Classes} & \textbf{\# Sources}\\

\midrule
Train & 445,985 & 39 & 68 & -- & -- & --\\ %
Val   & 9,999   & 38 & 63 & -- & -- & --\\
Test  & 23,043  & 39 & 66 & 4624 & 13 & 1\\
\bottomrule
\end{tabular}
\caption{\textbf{Dataset Split Statistics.} ReefNet provides train/val/test splits covering 39 coral classes (38 in validation). Al-Wajh is used only for cross-source testing and contains 4,624 annotations from 13 classes overlapping with the ReefNet label set.}
\label{tab:dataset-verification-summary}
\end{table}

\section{ReefNet Benchmarks}
\label{sec:reefnet-benchmarks}
\subsection{Dataset Splits and Evaluation Protocol}
\label{subsec:reefnet-protocol}
The benchmark uses the filtered subset from Sec.~\ref{subsec:annotation-qc}. Multilabel Stratified Shuffle Split~\cite{sechidis2011stratification,brady2017iterative} partitions original images within each source into training, validation, and test sets; every point and derived crop from one image stays in the same split. Table~\ref{tab:dataset-verification-summary} gives the counts. The 39 labels comprise 38 genera and Fungiidae; one rare label is absent from validation.

\textbf{Within-source evaluation} uses the 23,043-point ReefNet test split, drawn from sources represented in training. \textbf{Cross-source evaluation} uses 4,624 Al-Wajh points from 13 ground-truth genera. The standard protocol retains all \textbf{39 output labels}; predictions outside the 13 Al-Wajh genera count as errors. The Claude/GPT API runs are an exception: their prompts list 39 candidates on ReefNet but only 13 on Al-Wajh (Appendix~\ref{sup_api_models}). Their Al-Wajh scores therefore use a narrower candidate set. Macro recall averages class recall over the ground-truth labels present in each test (39 and 13, respectively). The differing averaging sets preclude interpreting their absolute difference as a directly comparable generalization gap. Al-Wajh is excluded from training, validation, hyperparameter tuning, and checkpoint selection.

We distinguish three adaptation regimes from these two evaluation settings: pretrained-model zero-shot recognition, few-shot adaptation, and full fine-tuning. Few-shot support sets contain $k\in\{1,5,10,20\}$ examples per label drawn from ReefNet training data and are nested across shot counts. All adaptation uses ReefNet training data, with checkpoint selection on ReefNet validation. Both test sets are reserved for final evaluation.

\section{Experiments}
\label{sec:experiment}
\subsection{Experimental Setup}
We compare embedding vision--language models (VLMs), generative multimodal large language models (MLLMs), and visual classifiers. Macro recall (balanced accuracy)~\cite{scikit-learn} gives equal weight to each ground-truth class despite the imbalance in Fig.~\ref{fig_distribution}. All reported scores are percentages. Model families differ in capacity, pretraining data, and optimization; their rankings are empirical comparisons, not controlled tests of pretraining alone. Visual encoders and VLMs use a 20-epoch protocol; all five open MLLMs use two-epoch LoRA adaptation. These protocols are \emph{not compute-matched}. Appendix~\ref{sec:exp_detail} describes the available training and prompting details.

\begin{table}[t]
\centering\small
\begin{tabular}{lrrrr}
\toprule
& \multicolumn{2}{c}{Name only} & \multicolumn{2}{c}{Name + description}\\
\cmidrule(lr){2-3}\cmidrule(lr){4-5}
Model & ReefNet & Al-Wajh & ReefNet & Al-Wajh\\
\midrule
InternVL-3.5~\cite{intern35} & 4.0 & 9.8 & 3.2 & 10.7\\
LLaVA-NeXT~\cite{llavanext} & 2.6 & 8.3 & -- & --\\
LLaVA-OneVision~\cite{llavaone} & 3.3 & 9.3 & -- & --\\
MiniCPM-V 4.5~\cite{minicpm} & 3.6 & 9.5 & 4.4 & 12.5\\
Qwen3-VL~\cite{qwen3vl} & 5.1 & 15.2 & \underline{5.9} & \underline{16.8}\\
Claude Opus 5$^\dagger$ & \underline{16.75} & 27.82 & -- & --\\
GPT-5.4$^\dagger$ & 14.18 & 26.39 & \textbf{14.92} & \textbf{32.22}\\
\midrule
BioCLIP2~\cite{bioclip2} & \textbf{23.5} & \textbf{43.2} & -- & --\\
BioCLIP~\cite{BioCLIP} & 11.3 & \underline{32.2} & -- & --\\
CLIP~\cite{clip} & 5.2 & 11.2 & -- & --\\
SigLIP~\cite{siglip} & 9.1 & 18.0 & -- & --\\
OpenCLIP~\cite{openclip} & 4.6 & 13.4 & -- & --\\
\bottomrule
\end{tabular}
\caption{\textbf{Zero-shot macro recall (\%).} Best and second-best available scores are bold and underlined \emph{within each column}, separately for each text condition. Dashes indicate unavailable description results. Both text conditions use the same test sets. $\dagger$Claude/GPT API prompts list 39 candidates on ReefNet and 13 on Al-Wajh, unlike the standard 39-label protocol; Al-Wajh comparisons across these candidate sets are not matched. API access: 26--27 July 2026. Appendix~\ref{sup_text_descriptions} states the documentation scope for the open-model description runs.}
\label{tab:zsl_results}
\end{table}

\subsection{Zero-Shot Evaluation of Pretrained Models}
\label{subsec:zsl}
Table~\ref{tab:zsl_results} retains the original name-only condition and adds matched name-plus-description results. BioCLIP2 leads name-only evaluation with 23.5\% macro recall on ReefNet and 43.2\% on Al-Wajh. Claude Opus 5 achieves 16.75\% and 27.82\%, while GPT-5.4 achieves 14.18\% and 26.39\%. These API scores exceed the original open MLLM scores, but the Al-Wajh comparison is not candidate-set matched. The biological pretraining of BioCLIP2 on TreeOfLife-200M~\cite{treeoflife_200m} is associated with strong transfer, but these comparisons do not isolate its causal contribution.

Descriptions improve all four evaluated MLLMs on Al-Wajh and three on ReefNet; InternVL-3.5 declines from 4.0\% to 3.2\% within-source. GPT-5.4 reaches 14.92\% and 32.22\% with descriptions, remaining below BioCLIP2's name-only scores. Descriptions therefore help in some settings without solving fine-grained recognition. Appendix~\ref{sup_text_descriptions} gives the matched comparison and a reasoning ablation; reasoning does not outperform direct classification under either matched Al-Wajh text condition.

\subsection{Few-Shot Adaptation of Pretrained Models}
\label{sec:fs}
\begin{table}[!t]
\centering
\scriptsize
\setlength{\tabcolsep}{2.5pt}
\renewcommand{\arraystretch}{0.9}
\resizebox{0.95\textwidth}{!}{%
\begin{tabular}{lcccc|cccc}
\toprule
\textbf{Model}
& \multicolumn{4}{c}{\textbf{Within-Source Test (ReefNet)}}
& \multicolumn{4}{c}{\textbf{Al-Wajh Test}} \\

\cmidrule(lr){2-5} \cmidrule(lr){6-9}

& \textbf{1-shot} & \textbf{5-shot} & \textbf{10-shot} & \textbf{20-shot}
& \textbf{1-shot} & \textbf{5-shot} & \textbf{10-shot} & \textbf{20-shot} \\

\midrule
Qwen3-VL~\cite{qwen3vl}                     & 5.56 & 9.15 & 17.24 & 27.22 & 10.93 & 13.77 & 25.87 & 31.27 \\ %
MiniCPM-V 4.5~\cite{minicpm}                    & 2.85 & 7.32 & 14.95 & 21.95 &  7.68 &  8.52 & 12.77 & 20.88 \\ %
InternVL-3.5~\cite{intern35}                 & 2.00 & 2.87 & 10.44 & 18.34 & 10.31 &  9.01 & 12.80 & 12.87 \\ %
\midrule
CLIP~\cite{clip}            & 8.10 & \underline{23.02} & 38.41 & 45.71 & \underline{23.16} & \underline{29.38} & \underline{47.64} & \underline{57.30} \\
OpenCLIP~\cite{openclip}    & \underline{9.70} & 21.45 & \underline{39.61} & \underline{45.95} & 18.40 & 23.27 & 45.84 & 46.51 \\
BioCLIP2~\cite{bioclip2}                     & \textbf{30.54} & \textbf{35.59} & \textbf{44.68} & \textbf{58.79} & \textbf{42.56} & \textbf{38.68} & \textbf{60.02} & \textbf{67.83} \\
\bottomrule
\end{tabular}
}

\caption{\textbf{Few-Shot Hard Coral Classification Macro Recall (\%).} Single-run models are fine-tuned with $k$ examples per label from ReefNet. Bold and underline denote the best and second-best result in each column. Results are reported for the within-source ReefNet test set and the cross-source Al-Wajh test set.}
\label{tab:few_shot}
\end{table}

We adapt the three strongest original open zero-shot MLLMs and three representative VLMs using the nested support sets. Table~\ref{tab:few_shot} shows that BioCLIP2 leads at every shot count on both tests. At 20 shots, it reaches 58.79\% on ReefNet and 67.83\% on Al-Wajh, compared with Qwen3-VL's 27.22\% and 31.27\%. Standard CLIP also adapts strongly, reaching 45.71\% and 57.30\%. Gains are not uniformly monotonic: BioCLIP2 drops from 42.56\% to 38.68\% on Al-Wajh between one and five shots. These single-support-set results do not quantify support-sampling variability. LLaVA-NeXT, LLaVA-OneVision, hosted MLLMs, and the added visual backbones are not included in this few-shot comparison.

\subsection{Full Fine-Tuning Evaluation}
\label{subsec:ft}
\begin{table}[!t]
\centering
\small
\setlength{\tabcolsep}{3pt}
\renewcommand{\arraystretch}{1.0}
\begin{tabular}{llcc}
\toprule
\textbf{Category} & \textbf{Model}
& \textbf{ReefNet Test}
& \textbf{Al-Wajh Test} \\
\midrule

\multirow{11}{*}{Vision}
& DINOv2 ViT-L/14~\cite{dinov2} & \underline{79.27} & \textbf{65.79} \\
& DINOv3 ViT-L/16~\cite{dinov3} & 77.42 & 59.76 \\
& SigLIP 2 ViT-L/16 (visual)~\cite{siglip2} & 74.21 & 56.71 \\
& Public MAE ViT-L/16~\cite{he2022masked} & 61.24 & 49.74 \\
& ViT-B/16~\cite{vit}        & \textbf{79.84} & 54.84 \\
& EfficientNet-B3~\cite{tan2019efficientnet} & 71.29 & 52.58 \\
& DeiT-B/16~\cite{deit}       & 76.22 & 51.52 \\
& ConvNeXt-B~\cite{liu2022convnet}      & 74.70 & 50.47 \\
& Swin-B~\cite{swin}          & 77.12 & 50.21 \\
& BEiT-B/16~\cite{bao2021beit}       & 66.66 & 48.23 \\
& ResNet-50~\cite{he2016deep}       & 45.55 & 36.93 \\

\midrule
\multirow{5}{*}{MLLMs (LoRA)}
& Qwen3-VL~\cite{qwen3vl}     & 72.99 & \underline{59.80} \\
& MiniCPM-V 4.5~\cite{minicpm}    & 63.36 & 48.09 \\
& LLaVA-NeXT~\cite{llavanext} & 61.70 & 38.62 \\
& LLaVA-OneVision~\cite{llavaone} & 60.50 & 35.39 \\
& InternVL-3.5~\cite{intern35} & 53.44 & 32.97 \\

\midrule
\multirow{3}{*}{VLMs}
& CLIP~\cite{clip}           & 76.34 & 58.51 \\
& OpenCLIP~\cite{openclip}   & 76.78 & 57.48 \\
& BioCLIP2~\cite{bioclip2}                    & 79.23 & 58.28 \\

\bottomrule
\end{tabular}

\caption{\textbf{Full-data fine-tuning macro recall (\%).} Scores are single-run point estimates; best and second-best across all listed models in each column are bold and underlined. Vision/VLM training uses 20 epochs; MLLM LoRA uses two, without compute matching. ViT and BioCLIP2 seed variability is reported separately in Appendix~\ref{sup_training_analysis}.}
\label{tab:full_ft}
\end{table}

Table~\ref{tab:full_ft} expands the visual classifiers with DINOv2 ViT-L/14~\cite{dinov2}, DINOv3 ViT-L/16~\cite{dinov3}, the SigLIP 2 ViT-L/16 visual encoder~\cite{siglip2}, and public MAE ViT-L/16~\cite{he2022masked}. These use pretrained visual representations with a 39-label classification head; the SigLIP 2 classifier is distinct from zero-shot SigLIP image--text matching, and the MAE result uses public pretraining rather than ReefNet-specific pretraining.

ViT-B/16 has the highest ReefNet single-run score (79.84\%), followed by DINOv2 (79.27\%) and BioCLIP2 (79.23\%). DINOv2 leads on Al-Wajh (65.79\%), followed by Qwen3-VL (59.80\%) and DINOv3 (59.76\%). The MLLM comparison now includes LLaVA-NeXT and LLaVA-OneVision under the same two-epoch LoRA protocol; Qwen3-VL remains strongest among all five open MLLMs. Small differences between single runs should not be interpreted as statistically established model rankings.

\subsection{Additional Analysis}
The training distribution is strongly skewed: the head group contains four labels and 78.4\% of training samples, the body 23 labels and 20.4\%, and the tail 12 labels and 1.2\%. Existing per-class results in Appendix~\ref{sec:more_results} show substantial remaining rare-class difficulty. Frequency groups and per-class sample counts must be interpreted with their stated populations.

Additional multi-seed results for selected models are provided in Appendix~\ref{sup_training_analysis}. A single augmentation run reaches 80.90\% and 55.47\%; this is limited evidence from one architecture, not a significance test or a universal assessment of augmentation. Appendix~\ref{sup_training_analysis} also reports the augmentation policy and available convergence evidence. Predictive-uncertainty validation remains open.

\section{Conclusion}
ReefNet harmonizes heterogeneous coral annotations into 39 biologically motivated labels and provides image-grouped splits for zero-shot, few-shot, and full-data adaptation. A stratified expert audit guides filtering without implying exhaustive verification. Biological VLMs lead zero-shot recognition, supervised adaptation yields substantial gains, and DINOv2 is strongest on the held-out Al-Wajh test among the reported fine-tuned models. Text descriptions improve several MLLM conditions, while the reasoning ablation does not improve the strongest matched direct-classification result. Rare taxa, residual label uncertainty, and geographic imbalance remain important limitations. Al-Wajh provides an independent regional stress test; broader region-disjoint evaluations are needed to assess global robustness.

\begin{ack}
This research was funded by the Ocean Science and Solutions Applied Research Institute (OSSARI), NEOM Company under the ``Towards AI Accelerated Reef Ecology Practice: Expert-informed Benchmarks and Machine Learning Systems for Coral Reef 2D /3D Visual Understanding'' project grant (RGC/3/6029-01) awarded to M. Elhoseiny, B. Jones, and F. Benzoni.

We thank Brian Owain Nieuwenhuis and the CoralNet team. We are grateful to the Red Sea Global team for providing the Al-Wajh dataset and to the Habb team members for their contributions to manual data verification. We thank the KAUST Supercomputing Core Lab (KSL) team for providing computational infrastructure and technical support for this research. We also acknowledge the original data contributors whose publicly shared CoralNet imagery and annotations made ReefNet possible; detailed attribution is provided in the supplementary material.
\end{ack}

\clearpage
\bibliographystyle{plainnat}
\bibliography{main}

\begin{thebibliography}{72}
\providecommand{\natexlab}[1]{#1}
\providecommand{\url}[1]{\texttt{#1}}
\expandafter\ifx\csname urlstyle\endcsname\relax
  \providecommand{\doi}[1]{doi: #1}\else
  \providecommand{\doi}{doi: \begingroup \urlstyle{rm}\Url}\fi

\bibitem[AIMS(2024)]{reefcloud}
AIMS.
\newblock Reefcloud, 2024.
\newblock URL \url{https://reefcloud.ai}.

\bibitem[Alonso et~al.(2019)Alonso, Yuval, Eyal, Treibitz, and
  Murillo]{Alonso2019}
Iñigo Alonso, Matan Yuval, Gal Eyal, Tali Treibitz, and Ana~C. Murillo.
\newblock Coralseg: Learning coral segmentation from sparse annotations.
\newblock \emph{Journal of Field Robotics}, 36:\penalty0 1456--1477, 12 2019.
\newblock ISSN 1556-4959.
\newblock \doi{10.1002/rob.21915}.
\newblock URL \url{https://onlinelibrary.wiley.com/doi/10.1002/rob.21915}.

\bibitem[Althaus et~al.(2015)Althaus, Hill, Ferrari, Edwards, Przeslawski,
  Schönberg, Stuart-Smith, Barrett, Edgar, Colquhoun, Tran, Jordan, Rees, and
  Gowlett-Holmes]{Althaus2015CATAMI}
Franziska Althaus, Nicole Hill, Renata Ferrari, Luke Edwards, Rachel
  Przeslawski, Christine H.~L. Schönberg, Rick Stuart-Smith, Neville Barrett,
  Graham Edgar, Jamie Colquhoun, Maggie Tran, Alan Jordan, Tony Rees, and Karen
  Gowlett-Holmes.
\newblock A standardised vocabulary for identifying benthic biota and substrata
  from underwater imagery: The catami classification scheme.
\newblock \emph{PLOS ONE}, 10\penalty0 (10):\penalty0 e0141039, 2015.
\newblock \doi{10.1371/journal.pone.0141039}.
\newblock URL \url{https://doi.org/10.1371/journal.pone.0141039}.

\bibitem[Bai et~al.(2025)Bai, Cai, Chen, Chen, Chen, Cheng, Deng, Ding, Gao,
  Ge, Ge, Guo, Huang, Huang, Huang, Hui, Jiang, Li, Li, Li, Li, Lin, Lin, Liu,
  Liu, Liu, Liu, Liu, Liu, Lu, Luo, Lv, Men, Meng, Ren, Ren, Song, Sun, Tang,
  Tu, Wan, Wang, Wang, Wang, Wang, Xie, Xu, Xu, Xu, Yang, Yang, Yang, Yang, Yu,
  Zhang, Zhang, Zhang, Zheng, Zhong, Zhou, Zhou, Zhou, Zhu, and Zhu]{qwen3vl}
Shuai Bai, Yuxuan Cai, Ruizhe Chen, Keqin Chen, Xionghui Chen, Zesen Cheng,
  Lianghao Deng, Wei Ding, Chang Gao, Chunjiang Ge, Wenbin Ge, Zhifang Guo,
  Qidong Huang, Jie Huang, Fei Huang, Binyuan Hui, Shutong Jiang, Zhaohai Li,
  Mingsheng Li, Mei Li, Kaixin Li, Zicheng Lin, Junyang Lin, Xuejing Liu,
  Jiawei Liu, Chenglong Liu, Yang Liu, Dayiheng Liu, Shixuan Liu, Dunjie Lu,
  Ruilin Luo, Chenxu Lv, Rui Men, Lingchen Meng, Xuancheng Ren, Xingzhang Ren,
  Sibo Song, Yuchong Sun, Jun Tang, Jianhong Tu, Jianqiang Wan, Peng Wang,
  Pengfei Wang, Qiuyue Wang, Yuxuan Wang, Tianbao Xie, Yiheng Xu, Haiyang Xu,
  Jin Xu, Zhibo Yang, Mingkun Yang, Jianxin Yang, An~Yang, Bowen Yu, Fei Zhang,
  Hang Zhang, Xi~Zhang, Bo~Zheng, Humen Zhong, Jingren Zhou, Fan Zhou, Jing
  Zhou, Yuanzhi Zhu, and Ke~Zhu.
\newblock Qwen3-vl technical report, 2025.
\newblock URL \url{https://arxiv.org/abs/2511.21631}.

\bibitem[Bao et~al.(2022)Bao, Dong, Piao, and Wei]{bao2021beit}
Hangbo Bao, Li~Dong, Songhao Piao, and Furu Wei.
\newblock {BE}it: {BERT} pre-training of image transformers.
\newblock In \emph{International Conference on Learning Representations}, 2022.
\newblock \doi{10.48550/arXiv.2106.08254}.
\newblock URL \url{https://openreview.net/forum?id=p-BhZSz59o4}.

\bibitem[Barbier et~al.(2011)Barbier, Hacker, Kennedy, Koch, Stier, and
  Silliman]{barbier2011economic}
Edward~B. Barbier, Sally~D. Hacker, Chris Kennedy, Evamaria~W. Koch, Adrian~C.
  Stier, and Brian~R. Silliman.
\newblock The value of estuarine and coastal ecosystem services.
\newblock \emph{Ecological Monographs}, 81:\penalty0 169--193, 5 2011.
\newblock ISSN 00129615.
\newblock \doi{10.1890/10-1510.1}.

\bibitem[Beijbom et~al.(2015)Beijbom, Edmunds, Roelfsema, Smith, Kline, Neal,
  Dunlap, Moriarty, Fan, Tan, Chan, Treibitz, Gamst, Mitchell, and
  Kriegman]{beijbom2015coralnet}
Oscar Beijbom, Peter~J. Edmunds, Chris Roelfsema, Jennifer Smith, David~I.
  Kline, Benjamin~P. Neal, Matthew~J. Dunlap, Vincent Moriarty, Tung-Yung Fan,
  Chih-Jui Tan, Stephen Chan, Tali Treibitz, Anthony Gamst, B.~Greg Mitchell,
  and David Kriegman.
\newblock Towards automated annotation of benthic survey images: Variability of
  human experts and operational modes of automation.
\newblock \emph{PLOS ONE}, 10:\penalty0 e0130312, 7 2015.
\newblock ISSN 1932-6203.
\newblock \doi{10.1371/journal.pone.0130312}.
\newblock URL \url{https://dx.plos.org/10.1371/journal.pone.0130312}.

\bibitem[Beijbom et~al.(2016)Beijbom, Treibitz, Kline, Eyal, Khen, Neal, Loya,
  Mitchell, and Kriegman]{Beijbom2016}
Oscar Beijbom, Tali Treibitz, David~I. Kline, Gal Eyal, Adi Khen, Benjamin
  Neal, Yossi Loya, B.~Greg Mitchell, and David Kriegman.
\newblock Improving automated annotation of benthic survey images using
  wide-band fluorescence.
\newblock \emph{Scientific Reports}, 6:\penalty0 23166, 3 2016.
\newblock ISSN 2045-2322.
\newblock \doi{10.1038/srep23166}.
\newblock URL \url{https://www.nature.com/articles/srep23166}.

\bibitem[Belcher et~al.(2023)Belcher, Bower, Burford, Celis, Fahimipour,
  Guevara, Katija, Khokhar, Manjunath, Nelson, et~al.]{Belcher2023}
Byron~T Belcher, Eliana~H Bower, Benjamin Burford, Maria~Rosa Celis, Ashkaan~K
  Fahimipour, Isabela~L Guevara, Kakani Katija, Zulekha Khokhar, Anjana
  Manjunath, Samuel Nelson, et~al.
\newblock Demystifying image-based machine learning: a practical guide to
  automated analysis of field imagery using modern machine learning tools.
\newblock \emph{Frontiers in Marine Science}, 10, 2023.
\newblock \doi{10.3389/fmars.2023.1157370}.

\bibitem[Bellwood et~al.(2004)Bellwood, Hughes, Folke, and
  Nyström]{bellwood2004coral}
D.~R. Bellwood, T.~P. Hughes, C.~Folke, and M.~Nyström.
\newblock Confronting the coral reef crisis.
\newblock \emph{Nature}, 429\penalty0 (6994):\penalty0 827--833, 2004.
\newblock \doi{10.1038/nature02691}.
\newblock URL \url{https://doi.org/10.1038/nature02691}.

\bibitem[Board(2024)]{worms}
WoRMS~Editorial Board.
\newblock Worms: World register of marine species, 2024.
\newblock URL \url{https://www.marinespecies.org}.

\bibitem[Brady(2017)]{brady2017iterative}
Trent Brady.
\newblock Iterative stratification.
\newblock \url{https://github.com/trent-b/iterative-stratification}, 2017.
\newblock Accessed: 2025-05-16.

\bibitem[Brandl et~al.(2019)Brandl, Rasher, Côté, Casey, Darling, Lefcheck,
  and Duffy]{Brandl2019}
Simon~J Brandl, Douglas~B Rasher, Isabelle~M Côté, Jordan~M Casey, Emily~S
  Darling, Jonathan~S Lefcheck, and J~Emmett Duffy.
\newblock Coral reef ecosystem functioning: eight core processes and the role
  of biodiversity.
\newblock \emph{Frontiers in Ecology and the Environment}, 17:\penalty0
  445--454, 10 2019.
\newblock ISSN 1540-9295.
\newblock \doi{10.1002/fee.2088}.
\newblock URL \url{https://onlinelibrary.wiley.com/doi/10.1002/fee.2088}.

\bibitem[Chen et~al.(2021)Chen, Beijbom, Chan, Bouwmeester, and
  Kriegman]{chen2021coral}
Qimin Chen, Oscar Beijbom, Stephen Chan, Jessica Bouwmeester, and David
  Kriegman.
\newblock A new deep learning engine for coralnet.
\newblock In \emph{2021 IEEE/CVF International Conference on Computer Vision
  Workshops (ICCVW)}, volume 2021-October, pages 3686--3695. IEEE, 10 2021.
\newblock ISBN 978-1-6654-0191-3.
\newblock \doi{10.1109/ICCVW54120.2021.00412}.
\newblock URL \url{https://ieeexplore.ieee.org/document/9607450/}.

\bibitem[Cooley et~al.(2023)Cooley, Schoeman, Bopp, Boyd, Donner, Ghebrehiwet,
  Ito, Kiessling, Martinetto, Ojea, Racault, Rost, and
  Skern-Mauritzen]{cooley2022coral}
S~Cooley, D~Schoeman, L~Bopp, P~Boyd, S~Donner, D~Y Ghebrehiwet, S.-I Ito,
  W~Kiessling, P~Martinetto, E~Ojea, M.-F Racault, B~Rost, and
  M~Skern-Mauritzen.
\newblock Oceans and coastal ecosystems and their services.
\newblock In \emph{Climate Change 2022 – Impacts, Adaptation and
  Vulnerability}, pages 379--550. Cambridge University Press, Cambridge and New
  York, 6 2023.
\newblock \doi{10.1017/9781009325844.005}.
\newblock URL
  \url{https://www.cambridge.org/core/product/identifier/9781009325844%23c3/type/book_part}.

\bibitem[Dosovitskiy et~al.(2021)Dosovitskiy, Beyer, Kolesnikov, Weissenborn,
  Zhai, Unterthiner, Dehghani, Minderer, Heigold, Gelly, Uszkoreit, and
  Houlsby]{vit}
Alexey Dosovitskiy, Lucas Beyer, Alexander Kolesnikov, Dirk Weissenborn,
  Xiaohua Zhai, Thomas Unterthiner, Mostafa Dehghani, Matthias Minderer, Georg
  Heigold, Sylvain Gelly, Jakob Uszkoreit, and Neil Houlsby.
\newblock An image is worth 16x16 words: Transformers for image recognition at
  scale.
\newblock In \emph{International Conference on Learning Representations}, 2021.
\newblock \doi{10.48550/arXiv.2010.11929}.
\newblock URL \url{https://openreview.net/forum?id=YicbFdNTTy}.

\bibitem[Edgar et~al.(2020)Edgar, Cooper, Baker, Barker, Barrett, Becerro,
  Bates, Brock, Ceccarelli, Clausius, Davey, Davis, Day, Green, Griffiths,
  Hicks, Hinojosa, Jones, Kininmonth, Larkin, Lazzari, Lefcheck, Ling, Mooney,
  Oh, Pérez-Matus, Pocklington, Riera, Sanabria-Fernandez, Seroussi, Shaw,
  Shields, Shields, Smith, Soler, Stuart-Smith, Turnbull, and
  Stuart-Smith]{Edgar2020RLS}
Graham~J. Edgar, Antonia Cooper, Susan~C. Baker, William Barker, Neville~S.
  Barrett, Mikel~A. Becerro, Amanda~E. Bates, Danny Brock, Daniela~M.
  Ceccarelli, Ella Clausius, Marlene Davey, Tom~R. Davis, Paul~B. Day, Andrew
  Green, Samuel~R. Griffiths, Jamie Hicks, Iván~A. Hinojosa, Ben~K. Jones,
  Stuart Kininmonth, Meryl~F. Larkin, Natali Lazzari, Jonathan~S. Lefcheck,
  Scott~D. Ling, Peter Mooney, Elizabeth Oh, Alejandro Pérez-Matus,
  Jacqueline~B. Pocklington, Rodrigo Riera, Jose~A. Sanabria-Fernandez, Yanir
  Seroussi, Ian Shaw, Derek Shields, Joe Shields, Margo Smith, German~A. Soler,
  Jemina Stuart-Smith, John Turnbull, and Rick~D. Stuart-Smith.
\newblock Establishing the ecological basis for conservation of shallow marine
  life using reef life survey.
\newblock \emph{Biological Conservation}, 252:\penalty0 108855, 12 2020.
\newblock ISSN 00063207.
\newblock \doi{10.1016/j.biocon.2020.108855}.
\newblock URL
  \url{https://linkinghub.elsevier.com/retrieve/pii/S0006320720309137}.

\bibitem[Edwards et~al.(2017)Edwards, Eynaud, Williams, Pedersen, Zgliczynski,
  Gleason, Smith, and Sandin]{edwards2017}
Clinton~B Edwards, Yoan Eynaud, Gareth~J Williams, Nicole~E Pedersen, Brian~J
  Zgliczynski, Arthur~CR Gleason, Jennifer~E Smith, and Stuart~A Sandin.
\newblock Large-area imaging reveals biologically driven non-random spatial
  patterns of corals at a remote reef.
\newblock \emph{Coral Reefs}, 36:\penalty0 1291--1305, 12 2017.
\newblock ISSN 0722-4028.
\newblock \doi{10.1007/s00338-017-1624-3}.
\newblock URL \url{http://link.springer.com/10.1007/s00338-017-1624-3}.

\bibitem[Ehrenberg et~al.(2022)Ehrenberg, Winston, Oliver, and
  Couch]{Ehrenberg2022}
Jon Ehrenberg, Morgan Winston, Thomas Oliver, and Courtney Couch.
\newblock Development of a semi-automated coral bleaching classifier in
  coralnet: A summary of standard operating procedures and report of results.
\newblock NOAA Technical Memorandum NMFS PIFSC 133, NOAA Pacific Islands
  Fisheries Science Center, 2022.
\newblock URL \url{https://repository.library.noaa.gov/view/noaa/47285}.

\bibitem[Friedman(2020)]{SQUIDLE}
Ariell Friedman.
\newblock Squidle+, a tool for managing, exploring \& annotating images, video
  \& large-scale mosaics, 2020.
\newblock URL \url{https://squidle.org/}.

\bibitem[Fu et~al.(2024)Fu, Wang, Pan, Huai, Qiu, Shangguan, Liu, Fu, Van~Gool,
  and Jiang]{fu2024cross}
Yuqian Fu, Yu~Wang, Yixuan Pan, Lian Huai, Xingyu Qiu, Zeyu Shangguan, Tong
  Liu, Yanwei Fu, Luc Van~Gool, and Xingqun Jiang.
\newblock Cross-domain few-shot object detection via enhanced open-set object
  detector.
\newblock In \emph{ECCV}, 2024.

\bibitem[Fu et~al.(2025)Fu, Wang, Ren, Sun, Gong, Fu, Paudel, Huang, and
  Van~Gool]{fu2025objectrelator}
Yuqian Fu, Runze Wang, Bin Ren, Guolei Sun, Biao Gong, Yanwei Fu, Danda~Pani
  Paudel, Xuanjing Huang, and Luc Van~Gool.
\newblock Objectrelator: Enabling cross-view object relation understanding
  across ego-centric and exo-centric perspectives.
\newblock In \emph{ICCV}, 2025.

\bibitem[Gharaee et~al.(2024)Gharaee, Gong, Pellegrino, Zarubiieva, Haurum,
  Lowe, McKeown, Ho, McLeod, Wei, Agda, Ratnasingham, Steinke, Chang, Taylor,
  and Fieguth]{BIOSCAN}
Zahra Gharaee, ZeMing Gong, Nicholas Pellegrino, Iuliia Zarubiieva,
  Joakim~Bruslund Haurum, Scott~C. Lowe, Jaclyn T.~A. McKeown, Chris C.~Y. Ho,
  Joschka McLeod, Yi-Yun~C Wei, Jireh Agda, Sujeevan Ratnasingham, Dirk
  Steinke, Angel~X. Chang, Graham~W. Taylor, and Paul Fieguth.
\newblock A step towards worldwide biodiversity assessment: The bioscan-1m
  insect dataset.
\newblock \emph{Advances in Neural Information Processing Systems}, 36, 2024.
\newblock URL \url{http://arxiv.org/abs/2307.10455}.

\bibitem[González-Rivero et~al.(2019)González-Rivero, Rodriguez-Ramirez,
  Beijbom, Dalton, Kennedy, Neal, Vercelloni, Bongaerts, Ganase, Bryant, Brown,
  Kim, Radice, Lopez-Marcano, Dove, Bailhache, Beyer, and
  Hoegh-Guldberg]{gonzalez2019seaview}
Manuel González-Rivero, Alberto Rodriguez-Ramirez, Oscar Beijbom, Peter
  Dalton, Emma~V. Kennedy, Benjamin~P. Neal, Julie Vercelloni, Pim Bongaerts,
  Anjani Ganase, Dominic~E.P. Bryant, Kristen Brown, Catherine Kim, Veronica~Z.
  Radice, Sebastian Lopez-Marcano, Sophie Dove, Christophe Bailhache,
  Hawthorne~L. Beyer, and Ove Hoegh-Guldberg.
\newblock Seaview survey photo-quadrat and image classification dataset, 2019.
\newblock URL \url{https://espace.library.uq.edu.au/view/UQ:734799}.

\bibitem[Gu et~al.(2025{\natexlab{a}})Gu, Stevens, Campolongo, Thompson, Zhang,
  Wu, Kopanev, Mai, White, Balhoff, Dahdul, Rubenstein, Lapp, Berger-Wolf,
  Chao, and Su]{bioclip2}
Jianyang Gu, Samuel Stevens, Elizabeth~G Campolongo, Matthew~J Thompson, Net
  Zhang, Jiaman Wu, Andrei Kopanev, Zheda Mai, Alexander~E. White, James
  Balhoff, Wasila Dahdul, Daniel Rubenstein, Hilmar Lapp, Tanya Berger-Wolf,
  Wei-Lun Chao, and Yu~Su.
\newblock Bio{CLIP} 2: Emergent properties from scaling hierarchical
  contrastive learning.
\newblock In \emph{The Thirty-ninth Annual Conference on Neural Information
  Processing Systems}, 2025{\natexlab{a}}.
\newblock URL \url{https://openreview.net/forum?id=yPC9zmkQgG}.

\bibitem[Gu et~al.(2025{\natexlab{b}})Gu, Stevens, Campolongo, Thompson, Zhang,
  Wu, Kopanev, Mai, White, Balhoff, Dahdul, Rubenstein, Lapp, Berger-Wolf,
  Chao, and Su]{treeoflife_200m}
Jianyang Gu, Samuel Stevens, Elizabeth~G Campolongo, Matthew~J Thompson, Net
  Zhang, Jiaman Wu, Andrei Kopanev, Zheda Mai, Alexander~E. White, James
  Balhoff, Wasila~M Dahdul, Daniel Rubenstein, Hilmar Lapp, Tanya Berger-Wolf,
  Wei-Lun Chao, and Yu~Su.
\newblock {T}ree{O}f{L}ife-200{M} (revision a8f38b4), 2025{\natexlab{b}}.
\newblock URL \url{https://huggingface.co/datasets/imageomics/TreeOfLife-200M}.

\bibitem[He et~al.(2016)He, Zhang, Ren, and Sun]{he2016deep}
Kaiming He, Xiangyu Zhang, Shaoqing Ren, and Jian Sun.
\newblock Deep residual learning for image recognition.
\newblock In \emph{Proceedings of the IEEE conference on computer vision and
  pattern recognition}, pages 770--778, 2016.
\newblock URL \url{https://doi.org/10.48550/arXiv.1512.03385}.

\bibitem[He et~al.(2022)He, Chen, Xie, Li, Doll{\'a}r, and
  Girshick]{he2022masked}
Kaiming He, Xinlei Chen, Saining Xie, Yanghao Li, Piotr Doll{\'a}r, and Ross
  Girshick.
\newblock Masked autoencoders are scalable vision learners.
\newblock In \emph{Proceedings of the IEEE/CVF conference on computer vision
  and pattern recognition}, pages 16000--16009, 2022.
\newblock URL \url{https://doi.org/10.48550/arXiv.2111.06377}.

\bibitem[Hill and Wilkinson(2004)]{hill2004methods}
Jos Hill and CLIVE Wilkinson.
\newblock Methods for ecological monitoring of coral reefs.
\newblock Technical report, Australian Institute of Marine Science, 2004.

\bibitem[Horn et~al.(2021)Horn, Cole, Beery, Wilber, Belongie, and
  Aodha]{iNat2021}
Grant~Van Horn, Elijah Cole, Sara Beery, Kimberly Wilber, Serge Belongie, and
  Oisin~Mac Aodha.
\newblock Benchmarking representation learning for natural world image
  collections.
\newblock In \emph{Proceedings of the IEEE/CVF conference on computer vision
  and pattern recognition}, 3 2021.
\newblock URL \url{http://arxiv.org/abs/2103.16483}.

\bibitem[Hughes et~al.(2017)Hughes, Barnes, Bellwood, Cinner, Cumming, Jackson,
  Kleypas, van~de Leemput, Lough, Morrison, Palumbi, van Nes, and
  Scheffer]{hughes2017coral}
Terry~P. Hughes, Michele~L. Barnes, David~R. Bellwood, Joshua~E. Cinner,
  Graeme~S. Cumming, Jeremy B.~C. Jackson, Joanie Kleypas, Ingrid~A. van~de
  Leemput, Janice~M. Lough, Tiffany~H. Morrison, Stephen~R. Palumbi, Egbert~H.
  van Nes, and Marten Scheffer.
\newblock Coral reefs in the anthropocene.
\newblock \emph{Nature}, 546:\penalty0 82--90, 6 2017.
\newblock ISSN 0028-0836.
\newblock \doi{10.1038/nature22901}.
\newblock URL \url{https://www.nature.com/articles/nature22901}.

\bibitem[Ilharco et~al.(2021)Ilharco, Wortsman, Wightman, Gordon, Carlini,
  Taori, Dave, Shankar, Namkoong, Miller, Hajishirzi, Farhadi, and
  Schmidt]{openclip}
Gabriel Ilharco, Mitchell Wortsman, Ross Wightman, Cade Gordon, Nicholas
  Carlini, Rohan Taori, Achal Dave, Vaishaal Shankar, Hongseok Namkoong, John
  Miller, Hannaneh Hajishirzi, Ali Farhadi, and Ludwig Schmidt.
\newblock Openclip, July 2021.
\newblock URL \url{https://doi.org/10.5281/zenodo.5143773}.

\bibitem[Li et~al.(2024{\natexlab{a}})Li, Zhang, Guo, Zhang, Li, Zhang, Zhang,
  Zhang, Li, Liu, and Li]{llavaone}
Bo~Li, Yuanhan Zhang, Dong Guo, Renrui Zhang, Feng Li, Hao Zhang, Kaichen
  Zhang, Peiyuan Zhang, Yanwei Li, Ziwei Liu, and Chunyuan Li.
\newblock Llava-onevision: Easy visual task transfer, 2024{\natexlab{a}}.
\newblock URL \url{https://arxiv.org/abs/2408.03326}.

\bibitem[Li et~al.(2024{\natexlab{b}})Li, Zhang, Zhang, Zhang, Li, Li, Ma, and
  Li]{llavanext}
Feng Li, Renrui Zhang, Hao Zhang, Yuanhan Zhang, Bo~Li, Wei Li, Zejun Ma, and
  Chunyuan Li.
\newblock Llava-next-interleave: Tackling multi-image, video, and 3d in large
  multimodal models, 2024{\natexlab{b}}.
\newblock URL \url{https://arxiv.org/abs/2407.07895}.

\bibitem[Liu et~al.(2021)Liu, Lin, Cao, Hu, Wei, Zhang, Lin, and Guo]{swin}
Ze~Liu, Yutong Lin, Yue Cao, Han Hu, Yixuan Wei, Zheng Zhang, Stephen Lin, and
  Baining Guo.
\newblock { Swin Transformer: Hierarchical Vision Transformer using Shifted
  Windows }.
\newblock In \emph{2021 IEEE/CVF International Conference on Computer Vision
  (ICCV)}, pages 9992--10002, Los Alamitos, CA, USA, October 2021. IEEE
  Computer Society.
\newblock \doi{10.1109/ICCV48922.2021.00986}.
\newblock URL
  \url{https://doi.ieeecomputersociety.org/10.1109/ICCV48922.2021.00986}.

\bibitem[Liu et~al.(2022)Liu, Mao, Wu, Feichtenhofer, Darrell, and
  Xie]{liu2022convnet}
Zhuang Liu, Hanzi Mao, Chao-Yuan Wu, Christoph Feichtenhofer, Trevor Darrell,
  and Saining Xie.
\newblock A convnet for the 2020s.
\newblock In \emph{Proceedings of the IEEE/CVF Conference on Computer Vision
  and Pattern Recognition}, pages 11976--11986, 2022.
\newblock URL \url{https://doi.org/10.48550/arXiv.2201.03545}.

\bibitem[Lowe et~al.(2024)Lowe, Misiuk, Xu, Abdulazizov, Baroi, Bastos, Best,
  Ferrini, Friedman, Hart, Hoegh-Guldberg, Ierodiaconou, Mackin-McLaughlin,
  Markey, Menandro, Monk, Nemani, O'Brien, Oh, Reshitnyk, Robert, Roelfsema,
  Sameoto, Schimel, Thomson, Wilson, Wong, Brown, and Trappenberg]{BenthicNet}
Scott~C. Lowe, Benjamin Misiuk, Isaac Xu, Shakhboz Abdulazizov, Amit~R. Baroi,
  Alex~C. Bastos, Merlin Best, Vicki Ferrini, Ariell Friedman, Deborah Hart,
  Ove Hoegh-Guldberg, Daniel Ierodiaconou, Julia Mackin-McLaughlin, Kathryn
  Markey, Pedro~S. Menandro, Jacquomo Monk, Shreya Nemani, John O'Brien,
  Elizabeth Oh, Luba~Y. Reshitnyk, Katleen Robert, Chris~M. Roelfsema,
  Jessica~A. Sameoto, Alexandre C.~G. Schimel, Jordan~A. Thomson, Brittany~R.
  Wilson, Melisa~C. Wong, Craig~J. Brown, and Thomas Trappenberg.
\newblock Benthicnet: A global compilation of seafloor images for deep learning
  applications, 5 2024.
\newblock URL \url{http://arxiv.org/abs/2405.05241}.

\bibitem[Miao et~al.(2024)Miao, Engelmann, Vysotska, Tombari, Pollefeys, and
  Bar{\'a}th]{scenegraphloc}
Yang Miao, Francis Engelmann, Olga Vysotska, Federico Tombari, Marc Pollefeys,
  and D{\'a}niel~B{\'e}la Bar{\'a}th.
\newblock Scenegraphloc: Cross-modal coarse visual localization on 3d scene
  graphs.
\newblock In \emph{ECCV}, 2024.

\bibitem[Miao et~al.(2026)Miao, Zaech, Wang, Despinoy, Paudel, and
  Gool]{miao2026langhopslanguagegroundedhierarchical}
Yang Miao, Jan-Nico Zaech, Xi~Wang, Fabien Despinoy, Danda~Pani Paudel, and
  Luc~Van Gool.
\newblock Langhops: Language grounded hierarchical open-vocabulary part
  segmentation.
\newblock In \emph{NeurIPS}, 2026.

\bibitem[{NOAA Pacific Islands Fisheries Science Center, Ecosystem Sciences
  Division}(2018{\natexlab{a}})]{NOAA_NCRMP_2018}
{NOAA Pacific Islands Fisheries Science Center, Ecosystem Sciences Division}.
\newblock {National Coral Reef Monitoring Program: Benthic cover derived from
  analysis of images collected during stratified random surveys (StRS) across
  the Mariana Archipelago}, 2018{\natexlab{a}}.
\newblock URL
  \url{https://www.ncei.noaa.gov/access/metadata/landing-page/bin/iso?id=gov.noaa.nodc:NCRMP-StRS-Images-Marianas}.

\bibitem[{NOAA Pacific Islands Fisheries Science Center, Ecosystem Sciences
  Division}(2018{\natexlab{b}})]{NOAA_NCRMP_HI_2018}
{NOAA Pacific Islands Fisheries Science Center, Ecosystem Sciences Division}.
\newblock {National Coral Reef Monitoring Program: Benthic cover derived from
  analysis of images collected during stratified random surveys (StRS) of the
  Hawaiian Archipelago}, 2018{\natexlab{b}}.
\newblock URL
  \url{https://www.ncei.noaa.gov/access/metadata/landing-page/bin/iso?id=gov.noaa.nodc:NCRMP-StRS-Images-HI}.

\bibitem[{NOAA Pacific Islands Fisheries Science Center, Ecosystem Sciences
  Division}(2018{\natexlab{c}})]{NOAA_NCRMP_SAMOA_2018}
{NOAA Pacific Islands Fisheries Science Center, Ecosystem Sciences Division}.
\newblock {National Coral Reef Monitoring Program: Benthic cover derived from
  analysis of images collected during stratified random surveys (StRS) across
  American Samoa}, 2018{\natexlab{c}}.
\newblock URL
  \url{https://www.ncei.noaa.gov/access/metadata/landing-page/bin/iso?id=gov.noaa.nodc:NCRMP-StRS-Images-AmSam}.

\bibitem[Oquab et~al.(2024)Oquab, Darcet, Moutakanni, Vo, Szafraniec, Khalidov,
  Fernandez, Haziza, Massa, El-Nouby, Assran, Ballas, Galuba, Howes, Huang, Li,
  Misra, Rabbat, Sharma, Synnaeve, Xu, Jegou, Mairal, Labatut, Joulin, and
  Bojanowski]{dinov2}
Maxime Oquab, Timothée Darcet, Théo Moutakanni, Huy Vo, Marc Szafraniec,
  Vasil Khalidov, Pierre Fernandez, Daniel Haziza, Francisco Massa, Alaaeldin
  El-Nouby, Mahmoud Assran, Nicolas Ballas, Wojciech Galuba, Russell Howes,
  Po-Yao Huang, Shang-Wen Li, Ishan Misra, Michael Rabbat, Vasu Sharma, Gabriel
  Synnaeve, Hu~Xu, Hervé Jegou, Julien Mairal, Patrick Labatut, Armand Joulin,
  and Piotr Bojanowski.
\newblock Dinov2: Learning robust visual features without supervision, 2024.
\newblock URL \url{https://arxiv.org/abs/2304.07193}.

\bibitem[Pan et~al.(2025)Pan, Liu, Fu, Ma, Li, Paudel, Van~Gool, and
  Huang]{pan2025locate}
Jiancheng Pan, Yanxing Liu, Yuqian Fu, Muyuan Ma, Jiahao Li, Danda~Pani Paudel,
  Luc Van~Gool, and Xiaomeng Huang.
\newblock Locate anything on earth: Advancing open-vocabulary object detection
  for remote sensing community.
\newblock In \emph{AAAI}, 2025.

\bibitem[Pedregosa et~al.(2011)Pedregosa, Varoquaux, Gramfort, Michel, Thirion,
  Grisel, Blondel, Prettenhofer, Weiss, Dubourg, Vanderplas, Passos,
  Cournapeau, Brucher, Perrot, and Duchesnay]{scikit-learn}
F.~Pedregosa, G.~Varoquaux, A.~Gramfort, V.~Michel, B.~Thirion, O.~Grisel,
  M.~Blondel, P.~Prettenhofer, R.~Weiss, V.~Dubourg, J.~Vanderplas, A.~Passos,
  D.~Cournapeau, M.~Brucher, M.~Perrot, and E.~Duchesnay.
\newblock Scikit-learn: Machine learning in {P}ython.
\newblock \emph{Journal of Machine Learning Research}, 12:\penalty0 2825--2830,
  2011.

\bibitem[Radford et~al.(2021)Radford, Kim, Hallacy, Ramesh, Goh, Agarwal,
  Sastry, Askell, Mishkin, Clark, Krueger, and Sutskever]{clip}
Alec Radford, Jong~Wook Kim, Chris Hallacy, Aditya Ramesh, Gabriel Goh,
  Sandhini Agarwal, Girish Sastry, Amanda Askell, Pamela Mishkin, Jack Clark,
  Gretchen Krueger, and Ilya Sutskever.
\newblock Learning transferable visual models from natural language
  supervision, 2021.
\newblock URL \url{https://arxiv.org/abs/2103.00020}.

\bibitem[Raphael et~al.(2020)Raphael, Dubinsky, Iluz, and
  Papathomas]{Raphael2020eilat}
A.~Raphael, Z.~Dubinsky, D.~Iluz, and M.~Papathomas.
\newblock Deep neural network recognition of shallow water corals in the gulf
  of eilat (aqaba).
\newblock \emph{Scientific Reports}, 10\penalty0 (1):\penalty0 12959, 2020.
\newblock \doi{10.1038/s41598-020-69201-w}.
\newblock URL \url{https://www.nature.com/articles/s41598-020-69201-w}.

\bibitem[Ren et~al.(2023)Ren, Liu, Song, Bi, Cucchiara, Sebe, and
  Wang]{ren2023masked}
Bin Ren, Yahui Liu, Yue Song, Wei Bi, Rita Cucchiara, Nicu Sebe, and Wei Wang.
\newblock Masked jigsaw puzzle: A versatile position embedding for vision
  transformers.
\newblock In \emph{CVPR}, 2023.

\bibitem[Ren et~al.(2025)Ren, Li, Zheng, Fu, Paudel, Yang, Van~Gool, and
  Sebe]{ren2025manifold}
Bin Ren, Yawei Li, Xu~Zheng, Yuqian Fu, Danda~Pani Paudel, Ming-Hsuan Yang, Luc
  Van~Gool, and Nicu Sebe.
\newblock Manifold-aware representation learning for degradation-agnostic image
  restoration.
\newblock \emph{arXiv preprint arXiv:2505.18679}, 2025.

\bibitem[Sauder et~al.(2025)Sauder, Domazetoski, Banc-Prandi, Perna, Meibom,
  and Tuia]{sauder2025coralscapesdatasetsemanticscene}
Jonathan Sauder, Viktor Domazetoski, Guilhem Banc-Prandi, Gabriela Perna,
  Anders Meibom, and Devis Tuia.
\newblock The coralscapes dataset: Semantic scene understanding in coral reefs,
  2025.
\newblock URL \url{https://arxiv.org/abs/2503.20000}.

\bibitem[Sechidis et~al.(2011)Sechidis, Tsoumakas, and
  Vlahavas]{sechidis2011stratification}
Konstantinos Sechidis, Grigorios Tsoumakas, and Ioannis Vlahavas.
\newblock On the stratification of multi-label data.
\newblock In Dimitrios Gunopulos, Thomas Hofmann, Donato Malerba, and Michalis
  Vazirgiannis, editors, \emph{Machine Learning and Knowledge Discovery in
  Databases}, volume 6913 of \emph{Lecture Notes in Computer Science}, pages
  145--158. Springer, Berlin, Heidelberg, 2011.
\newblock \doi{10.1007/978-3-642-23808-6_10}.

\bibitem[Sim{\'e}oni et~al.(2025)Sim{\'e}oni, Vo, Seitzer, Baldassarre, Oquab,
  Jose, Khalidov, Szafraniec, Yi, Ramamonjisoa, Massa, Haziza, Wehrstedt, Wang,
  Darcet, Moutakanni, Sentana, Roberts, Vedaldi, Tolan, Brandt, Couprie,
  Mairal, J{\'e}gou, Labatut, and Bojanowski]{dinov3}
Oriane Sim{\'e}oni, Huy~V. Vo, Maximilian Seitzer, Federico Baldassarre, Maxime
  Oquab, Cijo Jose, Vasil Khalidov, Marc Szafraniec, Seungeun Yi, Micha{\"e}l
  Ramamonjisoa, Francisco Massa, Daniel Haziza, Luca Wehrstedt, Jianyuan Wang,
  Timoth{\'e}e Darcet, Th{\'e}o Moutakanni, Leonel Sentana, Claire Roberts,
  Andrea Vedaldi, Jamie Tolan, John Brandt, Camille Couprie, Julien Mairal,
  Herv{\'e} J{\'e}gou, Patrick Labatut, and Piotr Bojanowski.
\newblock Dinov3, 2025.
\newblock URL \url{https://arxiv.org/abs/2508.10104}.

\bibitem[Spalding et~al.(2017)Spalding, Burke, Wood, Ashpole, Hutchison, and
  zu~Ermgassen]{spalding2017global}
Mark Spalding, Lauretta Burke, Spencer~A. Wood, Joscelyne Ashpole, James
  Hutchison, and Philine zu~Ermgassen.
\newblock Mapping the global value and distribution of coral reef tourism.
\newblock \emph{Marine Policy}, 82:\penalty0 104--113, 8 2017.
\newblock ISSN 0308597X.
\newblock \doi{10.1016/j.marpol.2017.05.014}.
\newblock URL
  \url{https://linkinghub.elsevier.com/retrieve/pii/S0308597X17300635}.

\bibitem[Spalding et~al.(2007)Spalding, Fox, Allen, Davidson, Ferdaña,
  Finlayson, Halpern, Jorge, Lombana, Lourie, Martin, McManus, Molnar, Recchia,
  and Robertson]{spalding_ecoregions}
Mark~D. Spalding, Helen~E. Fox, Gerald~R. Allen, Nick Davidson, Zach~A.
  Ferdaña, Max Finlayson, Benjamin~S. Halpern, Miguel~A. Jorge, Al~Lombana,
  Sara~A. Lourie, Kirsten~D. Martin, Edmund McManus, Jennifer Molnar, Cheri~A.
  Recchia, and James Robertson.
\newblock {Marine Ecoregions of the World: A Bioregionalization of Coastal and
  Shelf Areas}.
\newblock \emph{BioScience}, 57\penalty0 (7):\penalty0 573--583, 07 2007.
\newblock \doi{10.1641/B570707}.
\newblock URL \url{https://doi.org/10.1641/B570707}.

\bibitem[Stevens et~al.(2023{\natexlab{a}})Stevens, Wu, Thompson, Campolongo,
  Song, Carlyn, Dong, Dahdul, Stewart, Berger-Wolf, Chao, and Su]{BioCLIP}
Samuel Stevens, Jiaman Wu, Matthew~J Thompson, Elizabeth~G Campolongo, Chan~Hee
  Song, David~Edward Carlyn, Li~Dong, Wasila~M Dahdul, Charles Stewart, Tanya
  Berger-Wolf, Wei-Lun Chao, and Yu~Su.
\newblock Bioclip: A vision foundation model for the tree of life, 11
  2023{\natexlab{a}}.
\newblock URL \url{http://arxiv.org/abs/2311.18803}.

\bibitem[Stevens et~al.(2023{\natexlab{b}})Stevens, Wu, Thompson, Campolongo,
  Song, Carlyn, Dong, Dahdul, Stewart, Berger-Wolf, Chao, and
  Su]{treeoflife_10m}
Samuel Stevens, Jiaman Wu, Matthew~J Thompson, Elizabeth~G Campolongo, Chan~Hee
  Song, David~Edward Carlyn, Li~Dong, Wasila~M Dahdul, Charles Stewart, Tanya
  Berger-Wolf, Wei-Lun Chao, and Yu~Su.
\newblock Treeoflife-10m, 2023{\natexlab{b}}.
\newblock URL \url{https://huggingface.co/datasets/imageomics/TreeOfLife-10M}.

\bibitem[Tan and Le(2019)]{tan2019efficientnet}
Mingxing Tan and Quoc Le.
\newblock Efficientnet: Rethinking model scaling for convolutional neural
  networks.
\newblock In \emph{International Conference on Machine Learning}, pages
  6105--6114. PMLR, 2019.
\newblock \doi{10.48550/arXiv.1905.11946}.
\newblock URL \url{https://doi.org/10.48550/arXiv.1905.11946}.

\bibitem[Touvron et~al.(2021)Touvron, Cord, Douze, Massa, Sablayrolles, and
  Jegou]{deit}
Hugo Touvron, Matthieu Cord, Matthijs Douze, Francisco Massa, Alexandre
  Sablayrolles, and Herve Jegou.
\newblock Training data-efficient image transformers \&; distillation through
  attention.
\newblock In Marina Meila and Tong Zhang, editors, \emph{Proceedings of the
  38th International Conference on Machine Learning}, volume 139 of
  \emph{Proceedings of Machine Learning Research}, pages 10347--10357. PMLR,
  18--24 Jul 2021.
\newblock URL \url{https://proceedings.mlr.press/v139/touvron21a.html}.

\bibitem[Tschannen et~al.(2025)Tschannen, Gritsenko, Wang, Naeem,
  Alabdulmohsin, Parthasarathy, Evans, Beyer, Xia, Mustafa, H{\'e}naff,
  Harmsen, Steiner, and Zhai]{siglip2}
Michael Tschannen, Alexey Gritsenko, Xiao Wang, Muhammad~Ferjad Naeem, Ibrahim
  Alabdulmohsin, Nikhil Parthasarathy, Talfan Evans, Lucas Beyer, Ye~Xia, Basil
  Mustafa, Olivier H{\'e}naff, Jeremiah Harmsen, Andreas Steiner, and Xiaohua
  Zhai.
\newblock Siglip 2: Multilingual vision-language encoders with improved
  semantic understanding, localization, and dense features, 2025.
\newblock URL \url{https://arxiv.org/abs/2502.14786}.

\bibitem[Van~Horn et~al.(2015)Van~Horn, Branson, Farrell, Haber, Barry,
  Ipeirotis, Perona, and Belongie]{vanhorn2015nabirds}
Grant Van~Horn, Steve Branson, Ryan Farrell, Sam Haber, Jessie Barry, Panos
  Ipeirotis, Pietro Perona, and Serge Belongie.
\newblock {Building a Bird Recognition App and Large Scale Dataset with Citizen
  Scientists: The Fine Print in Fine-Grained Dataset Collection}.
\newblock In \emph{CVPR}, 2015.

\bibitem[Van~Horn et~al.(2018)Van~Horn, Mac~Aodha, Song, Cui, Sun, Shepard,
  Adam, Perona, and Belongie]{vanhorn2018inaturalist}
Grant Van~Horn, Oisin Mac~Aodha, Yang Song, Yin Cui, Chen Sun, Alex Shepard,
  Hartwig Adam, Pietro Perona, and Serge Belongie.
\newblock {The iNaturalist Species Classification and Detection Dataset}.
\newblock In \emph{CVPR}, pages 8769--8778, 2018.

\bibitem[Wah et~al.(2011)Wah, Branson, Welinder, Perona, and
  Belongie]{wah2011cub}
Catherine Wah, Steve Branson, Peter Welinder, Pietro Perona, and Serge
  Belongie.
\newblock {The Caltech-UCSD Birds-200-2011 Dataset}.
\newblock In \emph{Computation \& Neural Systems Technical Report}, 2011.

\bibitem[Wang et~al.(2025)Wang, Gao, Gu, Pu, Cui, Wei, Liu, et~al.]{intern35}
Weiyun Wang, Zhangwei Gao, Lixin Gu, Hengjun Pu, Long Cui, Xingguang Wei,
  Zhaoyang Liu, et~al.
\newblock Internvl3.5: Advancing open-source multimodal models in versatility,
  reasoning, and efficiency, 2025.
\newblock URL \url{https://arxiv.org/abs/2508.18265}.

\bibitem[Williams et~al.(2019)Williams, Couch, Beijbom, Oliver, Vargas-Angel,
  Schumacher, and Brainard]{williams2019coralnet}
Ivor~D. Williams, Courtney~S. Couch, Oscar Beijbom, Thomas~A. Oliver, Bernardo
  Vargas-Angel, Brett~D. Schumacher, and Russell~E. Brainard.
\newblock Leveraging automated image analysis tools to transform our capacity
  to assess status and trends of coral reefs.
\newblock \emph{Frontiers in Marine Science}, 6, 4 2019.
\newblock ISSN 2296-7745.
\newblock \doi{10.3389/fmars.2019.00222}.
\newblock URL
  \url{https://www.frontiersin.org/article/10.3389/fmars.2019.00222/full}.

\bibitem[Wu et~al.(2019)Wu, Hu, Guo, Liu, Zhao, Chen, Ji, et~al.]{wu2019ip102}
Jia Wu, Xixu Hu, Jie Guo, Jie Liu, Qiong Zhao, Zhen Chen, Xiangyang Ji, et~al.
\newblock {IP102: A Large-Scale Benchmark Dataset for Insect Pest Recognition}.
\newblock In \emph{CVPR Workshops}, 2019.

\bibitem[Yao et~al.(2024)Yao, Yu, Zhang, Wang, Cui, Zhu, Cai, Li, Zhao, He,
  et~al.]{minicpm}
Yuan Yao, Tianyu Yu, Ao~Zhang, Chongyi Wang, Junbo Cui, Hongji Zhu, Tianchi
  Cai, Haoyu Li, Weilin Zhao, Zhihui He, et~al.
\newblock Minicpm-v: A gpt-4v level mllm on your phone.
\newblock \emph{arXiv preprint arXiv:2408.01800}, 2024.

\bibitem[Yu et~al.(2023)Yu, Zhang, Xie, and Zhang]{yu2023point}
Haitao Yu, Deheng Zhang, Peiyuan Xie, and Tianyi Zhang.
\newblock Point-based radiance fields for controllable human motion synthesis.
\newblock \emph{arXiv preprint arXiv:2310.03375}, 2023.

\bibitem[Zhai et~al.(2023)Zhai, Mustafa, Kolesnikov, and Beyer]{siglip}
Xiaohua Zhai, Basil Mustafa, Alexander Kolesnikov, and Lucas Beyer.
\newblock Sigmoid loss for language image pre-training, 2023.
\newblock URL \url{https://arxiv.org/abs/2303.15343}.

\bibitem[Zheng et~al.(2023{\natexlab{a}})Zheng, Liu, Lu, Hua, Pan, Zhang, Tao,
  and Wang]{zheng2023deep}
Xu~Zheng, Yexin Liu, Yunfan Lu, Tongyan Hua, Tianbo Pan, Weiming Zhang, Dacheng
  Tao, and Lin Wang.
\newblock Deep learning for event-based vision: A comprehensive survey and
  benchmarks.
\newblock \emph{arXiv preprint arXiv:2302.08890}, 2023{\natexlab{a}}.

\bibitem[Zheng et~al.(2023{\natexlab{b}})Zheng, Zhu, Liu, Cao, Fu, and
  Wang]{zheng2023both}
Xu~Zheng, Jinjing Zhu, Yexin Liu, Zidong Cao, Chong Fu, and Lin Wang.
\newblock Both style and distortion matter: Dual-path unsupervised domain
  adaptation for panoramic semantic segmentation.
\newblock In \emph{CVPR}, 2023{\natexlab{b}}.

\bibitem[Zheng et~al.(2024)Zheng, Liang, Hua, Wong, Ang, Pui, Chui, and
  Yeung]{coralscop}
Ziqiang Zheng, Haixin Liang, Binh-Son Hua, Yue~Him Wong, Put Ang, Apple Pui,
  Yi~Chui, and Sai-Kit Yeung.
\newblock Coralscop: Segment any coral image on this planet.
\newblock In \emph{Proceedings of the IEEE/CVF Conference on Computer Vision
  and Pattern Recognition}, 2024.
\newblock URL \url{https://coralscop.hkustvgd.com}.

\bibitem[Ziqiang et~al.(2023)Ziqiang, Yaofeng, Haixin, Zhibin, and
  Yeung]{coralvos}
Zheng Ziqiang, Xie Yaofeng, Liang Haixin, Yu~Zhibin, and Sai-Kit Yeung.
\newblock Coralvos: Dataset and benchmark for coral video segmentation, 2023.

\end{thebibliography}

\clearpage
\appendix

\noindent This appendix is organized as follows.
Appendix~\ref{appendix_reefnet_in_relation_to_coralnet} positions ReefNet relative to CoralNet, detailing the limitations of CoralNet for ML use and how ReefNet addresses them.
Appendix~\ref{sup_data_collection} describes the full ReefNet data collection methodology, including the multi-stage source filtering pipeline.
Appendix~\ref{sup_quality_control} details the annotation quality control process and the custom verification platform.
Appendix~\ref{sup_verification_protocol} provides further details on the centralized re-verification protocol, reviewer background, and sampling strategy.
Appendix~\ref{sup_covariate_diversity} reports covariate diversity across ecoregions, image white balance, and resolution statistics.
Appendix~\ref{sup_thealwajh_testset} describes the Al-Wajh Lagoon dataset used as the cross-source benchmark test set.
Appendix~\ref{sec:exp_detail} provides available training details, including vision model architectures, MLLM LoRA fine-tuning setup, and zero-shot prompting protocols.
Appendix~\ref{sec:more_results} presents additional quantitative results, including per-class recall, overall accuracy, and precision/F1 scores.
Appendix~\ref{sup_text_descriptions} describes the text and reasoning experiments. Finally, we provide historical qualitative examples, limitations of the dataset, contributor attribution, and information on data and code availability.

\section{ReefNet in Relation to CoralNet}
\label{appendix_reefnet_in_relation_to_coralnet}
Despite CoralNet's~\cite{beijbom2015coralnet} wide adoption as a platform for coral annotations and dataset publication, it presents several challenges that limit its usability for machine-learning practitioners. This stands in contrast to other biological domains, where the community has built well-established machine-learning-ready datasets (e.g., CUB-200-2011, NABirds, iNaturalist, IP102)~\cite{wah2011cub,vanhorn2015nabirds,vanhorn2018inaturalist,wu2019ip102}. ReefNet addresses these shortcomings, making it a more reliable resource for scientific exploration and algorithmic development. The key limitations of CoralNet are:

\noindent\textbf{- Lack of a unified label set across sources.} CoralNet aggregates data with heterogeneous and often incompatible label sets, making large-scale integration infeasible.

\noindent\textbf{- Absence of taxonomically verified hard coral labels.} Many sources use outdated, ambiguous, or generic labels instead of scientifically recognized names (e.g., those in the World Register of Marine Species).

\noindent\textbf{- No standardized quality control.} Without systematic quality checks, it is difficult to identify reliable samples, which is critical for both training and evaluation.

\noindent\textbf{- No large-scale benchmark framework.} CoralNet does not provide a standardized evaluation setting for consistent model comparison.

\begin{figure}[!ht]
  \centering
  \includegraphics[width=\linewidth]{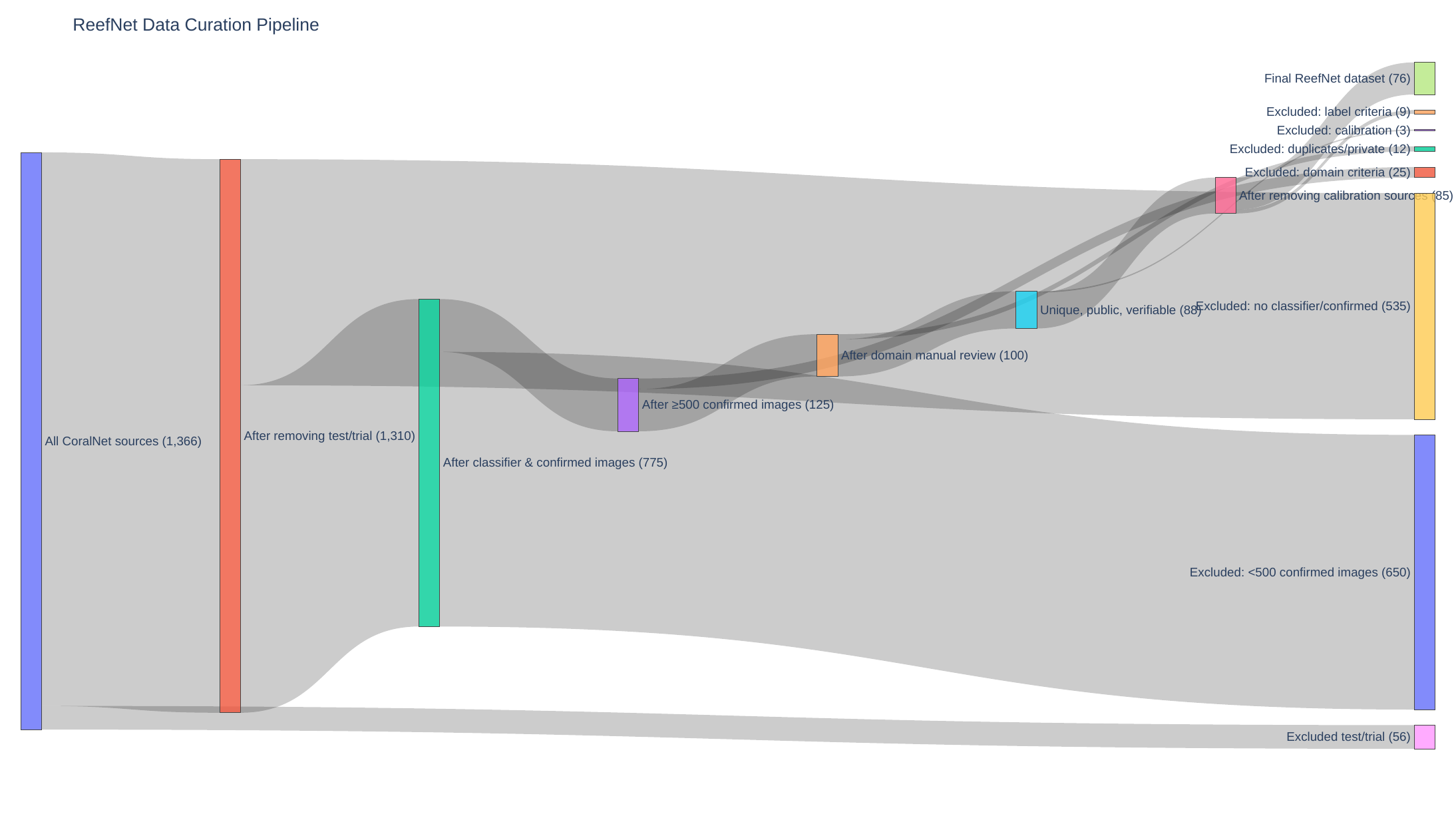}
  \caption{\textbf{ReefNet Curation Pipeline.}
  The diagram illustrates the sequential filtering and exclusion steps applied to 1,366 CoralNet image sources to arrive at the final ReefNet dataset of 76 sources.
  Sources were excluded due to an insufficient number of human-verified annotations, ecological or privacy concerns, and calibration-related issues.
  }

  \label{data_pipeline_static_friendly}
\end{figure}

To address these challenges, we propose ReefNet with the following key contributions:

\noindent\textbf{- World Register of Marine Species (WoRMS)~\cite{worms}-based unified taxonomic labeling:}
ReefNet adopts a unified labeling scheme grounded in WoRMS to integrate heterogeneous sources. This harmonization also enabled the inclusion of a Red Sea–focused dataset. We anticipate that this taxonomy will encourage future work to adopt scientifically sound annotation practices.

\noindent\textbf{- Rigorous re-verification:}
CoralNet sources were annotated by distinct groups with variable expertise, leading to inconsistencies in label provenance. While acceptable within individual sources, these inconsistencies hinder multi-source integration. ReefNet mitigates this issue by introducing a centralized verification step: a dedicated review team, composed of individuals with complementary backgrounds, re-evaluated all included sources. This ensures a consistent, cross-source quality standard unattainable through direct dataset merging.

\noindent\textbf{- Standardized benchmarks:}
ReefNet separates three adaptation regimes (zero-shot recognition, few-shot adaptation of pretrained models, and full fine-tuning) from two evaluation settings. Within-source evaluation uses ReefNet's 39 labels; the standard cross-source protocol retains the same 39-way output on Al-Wajh. The Claude/GPT API prompts instead list only the 13 Al-Wajh genera, so their Al-Wajh scores are not candidate-set matched to the standard protocol (Appendix~\ref{sup_api_models}). Predictions outside its 13 present genera count as errors, and macro recall is averaged over those 13 ground-truth genera. Neither the Al-Wajh images nor their annotations are used for training, validation, hyperparameter tuning, or checkpoint selection. Source-level metadata and image-grouped splits support study of heterogeneous surveys without establishing global geographic robustness.

\section{Additional Information on the Dataset}

\subsection{ReefNet Data Collection Methodology}
\label{sup_data_collection}
The \textbf{ReefNet} dataset was constructed through a multi-stage curation pipeline applied to publicly available sources hosted on \textit{CoralNet}. Figure~\ref{data_pipeline_static_friendly} summarizes this pipeline. Beginning with all \textbf{1,366 publicly listed CoralNet sources}, we applied a series of semantic, ecological, and technical filters to isolate high-quality, taxonomically relevant reef imagery and annotations. These steps ensured that retained sources featured dense, consistent, and biologically meaningful annotation data.

We first excluded sources with names containing keywords such as \textit{``test''} or \textit{``trial''}---commonly created by users experimenting with the platform---reducing the pool to \textbf{1,310 sources}. An exception was made for \textit{CoralNet Assistance Test}, which was retained after manual inspection confirmed its annotation reliability and ecological relevance.

Next, we removed sources lacking a trained classifier or any \textit{confirmed} (i.e., human-annotated) images, yielding \textbf{775 sources}. Applying a minimum threshold of \textbf{500 confirmed images} further reduced the set to \textbf{125 sources}. These were manually reviewed against two domain-specific criteria: (i) the presence of annotations for \textbf{Scleractinian} (reef-building hard coral) taxa, and (ii) use of \textbf{in situ} imagery from shallow, tropical reef environments, resulting in \textbf{100 sources}.

To ensure uniqueness, we eliminated duplicate annotations based on filename, label identity, and point-level annotation coordinates---occasionally leading to the exclusion of entire sources. We also removed any sources that had become private after our data crawl, yielding \textbf{88 public, verifiable sources}. Following consultation with data owners, we excluded three calibration sources used for training novice annotators, resulting in \textbf{85 sources}.

A final filtering stage retained only labels that met all of the following criteria: (i) ecologically relevant (i.e., hard coral taxa), (ii) appeared at least 100 times, (iii) were present in a minimum of three distinct sources with at least 10 annotations per source, (iv) exhibited consistent, visually distinguishable patterns, and (v) were taxonomically valid per WoRMS. After filtering and consolidation, the final \textbf{ReefNet} dataset consisted of \textbf{76 sources} and approximately \textbf{925{,}000 hard coral annotations}.

In parallel, we standardized associated metadata for each source, including geographic location, contributor affiliation, and source history (see Table~\ref{tab_contributors}). Public source metadata and taxonomy mapping files are listed in Appendix~\ref{sup_resources}.

\noindent\textbf{CoralNet Point Sampling Strategies}
Randomized point sampling is a long-established ecological standard and has also been widely adopted in terrestrial ecology (e.g., canopy cover, bird populations, vegetation structure). While ReefNet models treat each annotation point as a patch-level sample (crops from the same image may be dependent), the underlying point annotations originate from CoralNet-selected sources, which employ automated sampling methods across sites with:
\begin{itemize}
    \item \textbf{Simple random:} 51 sources
    \item \textbf{Stratified random:} 23 sources
    \item \textbf{Uniform grid:} 2 sources
\end{itemize}

Source sampling schemes apply to all benthic annotations, whereas the benchmark retains only points with an accepted hard-coral label. Per-image sampling density therefore differs from retained hard-coral point counts. We do not conflate these populations.

\subsection{ReefNet Data Quality Control}
\label{sup_quality_control}

To ensure taxonomic reliability and consistency across ReefNet annotations, we implemented a multi-stage expert verification and filtering process. The filtering process starts from 920,017 hard-coral annotations entering quality control, retains 860,463 after source/taxon filtering, and yields the 479,027-annotation benchmark after source--taxon filtering. These stage counts are separate from the full-release and mapped-label totals. This process was supported by a custom web-based application developed specifically for large-scale, structured review of coral genus annotations by marine biologists. Expert feedback collected through this platform was used to assess annotation quality, identify systematic labeling errors, and curate high-confidence subsets for benchmark construction.

\begin{figure}[!ht]
    \centering
    \includegraphics[width=\textwidth]{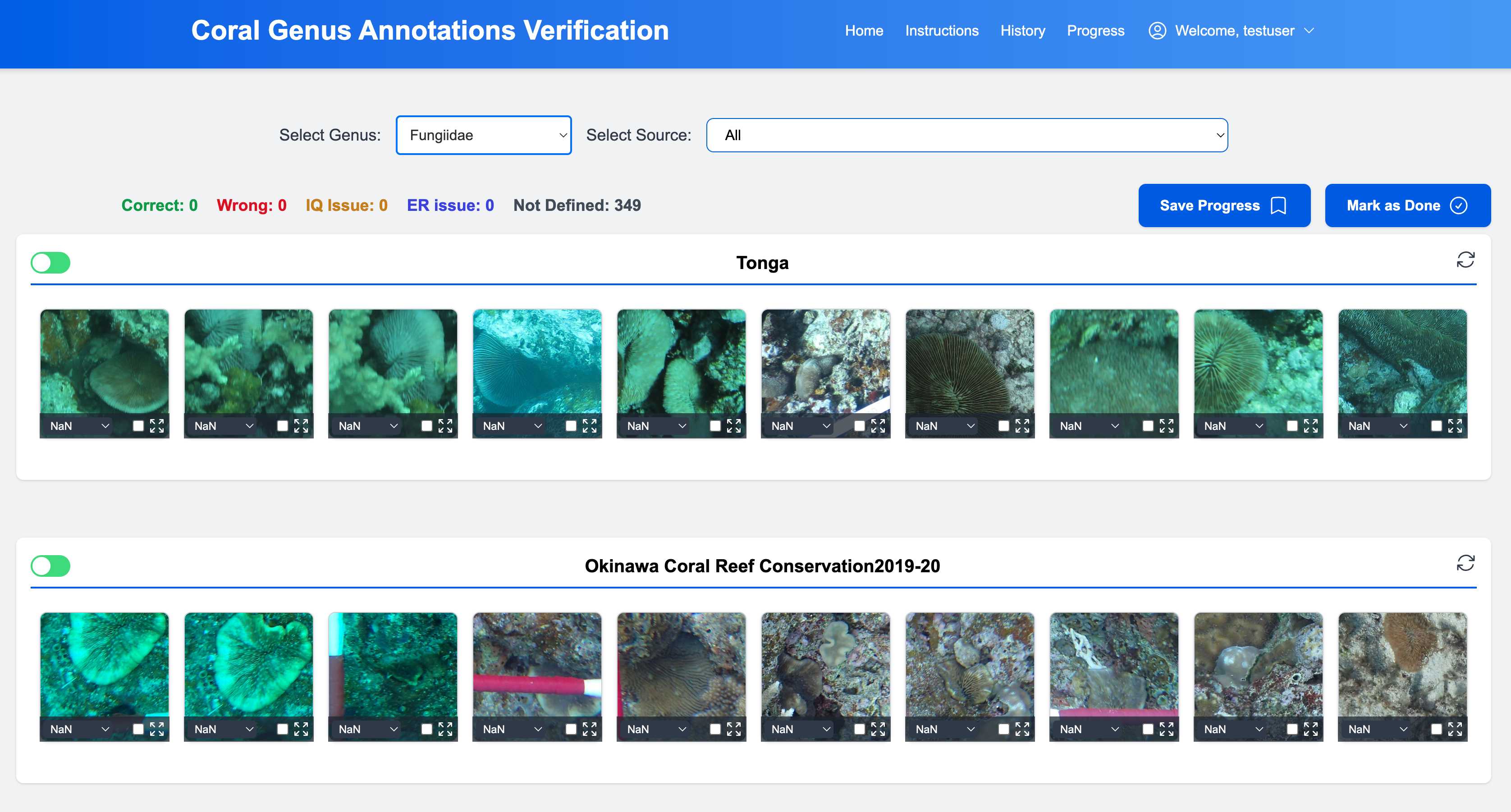}
    \caption{\textbf{Screenshot of the ReefNet manual verification platform.} Experts assess genus-level annotations by reviewing image patches alongside full-image context and assign structured labels through a dropdown interface. The tool supports efficient, large-scale quality control of coral annotations.}
    \label{fig:verification_tool}
\end{figure}

\paragraph{Manual Verification Tool.}
We developed a custom web-based platform to support expert review of coral genus annotations (Figure~\ref{fig:verification_tool}). Marine biologists used the tool to verify stratified subsets of annotations across sources and taxa. Each annotation was labeled as \textit{Correct}, \textit{Incorrect}, \textit{Low Quality Image}, or \textit{Hard to Decide}, based on a zoomable image patch together with the full-image context. The interface supports genus- and source-specific filtering, progress tracking, and pseudonymized user sessions. The system is lightweight, scalable, and integrated into the ReefNet curation pipeline. All verification actions are logged with metadata to support auditability and downstream filtering. Expert feedback informed exclusion criteria and helped construct the high-confidence benchmark subset, which achieved 92\% agreement. These labels were also used to refine taxonomic definitions and guide model retraining.

\subsection{More Details on the Centralized Re-Verification Protocol}
\label{sup_verification_protocol}

Our reviewers are not inherently superior to the original CoralNet annotators. However, our audit revealed systematic issues, including taxonomic inconsistencies across sources, frequent mislabeling (e.g., genus-level errors in \textit{Acropora}), and the absence of standardized label conventions. Moreover, because each CoralNet source was annotated by a distinct group with varying taxonomic expertise, labels that may be acceptable within individual sources can introduce substantial noise and bias when merged across datasets. ReefNet addresses this issue by applying centralized expert verification across all sources, thereby establishing a consistent cross-source quality baseline that CoralNet alone cannot provide.

\noindent\textbf{Expert Reviewers' Background.}
The verification team included one highly experienced coral taxonomist, three senior coral ecologists, and six PhD-level specialists in coral systematics and reef monitoring. Contributions to manual data verification are acknowledged in the main paper.

\noindent\textbf{Review Protocol.}
The stratified audit covered 8,962 patches, with approximately ten per source--taxon group. Reviewers categorized patches as Correct, Incorrect, Low Quality Image, or Hard to Decide. Agreement is the fraction categorized as Correct, including uncertain and low-quality cases in the denominator. Source/taxon filtering at 50\% audit agreement raised agreement on the remaining reviewed subset from 73\% to 78\%; retaining source--taxon pairs with at least 70\% agreement raised it to 92\%. This is agreement on a non-proportional audit subset, not an exhaustive error rate or lower bound for all retained annotations.

\noindent\textbf{Goals of Expert Filtering.}
The objective of our verification process is not to override CoralNet annotations, but to augment them with standardized, reproducible curation. Specifically, the process (i) standardizes labels through WoRMS AphiaIDs, (ii) resolves cross-source inconsistencies, and (iii) produces a high-confidence dataset suitable for benchmarking machine-learning models in ecological research.

\noindent\textbf{Scalability Considerations.}
Although the verification tool enables efficient, structured feedback from expert reviewers, its scalability remains constrained by the availability of qualified taxonomists. As ReefNet grows or is extended to more diverse regions, broader human-in-the-loop verification workflows will be needed to support higher annotation throughput. Future directions include active learning to prioritize ambiguous or low-confidence cases, hierarchical review pipelines with tiered expertise, and exploration of consensus-based crowdsourcing for preliminary filtering stages. Addressing these limitations will be critical for sustaining high-quality annotation pipelines in large-scale ecological monitoring.

\subsection{Additional Details on Covariate Diversity}
\label{sup_covariate_diversity}
\begin{figure}[!ht]
    \centering
    \includegraphics[width=\textwidth]{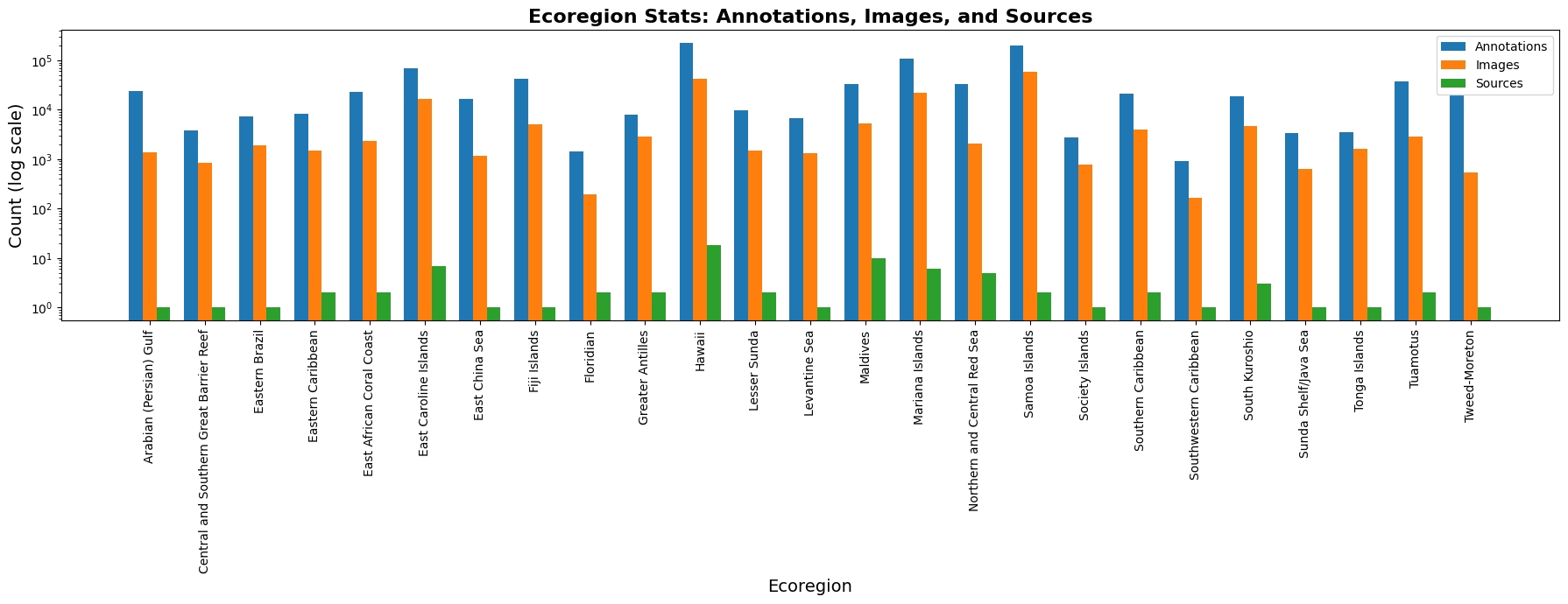}
    \caption{
    \textbf{Distribution of annotations, images, and sources across marine ecoregions.}
    Each bar group represents one of 26 marine ecoregions covered in the ReefNet dataset, as defined by the MEOW classification. The plot shows the number of annotations, distinct images, and image sources per ecoregion (log scale). Geographic coverage is notably high in regions like Hawaii, Samoa, and the Mariana Islands, but more limited in some Atlantic and Western Indian Ocean regions.
    }
    \label{fig:ecoregion_distribution}
\end{figure}

This section highlights three key aspects of metadata in ReefNet that are essential for both machine learning and biological applications: i) geographic location (Figure.~\ref{fig:ecoregion_distribution}), ii) image white balance, and iii) image resolution.

\noindent\textbf{Ecoregions Distribution.} To assess the geographic and ecological diversity of ReefNet, we grouped all annotations by marine ecoregion using the Marine Ecoregions of the World (MEOW) classification system. Figure~\ref{fig:ecoregion_distribution} summarizes the distribution of sources, images, and annotations across 26 unique ecoregions.

The dataset spans a wide range of coral reef habitats, including the Red Sea, Caribbean, Central Indo-Pacific, and Central Pacific. Notably, several biodiversity hotspots are heavily represented: the Hawaiian Islands (18 sources, 221,419 annotations), Samoa Islands (2 sources, 201,685 annotations), and Mariana Islands (6 sources, 107,553 annotations). Other regions, such as the East Caroline Islands and Fiji, also contribute substantial volumes of data, enriching the taxonomic and ecological breadth of the dataset.

The number of image sources per ecoregion varies significantly, reflecting uneven global monitoring efforts. While some ecoregions are represented by multiple contributors and thousands of images, others---such as Tweed--Moreton or the Southwestern Caribbean---appear undersampled. This variation is important for evaluating model robustness and generalization across biogeographically distinct environments.

All counts are shown on a log scale to accommodate differences spanning several orders of magnitude. Overall, this distribution highlights ReefNet’s potential to support geographically robust coral classification and cross-region generalization studies.

\begin{figure}[p]
    \centering
    \includegraphics[width=\textwidth]{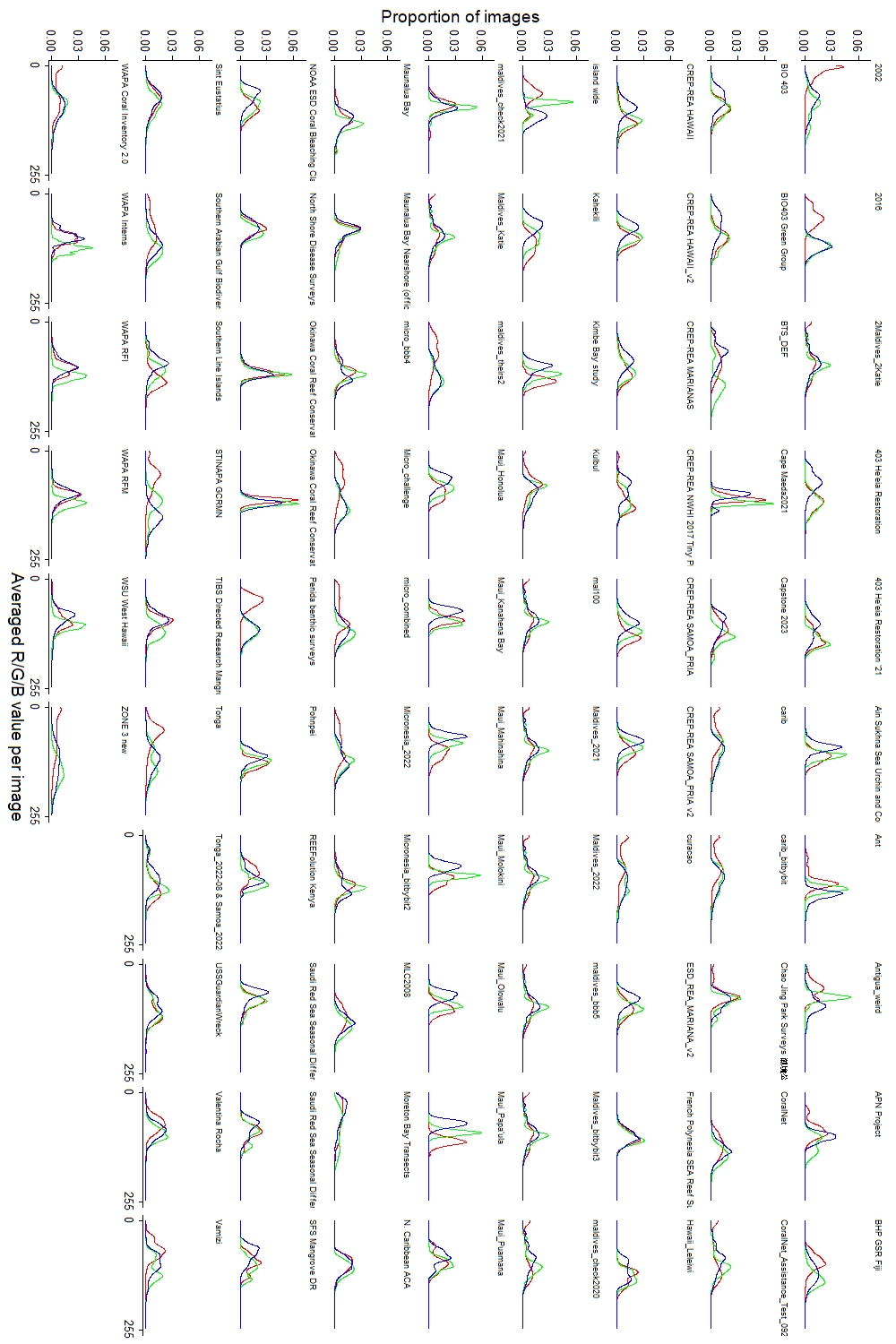}
    \caption{Density plots per CoralNet source of the average Red, Green, and Blue values of each image.}
    \label{fig_RGB}
\end{figure}

\noindent\textbf{Image White Balance.} Variations in water clarity, depth, light penetration, and camera configuration can lead to strong color casts in underwater imagery. We therefore document the average intensities of the red, green, and blue color channels across CoralNet sources (Figure~\ref{fig_RGB}). Sources with non-overlapping peaks in their RGB distributions generally exhibit strong color casts due to poor white balancing. The variety of color casts captured in ReefNet exposes models to heterogeneous imaging conditions; certain applications may benefit from using imagery whose lighting conditions more closely match the target deployment setting.

\noindent\textbf{Image Resolution.} Image resolution in ReefNet ranges from 0.2 to 27 megapixels, with a median resolution of 12 megapixels. This diversity permits analysis of resolution variation, but does not itself establish model robustness. As with white balance, however, users may prefer to extract images with a more consistent resolution for specific tasks. Most sources in ReefNet exhibit relatively uniform resolution distributions as a result of standard camera systems, whereas a few sources show substantial variation, generally because benthic quadrats were manually cropped from the original images (Figure~\ref{fig_resolution}).

To facilitate the creation of customized sub-datasets, source-level metadata---including geographic location, RGB color profiles, and resolution statistics---will be released with the ReefNet dataset on Hugging Face.

\subsection{Al-Wajh Lagoon Dataset}
\label{sup_thealwajh_testset}
\begin{figure}[!ht]
    \centering
    \includegraphics[width=\textwidth]{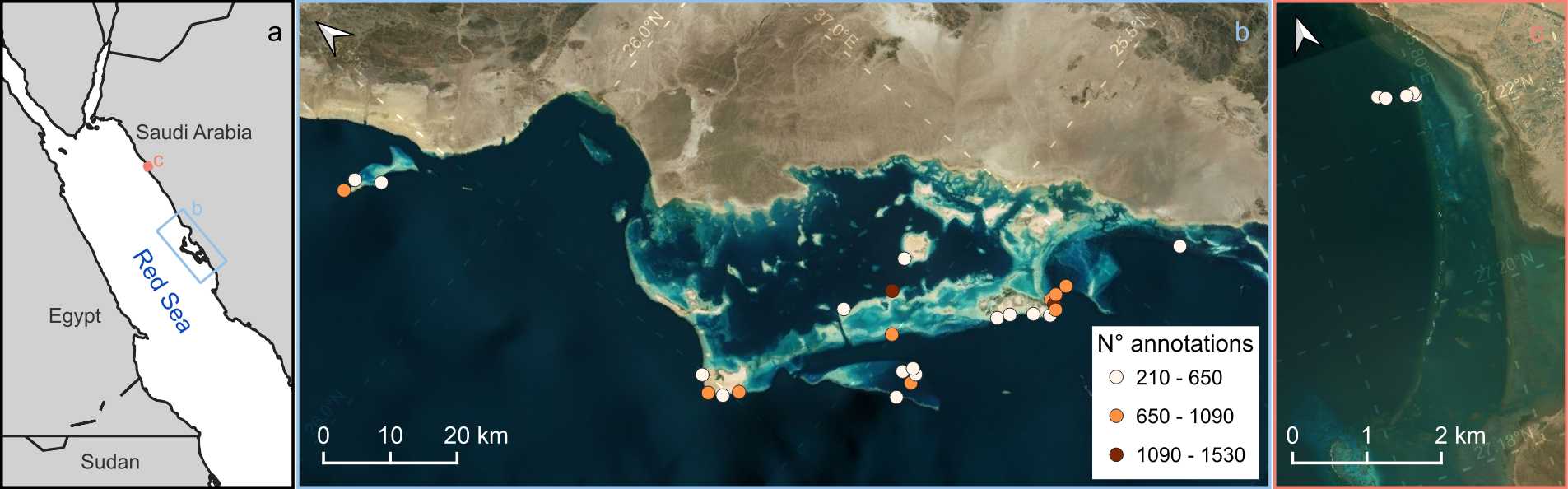}
    \caption{\textbf{Map showing the origin of the imagery in the Al-Wajh test dataset.} Panel a shows the location of panel b and c in the Red Sea. Panel b shows the Al-Wajh lagoon with the specific locations of the annotations in the test set. Panel c shows the location of a limited set of images taken outside the Al-Wajh lagoon. Map contains Natural Earth data and Bing satellite imagery.}
    \label{Map_RSG_dataset}
\end{figure}

\begin{figure}[p]
    \centering
    \includegraphics[width=\textwidth]{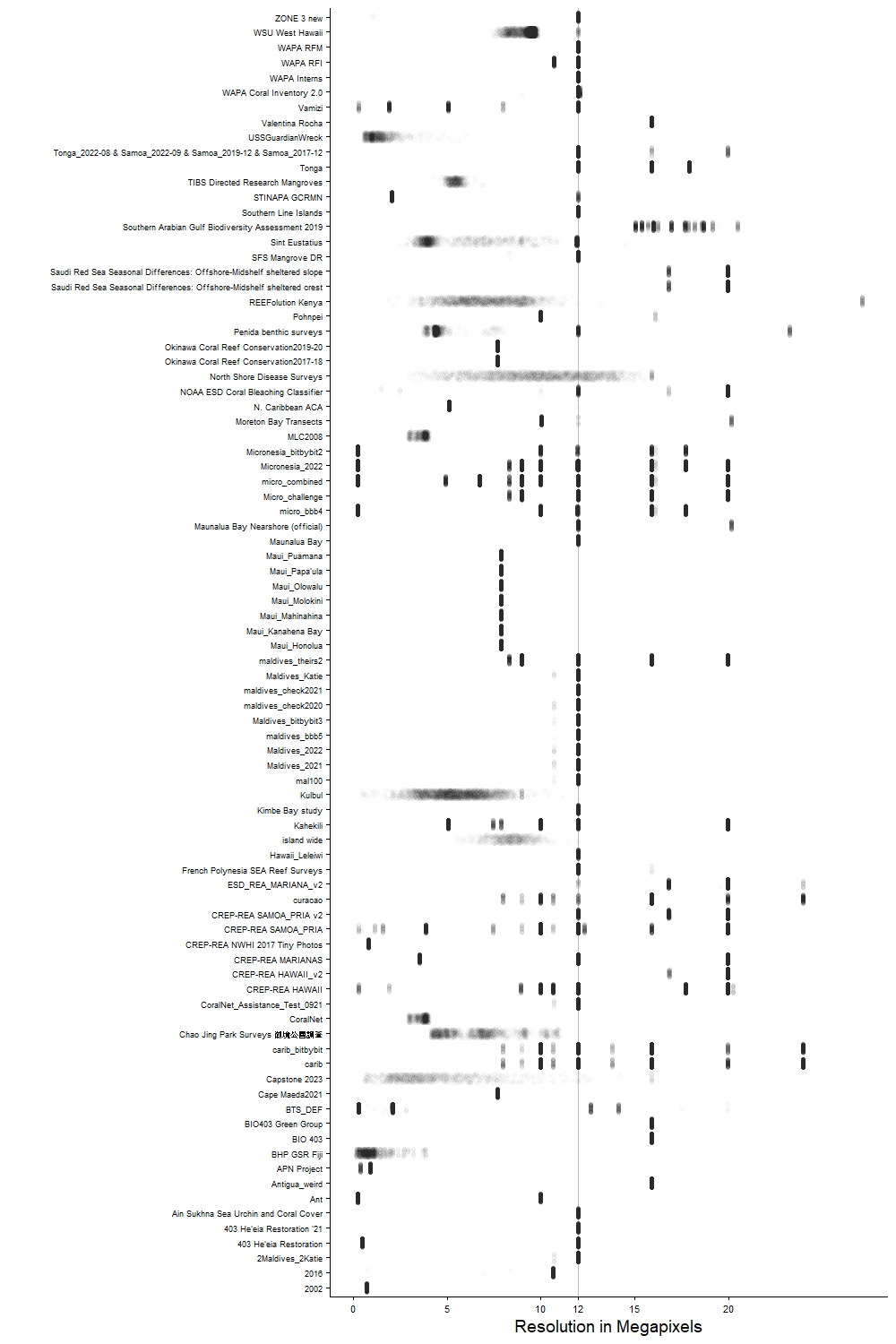}
    \caption{Scatter plot showing the resolution of each annotated image per source.}
    \label{fig_resolution}
\end{figure}

\noindent{\textbf{Dataset Overview.}}
As part of a larger environmental monitoring effort, nearly 300 coral reefs were surveyed in the Al-Wajh lagoon (25.6°N, 36.8°E) between March and September 2021. Surveyed reefs were strategically selected using a stratified random design to ensure representation of the region’s diverse reef habitats, including Reef Walls, Reef Crests, Reef Slopes, Patch Reefs, and Algal Reefs. The Al-Wajh Test dataset is composed of a subset of this imagery collected from the outer reefs of the lagoon. Imagery in the Al-Wajh Test dataset was collected using purpose-built photoquadrats, ensuring images were taken from a consistent distance from the seafloor. The camera was Canon PowerShot G9 X Mark II which shoots images of 20.2 MP. This represents the upper end of the resolutions available in the ReefNet dataset (Figure~\ref{fig_resolution}). A total of \textbf{1,376 images} were manually annotated on the CoralNet platform. Per image, 10 points were generated following a stratified random design of 2 rows and five columns, giving 13,760 sampled benthic points, of which \textbf{4,624} are retained hard-coral annotations. The geographic coverage and annotation locations of this benchmark subset are shown in Figure~\ref{Map_RSG_dataset}.

\begin{figure}[!ht]
\centering
\includegraphics[width=0.9\linewidth]{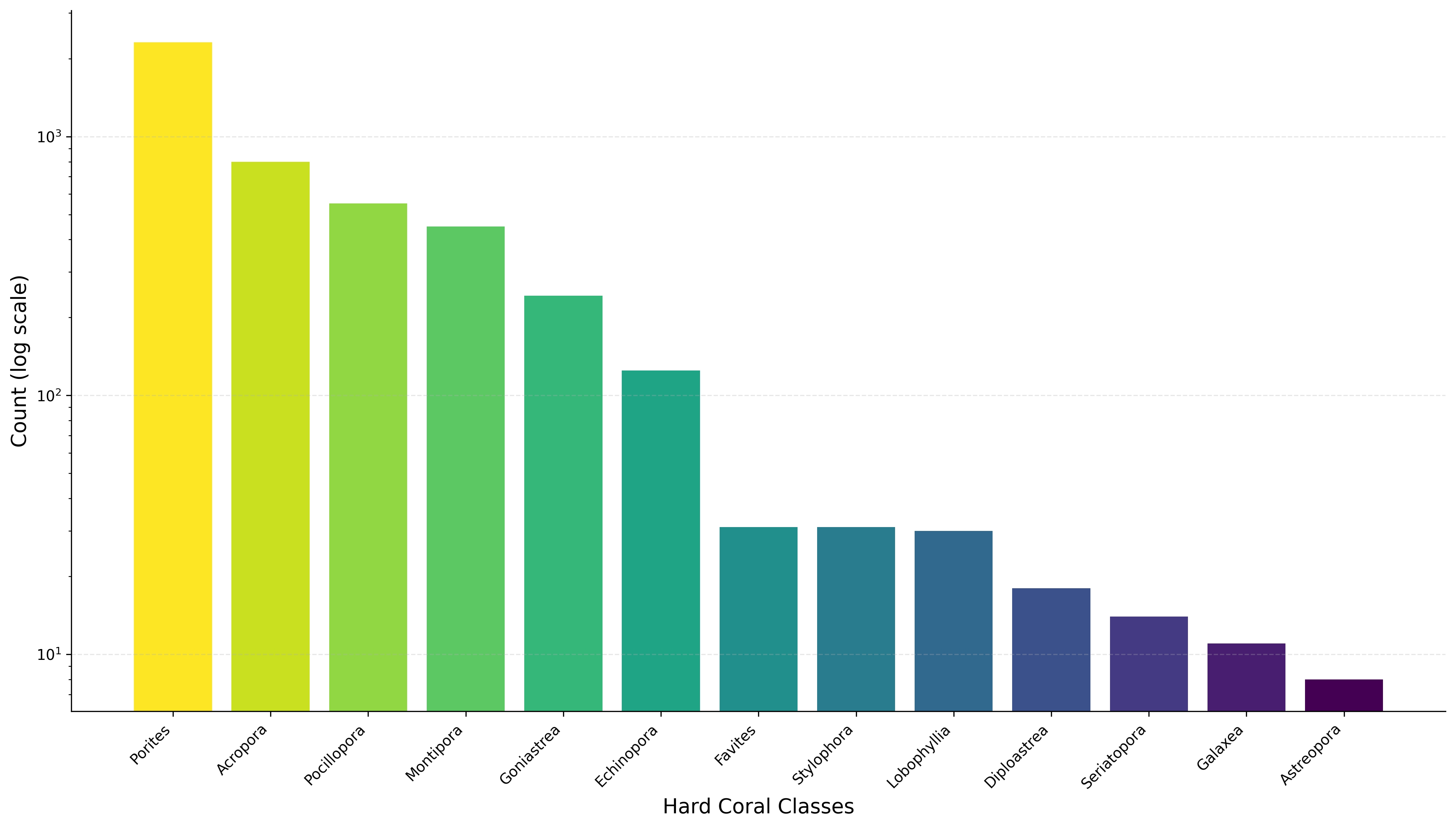}
\caption{\textbf{Log-scale Distribution of Annotations Across 13 Hard Coral Genera in the Al-Wajh Lagoon Dataset.} The
distribution shows a bias toward some classes, comparable to the distribution in ReefNet.}
\label{fig_alwajh_dist}
\end{figure}

\noindent\textbf{Annotation Distribution.}
The Al-Wajh Lagoon dataset covers 13 coral genera that overlap with those used to train the different models. Figure~\ref{fig_alwajh_dist} shows the annotation distribution, which exhibits a bias toward more abundant classes such as \textit{Porites}, \textit{Acropora}, \textit{Pocillopora}, and \textit{Montipora}.

\subsection{Point-Centered Crops: Composition and Overlap}
\label{sup_crop_analysis}
Each sample is a $1024\times1024$ crop centered on a labeled point; the target is the organism under that point. Other labeled taxa may occur in the surrounding image region. Table~\ref{tab:crop_composition} reports the number of distinct observed labels per benchmark crop. Of the 479,027 crops, 29,713 (6.20\%) contain at least two observed labels. Corresponding counts are 1,469/23,043 (6.38\%) on ReefNet test and 129/4,624 (2.79\%) on Al-Wajh. These are lower bounds because point annotations do not exhaustively identify every organism in a crop. Multiple observed taxa do not, by themselves, imply an incorrect center-point label.

\begin{table}[ht]
\centering
\begin{tabular}{lrr}
\toprule
Observed labels per crop & Crops & Fraction (\%)\\
\midrule
One & 449,314 & 93.80\\
Two & 26,991 & 5.63\\
Three & 2,431 & 0.51\\
Four or more & 291 & 0.06\\
\bottomrule
\end{tabular}
\caption{Label composition of all 479,027 benchmark crops, measured using observed point annotations.}
\label{tab:crop_composition}
\end{table}

For each unordered pair of crops from the same original image, overlap is the shared valid source-image area divided by the nominal crop area $1024^2$. Padding is excluded from the shared area. This measure differs from intersection over union (IoU). Table~\ref{tab:crop_overlap} reports all-pair and positive-overlap averages separately. All crops from an original image remain in one split; overlapping crops create within-image dependence but do not cross that image's split boundary. The source curation also removes duplicate annotations by filename, label, and point coordinates; this is not an exhaustive visual duplicate-image audit across sources.

\begin{table}[ht]
\centering\small
\begin{tabular}{lrrrr}
\toprule
Set & Pairs & Any overlap (\%) & \multicolumn{2}{c}{Mean overlap (\%)}\\
\cmidrule(lr){4-5}
& & & All pairs & Overlapping pairs\\
\midrule
ReefNet benchmark & 2,891,942 & 43.75 & 13.77 & 31.47\\
ReefNet test & 111,895 & 42.83 & 13.70 & 31.98\\
Al-Wajh & 8,473 & 23.57 & 7.39 & 31.34\\
\bottomrule
\end{tabular}
\caption{Spatial overlap among same-image crop pairs. The ReefNet benchmark row includes training, validation, and test crops.}
\label{tab:crop_overlap}
\end{table}
         %

\section{Experimental details}
\label{sec:exp_detail}
\subsection{Training Setup}

\noindent\textbf{General Settings.}
Unless otherwise specified, all vision-only models and VLMs were trained for \textbf{20 epochs} using the AdamW optimizer with a base learning rate of $2 \times 10^{-5}$ and a batch size of 16 per GPU across 3 A100 GPUs. Unless overridden in Table~\ref{tab:foundation_preprocessing}, inputs were normalized using ImageNet statistics: mean $[0.485, 0.456, 0.406]$ and standard deviation $[0.229, 0.224, 0.225]$. The original reference runs used no data augmentation; the four added visual backbones use the random-resized training crops specified in Table~\ref{tab:foundation_preprocessing}. The separate ViT augmentation ablation is described in Appendix~\ref{sup_training_analysis}. Training was conducted with automatic mixed precision (BF16) and cosine learning rate scheduling with 5 warmup epochs. Each training sample consists of a $1024 \times 1024$ crop extracted on-the-fly from the original full-resolution CoralNet image, centered on each point annotation. A held-out validation set was used to monitor performance and select the best-performing model checkpoint. No validation samples were included in training, ensuring a clean separation between training and evaluation.

\noindent\textbf{Architectures.}
The original seven vision-only models were: ViT-B/16, DeiT-B/16, Swin-B, ResNet-50, ConvNeXt-B, EfficientNet-B3, and BEiT-B/16. These seven models used pretrained weights from ImageNet-21K or ImageNet-1K (as publicly available via \texttt{timm}). The extracted point-centered crop is $1024\times1024$; model input preprocessing is distinct from crop construction.
We additionally evaluate DINOv2 ViT-L/14~\cite{dinov2}, DINOv3 ViT-L/16~\cite{dinov3}, SigLIP 2 ViT-L/16~\cite{siglip2}, and public MAE ViT-L/16~\cite{he2022masked}, each with a 39-label classifier under the 20-epoch, validation-selected protocol. SigLIP 2 uses only its visual encoder in this comparison.
Their pretrained initializations and model-specific preprocessing are listed in Table~\ref{tab:foundation_preprocessing}.
 We also evaluated five VLMs: CLIP~\cite{clip}, OpenCLIP~\cite{openclip}, SigLIP~\cite{siglip}, BioCLIP~\cite{BioCLIP}, and BioCLIP2~\cite{bioclip2}. Among these, CLIP, OpenCLIP, and BioCLIP2 were additionally fine-tuned under full and few-shot settings; SigLIP and BioCLIP were evaluated zero-shot only. For MLLMs, five models were evaluated zero-shot: Qwen3-VL-8B~\cite{qwen3vl}, InternVL3.5-8B~\cite{intern35}, MiniCPM-V-4.5~\cite{minicpm}, LLaVA-NeXT~\cite{llavanext}, and LLaVA-OneVision~\cite{llavaone}. All five open MLLMs were additionally fine-tuned using LoRA; few-shot adaptation covers only Qwen3-VL, MiniCPM-V, and InternVL-3.5.

\noindent\textbf{VLM Fine-Tuning Setup.}
CLIP, OpenCLIP, and BioCLIP2 were fine-tuned for 20 epochs using AdamW with learning rate $2 \times 10^{-5}$, cosine scheduling, and no augmentation. For each retained label, we constructed a taxonomy string by concatenating all hierarchical levels through the retained taxonomic rank based on WoRMS (e.g., \texttt{"Animalia;\allowbreak{}Cnidaria;\allowbreak{}Hexacorallia;\allowbreak{}Scleractinia;\allowbreak{}Acroporidae;\allowbreak{}Acropora"}), and used these strings as text queries. BioCLIP2 was initialized from pretrained weights on TreeOfLife-200M~\cite{treeoflife_200m}. Training used local and global contrastive loss objectives as in the original BioCLIP setup.

\begin{table}[htbp]
\centering
\small
\setlength{\tabcolsep}{4pt}
\renewcommand{\arraystretch}{1.15}
\begin{tabular}{@{}>{\raggedright\arraybackslash}p{1.6cm}>{\raggedright\arraybackslash}p{0.9cm}>{\raggedright\arraybackslash}p{2.7cm}>{\raggedright\arraybackslash}p{4.0cm}>{\raggedright\arraybackslash}p{2.1cm}@{}}
\toprule
Model & Input & Fine-tuning crop & Validation/test\newline preprocessing & Normalization\\
\midrule
DINOv2 ViT-L/14 & $224^2$ & Bicubic random-resized crop to $224^2$ & Resize shorter edge to 224; center crop $224^2$ & ReefNet mean/std\\
\multicolumn{5}{@{}l@{}}{\footnotesize Initialization: \texttt{vit\_large\_patch14\_dinov2.lvd142m}}\\[3pt]
DINOv3 ViT-L/16 & $256^2$ & Bicubic random-resized crop to $256^2$ & Resize shorter edge to 256; center crop $256^2$ & ImageNet mean/std\\
\multicolumn{5}{@{}l@{}}{\footnotesize Initialization: \texttt{vit\_large\_patch16\_dinov3\_qkvb.lvd1689m}}\\[3pt]
SigLIP 2 ViT-L/16 & $256^2$ & Bicubic random-resized crop to $256^2$ & Resize shorter edge to 284; center crop $256^2$ (\texttt{crop\_pct=0.9}) & Mean and std both $(0.5,0.5,0.5)$\\
\multicolumn{5}{@{}l@{}}{\footnotesize Initialization: \texttt{vit\_large\_patch16\_siglip\_256.v2\_webli}}\\[3pt]
MAE ViT-L/16 & $224^2$ & Bicubic random-resized crop to $224^2$ & Resize shorter edge to 256; center crop $224^2$ (\texttt{crop\_pct=0.875}) & ReefNet mean/std\\
\multicolumn{5}{@{}l@{}}{\footnotesize Initialization: official \texttt{mae\_pretrain\_vit\_large.pth}}\\
\bottomrule
\end{tabular}
\caption{Pretrained initialization and preprocessing for the four added visual backbones. Input dimensions and resize lengths are in pixels. Each model uses a 39-label classifier, 20 fine-tuning epochs, and checkpoint selection on ReefNet validation. ImageNet normalization uses the mean and standard deviation listed in General Settings; ReefNet denotes the dataset-specific mean/std used for those runs. SigLIP 2 uses its visual encoder, and MAE uses the public pretrained checkpoint.}
\label{tab:foundation_preprocessing}
\end{table}

\subsection{MLLM Fine-Tuning Setup}
\label{sup_mllm_finetuning}
All five open MLLMs were adapted using LoRA via the LLaMA-Factory framework. We used LoRA rank 8 applied to all linear layers (\texttt{lora\_target: all}), training for \textbf{2 epochs} with AdamW (learning rate $5 \times 10^{-5}$, cosine schedule, warmup ratio 0.03). Training used BF16 precision with a per-device batch size of 8 and no gradient accumulation. Images were resized to a maximum of $1024 \times 1024$ pixels. Flash Attention 2 was enabled where supported. The fine-tuning prompt was: \texttt{"What is the genus in the center of the image? Answer with one word."} with the genus label as the target response. 
 The five MLLMs fine-tuned were Qwen3-VL-8B~\cite{qwen3vl}, InternVL3.5-8B~\cite{intern35}, MiniCPM-V-4.5~\cite{minicpm}, LLaVA-NeXT~\cite{llavanext}, and LLaVA-OneVision~\cite{llavaone}. The original three-model adaptation subset was selected using zero-shot performance and available compute; full-data fine-tuning is now reported for all five, while the few-shot subset is unchanged.

\subsection{Zero-Shot Prompting Protocol}
\label{sup_zeroshot_prompting}
\noindent\textbf{Vision--Language Models (VLMs):} For all VLM-based models (CLIP, OpenCLIP, SigLIP, BioCLIP, and BioCLIP2), we use a simple zero-shot text template: “A photo of a \{genus\}.”

For each of the 39 candidate coral labels, we instantiate the template with the genus name and compute the image–text similarity score. The predicted genus corresponds to the text prompt with the highest similarity score.

\noindent\textbf{Multimodal Large Language Models (MLLMs):} For MLLMs (Qwen3-VL, InternVL3.5, MiniCPM-V, LLaVA-NeXT, and LLaVA-OneVision), we use the following prompt structure.

\begin{tcolorbox}[colback=blue!5!white,colframe=blue!50!black]
You are an expert marine biologist specializing in hard coral genera recognition.\\
Task: \\
You are given: \\
1. An image of a coral colony. \\
2. A list of 39 possible coral genera. \\
Your job is to identify which genus the coral in the image belongs to. \\
Rules: \\
- You must select exactly ONE genus. \\
- The answer MUST be one of the provided genera. \\
- Do NOT provide any explanation. \\
- Do NOT describe the coral. \\
- Output ONLY the genus name. \\
Genus List: \\
{genus\_list}
\end{tcolorbox}

The quoted prompt preserves the experimental wording ``genus'', although Fungiidae is a family-level candidate. The placeholder \texttt{\{genus\_list\}} contains the 39 candidate label names (38 genera plus Fungiidae), which are explicitly included in the prompt.

\noindent\textbf{Output Parsing}: We parse the model output using string matching against the provided list of genus names. If the model output contains additional tokens, we extract the substring corresponding to the closest matching genus name from the candidate list.

\subsection{Claude/GPT API Evaluation}
\label{sup_api_models}
\paragraph{Model identifiers and access.}
Claude Opus 5 (\texttt{anthropic/claude-opus-5}) was queried on 27 July 2026 and GPT-5.4 (\texttt{openai/gpt-5.4}) on 26--27 July 2026 through the OpenRouter chat-completions API, using default provider routing. No provider or immutable model snapshot was pinned or recorded; the identifiers and dates describe the recorded calls, not an immutable model version.

\paragraph{Inputs and candidate sets.}
Each request contained one user message, with prompt text followed by the image and no system message. All 23,043 ReefNet and 4,624 Al-Wajh test crops were evaluated. Original $1024\times1024$ crops were sent as stored, base64-encoded (JPEG for ReefNet, PNG for Al-Wajh), without client-side resizing, cropping, or re-encoding. The image-detail option was unset; provider-side downscaling was not recorded.

The saved prompts list 39 candidates for ReefNet and only 13 for Al-Wajh in every API condition, including reasoning and descriptions. The Al-Wajh candidates are Acropora, Astreopora, Diploastrea, Echinopora, Favites, Galaxea, Goniastrea, Lobophyllia, Montipora, Pocillopora, Porites, Seriatopora, and Stylophora. This narrower prompt candidate set differs from the standard 39-label protocol; it does not imply a hard output mask, since models can produce off-list answers. Macro recall averages over the 39 and 13 ground-truth labels, respectively. Cross-model Al-Wajh comparisons involving these API runs are therefore not candidate-set matched.

\paragraph{Decoding and reasoning.}
All calls used temperature 0, top-$p$ 1, and seed 0. Direct classification explicitly disabled internal reasoning (\texttt{reasoning: \{enabled: false\}}), requested plain text, and capped output at 64 tokens. GPT reasoning conditions requested JSON containing \texttt{reasoning} and \texttt{genus}; provider reasoning settings were left at their defaults, with no explicit effort level or reasoning-token budget. The output cap was 1,024 tokens for name-only reasoning and 2,048 for reasoning with descriptions. Claude was evaluated only in direct, name-only mode. GPT used direct and reasoning name-only conditions on both tests, direct descriptions on both tests, and reasoning with descriptions only on Al-Wajh.

\paragraph{Parsing, scoring, and retries.}
Direct predictions are the response after trimming surrounding whitespace. Reasoning predictions use the JSON \texttt{genus} field; if JSON parsing fails, the parser tries the outermost braced substring, then a regular-expression extraction of that field, and otherwise returns an empty prediction. Scoring uses case-sensitive exact string equality with the ground-truth label, without fuzzy matching or label remapping. Invalid predictions, including empty, truncated, off-list, and refused answers, count as errors and remain in the denominator. This API parser is separate from the open-model parser described above.

Transient API failures were retried up to five times with exponential backoff; remaining failed requests were re-sent until all images had a response. The Claude Al-Wajh run required a second pass for 335 initially failed requests. No final scored file contained an API error. Table~\ref{tab:api_invalid} records the remaining invalid predictions; these counts come from the saved evaluation audit. Direct runs had no empty predictions, while each GPT name-only reasoning run had one empty prediction after a truncated JSON response. The seven off-list GPT Al-Wajh direct answers were \textit{Platygyra}, a ReefNet label absent from the 13 candidates.

\begin{table}[htbp]
\centering\small
\begin{tabular}{lllrr}
\toprule
Model & Test & Condition & Empty & Off-list\\
\midrule
Claude Opus 5 & ReefNet & Direct, names & 0 & 2\\
Claude Opus 5 & Al-Wajh & Direct, names & 0 & 0\\
GPT-5.4 & ReefNet & Direct, names & 0 & 6\\
GPT-5.4 & ReefNet & Reasoning, names & 1 & 0\\
GPT-5.4 & ReefNet & Direct, descriptions & 0 & 4\\
GPT-5.4 & Al-Wajh & Direct, names & 0 & 7\\
GPT-5.4 & Al-Wajh & Reasoning, names & 1 & 0\\
GPT-5.4 & Al-Wajh & Direct, descriptions & 0 & 0\\
GPT-5.4 & Al-Wajh & Reasoning, descriptions & 0 & 0\\
\bottomrule
\end{tabular}
\caption{Invalid API predictions retained as errors. Off-list counts exclude empty predictions. Each ReefNet run has 23,043 samples; each Al-Wajh run has 4,624.}
\label{tab:api_invalid}
\end{table}

\paragraph{Exact name-only prompt wording.}
The following direct template reproduces the saved ReefNet API prompt, replacing only its final label list with \texttt{\{genus\_list\}}. The placeholder expands to the 39 labels in Table~\ref{tab:api_descriptions}, in alphabetical order and separated by newlines. Al-Wajh replaces ``39'' with ``13'' and uses the 13-label list above in the same format. Claude and GPT used the same direct prompt. Historical use of ``genus'' is preserved although Fungiidae is a family.
\begin{Verbatim}
You are a marine biologist specializing in coral taxonomy.
Task:
You are given:
1. An image of a coral colony.
2. A list of 39 possible coral genera.

Your job is to identify which genus the coral in the image belongs to.
Rules:
- You must select exactly ONE genus.
- The answer MUST be one of the provided genera.
- Do NOT provide any explanation.
- Do NOT describe the coral.
- Output ONLY the genus name.

Genus List:
{genus_list}
\end{Verbatim}

The GPT name-only reasoning template uses the same dataset-specific candidate lists and count substitution:
\begin{Verbatim}
You are a marine biologist specializing in coral taxonomy.
Task:
You are given:
1. An image of a coral colony.
2. A list of 39 possible coral genera.

Your job is to identify which genus the coral in the image belongs to.
Rules:
- First reason about the visually diagnostic features of the coral in the center
  of the image: corallite size, shape and arrangement, septa and costae, wall
  structure, colony growth form, surface texture, and colour.
- Then select exactly ONE genus.
- The answer MUST be one of the provided genera.
- Respond with ONLY a single JSON object, with no markdown fences and no text
  before or after it, in exactly this form:
  {"reasoning": "<your concise reasoning>", "genus": "<one genus from the list>"}
- The "genus" value must be the genus name alone, spelled exactly as it appears
  in the list below.

Genus List:
{genus_list}
\end{Verbatim}

\subsection{Training Variability, Augmentation, and Convergence}
\label{sup_training_analysis}
Table~\ref{tab:training_seeds} reports three independent runs for full-data ViT-B/16 and BioCLIP2, and a separate class-capped Qwen3-VL study. Within each study, runs use the same protocol, splits, and checkpoint-selection rule. For ViT-B/16 and BioCLIP2, the first run is the reference result in the main full-fine-tuning table. Standard deviations use the sample denominator $n-1$ and are measured in percentage points; these summarize training-run variability, not test-set sampling uncertainty. Main-table scores remain single-run point estimates, with available seed studies reported here. Close single-run rankings do not establish statistically significant differences.

\begin{table}[ht]
\centering\small
\begin{tabular}{lllrrrr}
\toprule
Model & Training & Test & Run 1 & Run 2 & Run 3 & Mean $\pm$ SD\\
\midrule
ViT-B/16 & Full & ReefNet & 79.84 & 79.90 & 79.01 & $79.58\pm0.50$\\
ViT-B/16 & Full & Al-Wajh & 54.84 & 55.37 & 55.62 & $55.28\pm0.40$\\
BioCLIP2 & Full & ReefNet & 79.23 & 78.96 & 78.39 & $78.86\pm0.43$\\
BioCLIP2 & Full & Al-Wajh & 58.28 & 57.83 & 57.80 & $57.97\pm0.27$\\
\midrule
Qwen3-VL & Capped & ReefNet & 50.96 & 50.45 & 50.16 & $50.52\pm0.41$\\
Qwen3-VL & Capped & Al-Wajh & 34.72 & 34.68 & 32.66 & $34.02\pm1.18$\\
\bottomrule
\end{tabular}
\caption{Macro recall (\%) across three independent seeds per study, with mean and sample standard deviation. Run indices do not denote numeric seed values. Qwen3-VL uses a reduced, class-capped training set and is not a full-data comparison.}
\label{tab:training_seeds}
\end{table}

\paragraph{Scope of the capped MLLM study.}
Qwen3-VL uses at most 300 training patches per label: 11,072 of the 445,985 training samples and 692 optimizer updates, compared with 27,876 in the full-data run. This cap also changes class balance. Its $50.52\pm0.41$ ReefNet and $34.02\pm1.18$ Al-Wajh results characterize only this capped protocol; neither the means nor their variability should be presented as full-data Qwen3-VL results. In particular, they do not replace the full-data single-run scores of 72.99\% and 59.80\%. Full-data multi-seed results are available only for ViT-B/16 and BioCLIP2; training-run variability for the other full-data models is not quantified.

\paragraph{Augmentation ablation.}
The original ViT reference runs use no augmentation; this is separate from the random-resized crops used by the four added backbones in Table~\ref{tab:foundation_preprocessing}. A separate single-seed ViT-B/16 run with horizontal flipping, color jitter, RandAugment magnitude 7, and random erasing probability 0.1 achieves 80.90\% macro recall on ReefNet and 55.47\% on Al-Wajh. The corresponding three-seed no-augmentation reference is $79.58\pm0.50$ and $55.28\pm0.40$. This limited comparison suggests potential within-source benefit, but a single augmented run cannot establish significance, characterize augmentation variance, or determine effects for other architectures. Zero-shot evaluations are unaffected by training augmentation.

\paragraph{Available convergence evidence.}
Recorded validation loss decreases from 0.0854 to 0.0443 for Qwen3-VL and from 0.1087 to 0.0606 for InternVL-3.5 over the reported training trajectory. The first epoch accounts for 90.4\% and 86.2\% of the respective total reductions. Validation loss was reported to decrease at every evaluation without an upturn; over the final tenth of training, the change in mean training loss was smaller than its within-window standard deviation. Reported second-epoch macro-recall gains were 3.97, 2.04, and 3.87 percentage points for Qwen3-VL, MiniCPM-V, and InternVL-3.5, respectively. These summaries support diminishing validation-loss improvement but do not prove global convergence or replace complete learning curves. An additional half-epoch Qwen3-VL checkpoint achieves 72.75\%/56.94\% (ReefNet/Al-Wajh), compared with 72.99\%/59.80\% for the two-epoch result. Its restarted learning-rate schedule and smaller effective batch make this a descriptive comparison rather than a matched epoch-budget ablation; it is not used to tune on Al-Wajh.

The reported two-epoch MLLM run required 27,876 optimizer updates and more than 24 hours on four GPUs. Two-epoch MLLM LoRA and 20-epoch encoder fine-tuning are not compute-matched: the aim is a standardized protocol within each family, not a compute-normalized ranking across families. Epoch counts alone cannot normalize trainable parameters, optimizer updates, accelerator use, or wall time across these families.
 %

\section{Additional Results}
\label{sec:more_results}

\subsection{Head, Body, and Tail Breakdown}
\label{sup_head_body_tail}

\begin{figure}[t]
    \centering
    \includegraphics[width=\textwidth]{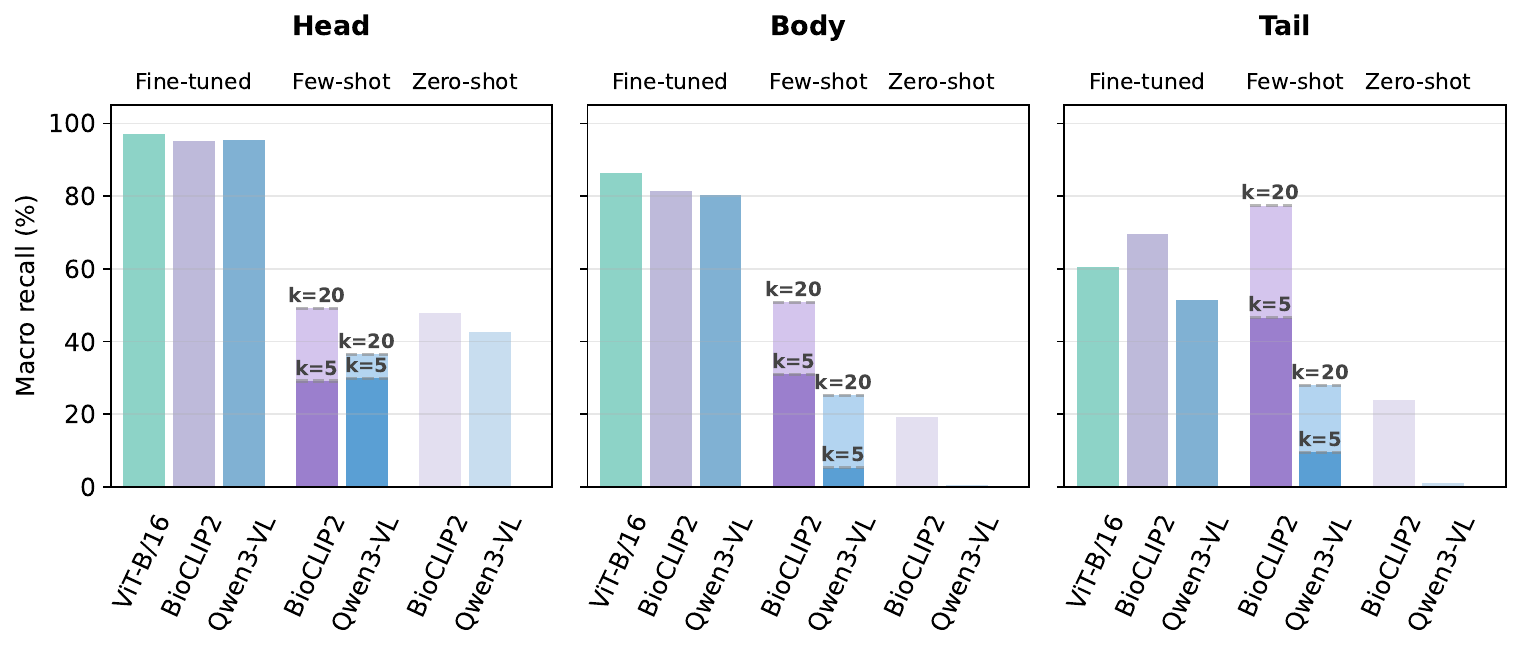}
    \caption{\textbf{Head, body, and tail macro recall.} Selected original-suite models on the ReefNet within-source test set (39 labels, 23,043 samples). Each panel separates full fine-tuning, few-shot adaptation, and zero-shot recognition. The lower few-shot segment shows $k=5$ recall and the total bar height shows $k=20$ recall. Groups contain 4 head, 23 body, and 12 tail labels, defined by training frequency. The ViT series uses the auxiliary per-class record described below.}
    \label{head_body_tail}
\end{figure}

The fixed head/body/tail label groups have observed training-count ranges of 57,530--116,459, 1,329--9,871, and 8--977 samples per label, respectively (main-paper training-frequency histogram). They contain 78.4\%, 20.4\%, and 1.2\% of the 445,985 training samples. Groups are specified by fixed label memberships; these intervals report their observed training-count ranges. The separate ``fewer than 200 test samples'' highlight in the per-class plot below does not define these training-frequency groups.

This original-model comparison excludes the added foundation backbones and API models. The Qwen3-VL fine-tuned and few-shot bars use the ReefNet test summaries. Fine-tuned Qwen3-VL reaches 95.50\%, 80.30\%, and 51.47\% recall on head, body, and tail labels, respectively. BioCLIP2 few-shot tail recall exceeds its head recall at both $k=5$ and $k=20$, so lower tail performance is not universal across adaptation settings.

\noindent\textbf{ViT record scope.} The available ViT-B/16 per-class record underlying this group plot gives 79.48\% overall macro recall, whereas the main result table reports 79.84\%. Their checkpoint correspondence is unverified. We retain the main-table result and present this auxiliary group breakdown separately; it should not be interpreted as a decomposition of the 79.84\% result.

\subsection{Per-Class Analysis}
\label{sup_per_class_vit}

\begin{figure}[t]
    \centering
    \includegraphics[width=0.72\textwidth]{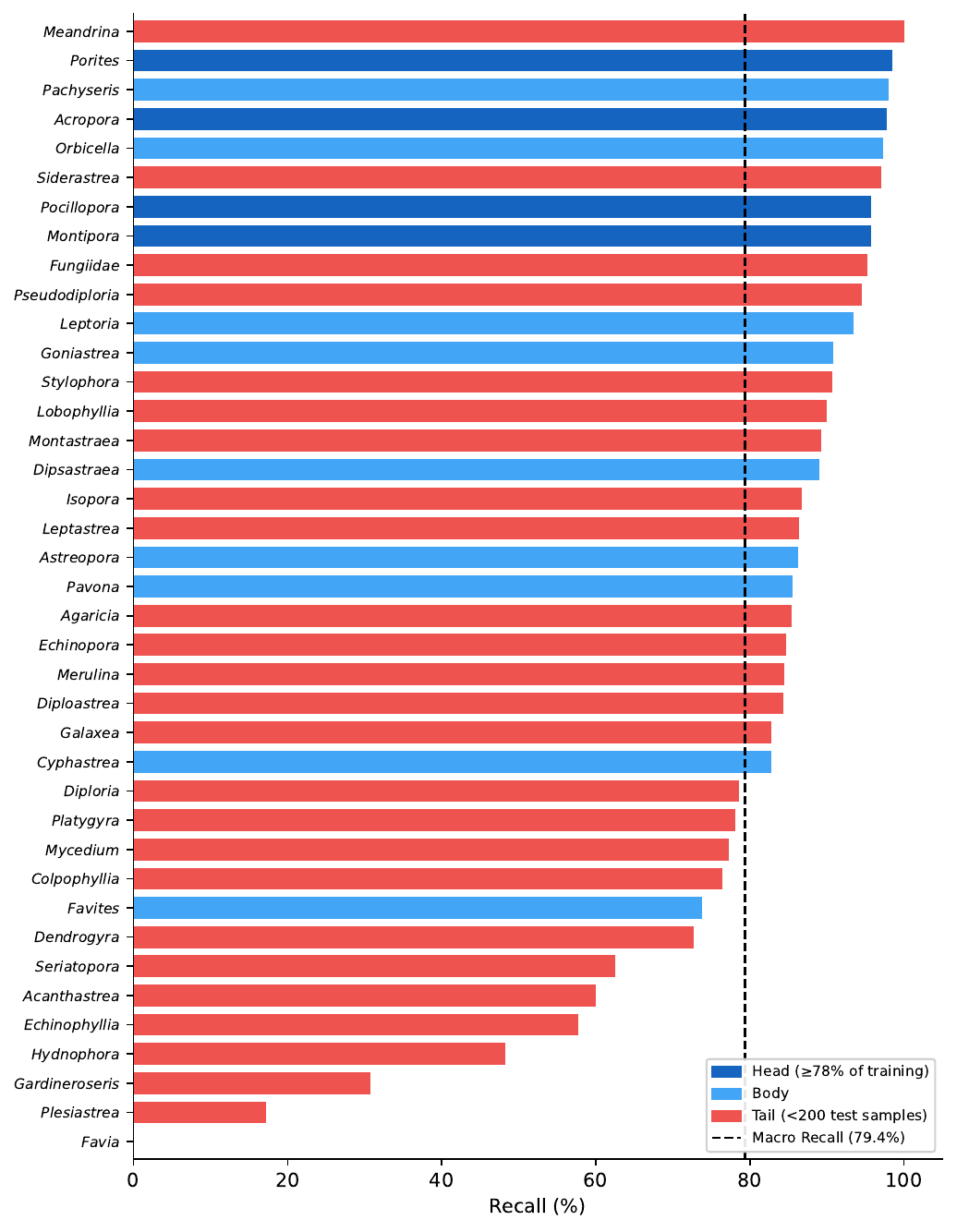}
    \caption{\textbf{Per-class recall of ViT-B/16 on the ReefNet within-source test set.}
    Classes are sorted by recall (ascending).
    Dark blue bars indicate the four head genera (collectively 78.4\% of training samples);
    medium blue bars are body genera;
    red bars identify classes with fewer than 200 test samples; this test-count criterion is distinct from the training-frequency groups above.
    The dashed line shows the macro recall (79.8\%).}
    \label{fig:vit_perclass_recall}
\end{figure}

\begin{figure}[t]
    \centering
    \includegraphics[width=\textwidth]{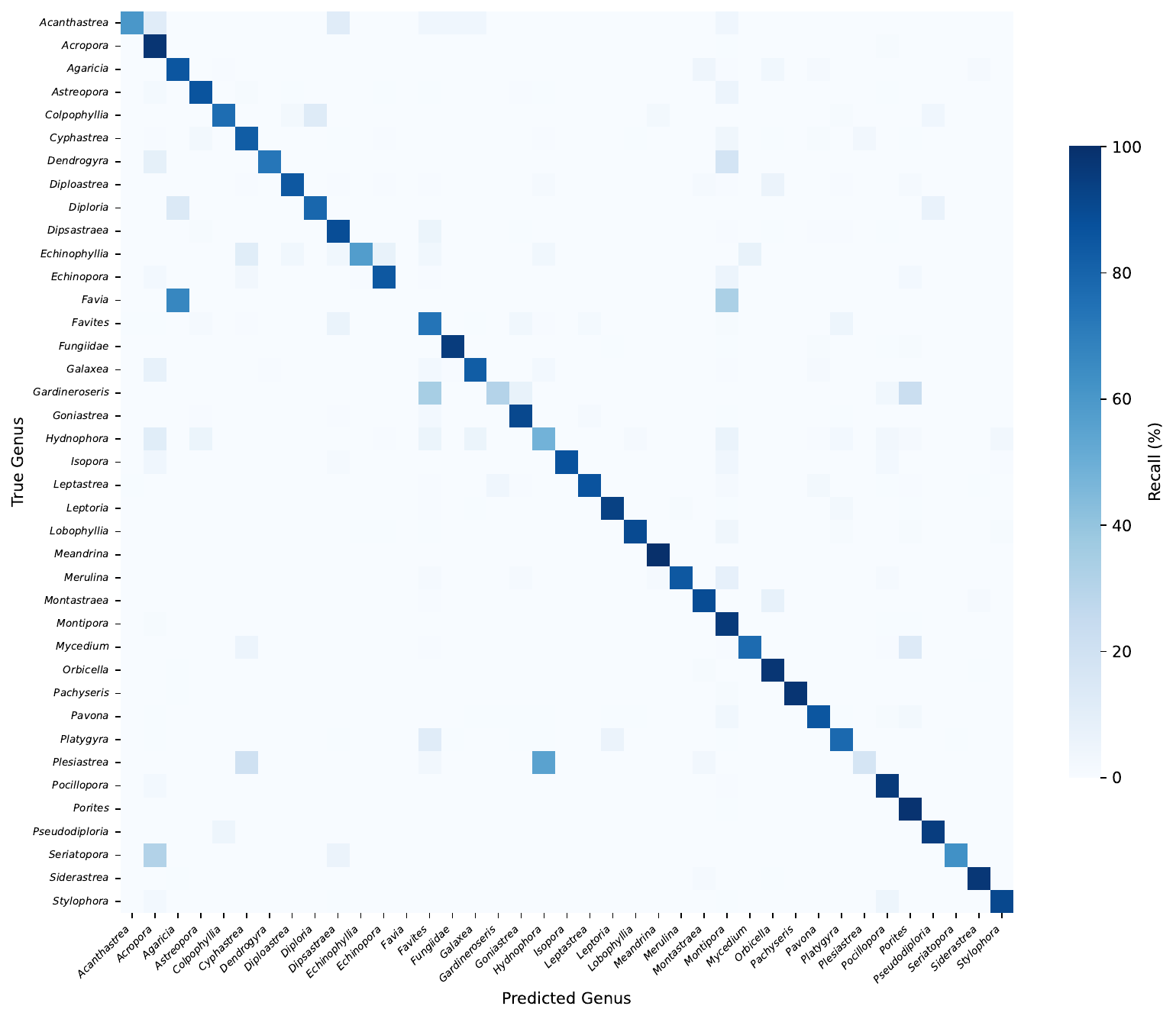}
    \caption{\textbf{Normalized confusion matrix for ViT-B/16 on the ReefNet within-source test set.}
    Each row is normalized by the true class count (row sum = 100\%), so diagonal entries equal per-class recall.
    Notable off-diagonal clusters include the abundant genera (\textit{Acropora}, \textit{Pocillopora}, \textit{Montipora}), the compact-corallite group (\textit{Acanthastrea}, \textit{Dipsastraea}, \textit{Favites}), and pairs such as \textit{Platygyra}$\leftrightarrow$\textit{Favites} and \textit{Astreopora}$\to$\textit{Montipora}.}
    \label{fig:vit_confusion}
\end{figure}

Figures~\ref{fig:vit_perclass_recall} and~\ref{fig:vit_confusion} show the per-class recall and normalized confusion matrix for ViT-B/16 on the ReefNet within-source test set (39 labels, 23{,}043 samples; macro recall 79.8\%).

\noindent\textbf{Per-class recall.}
The four dominant genera—\textit{Acropora}, \textit{Montipora}, \textit{Pocillopora}, and \textit{Porites}—are correctly classified in over 95\% of cases, reflecting their abundance in training (collectively 78.4\% of samples).
Performance degrades progressively for body and tail classes.
The three lowest-recall genera are \textit{Favia} (0\%, $n{=}3$ test samples), \textit{Plesiastrea} (17\%, $n{=}29$), and \textit{Gardineroseris} (31\%, $n{=}26$); for \textit{Favia} the test set contains only 3 samples, making any estimate unreliable.

\noindent\textbf{Confusion patterns.}
The most frequent misclassifications occur among the three most abundant genera: \textit{Acropora}$\to$\textit{Pocillopora} (1.3\%), \textit{Pocillopora}$\to$\textit{Acropora} (2.4\%), and \textit{Montipora}$\to$\textit{Acropora} (1.4\%).
Beyond these abundance-driven confusions, several ecologically interpretable patterns emerge.
Genera with compact, polygonal corallites—\textit{Acanthastrea}, \textit{Dipsastraea}, and \textit{Favites}—are mutually confused: \textit{Acanthastrea}$\to$\textit{Dipsastraea} (12.0\%), \textit{Dipsastraea}$\to$\textit{Favites} (5.9\%), and \textit{Favites}$\to$\textit{Dipsastraea} (6.9\%).
\textit{Platygyra} is confused with \textit{Favites} in 11.7\% of cases, reflecting their similar polygonal skeletal patterns.
\textit{Astreopora} is misclassified as \textit{Montipora} in 5.8\% of cases, a known difficulty arising from their inconspicuous corallite structures in patch-scale images.
\textit{Echinopora} is confused with \textit{Montipora} (5.6\%) and \textit{Porites} (2.4\%), genera sharing comparable surface textures and encrusting growth forms.
Among the lowest-recall genera, \textit{Plesiastrea} ($n{=}29$) is predominantly misclassified as \textit{Hydnophora} (55.2\%) and \textit{Cyphastrea} (20.7\%), while \textit{Gardineroseris} ($n{=}26$) is most often confused with \textit{Favites} (34.6\%) and \textit{Porites} (23.1\%).

\subsection{Overall Accuracy, Precision, and F1 Score}
\label{sup_accuracy_precision_f1}

Table~\ref{tab:supp_additional_metrics} reports macro precision, macro F1, and overall accuracy (micro recall) for the original model suite evaluated on the ReefNet within-source test set (39 labels, 23{,}043 samples), complementing the macro recall reported in the main paper.
Macro recall is included in the table for reference.
A note on precision and F1: because some models—particularly in the zero-shot setting—never predict certain classes, macro-averaged precision and F1 may be artificially deflated for those models (per-class precision is defined as 0 for classes with no predicted positives).
This does not affect macro recall, which depends only on predicted vs.\ ground-truth labels within each class.
As an extreme example, LLaVA-NeXT in zero-shot mode collapses to predicting a single genus (\textit{Favia}) for virtually all inputs, yielding near-zero precision and F1.

\begin{table}[t]
\centering
\caption{\textbf{Additional metrics on the ReefNet within-source test set} (39 labels, 23{,}043 samples).
Macro recall—the primary metric reported in the main paper—is included for reference.
Precision and F1 may be affected by classes absent from model predictions; a caveat noted in Section~\ref{sup_accuracy_precision_f1}.
Bold indicates the best result within each model group.}
\label{tab:supp_additional_metrics}
\resizebox{\textwidth}{!}{%
\begin{tabular}{l cccc}
\toprule
\textbf{Model} & \textbf{Macro Recall} & \textbf{Macro Precision} & \textbf{Macro F1} & \textbf{Accuracy} \\
\midrule
\multicolumn{5}{l}{\textit{Vision Models}} \\
\midrule
\quad ResNet-50 & 45.6 & 55.9 & 48.3 & 80.7 \\
\quad EfficientNet-B3 & 71.3 & 75.7 & 72.4 & 90.2 \\
\quad ViT-B/16 & \textbf{79.8} & \textbf{83.0} & \textbf{80.7} & \textbf{94.6} \\
\quad DeiT-B/16 & 76.2 & 80.8 & 77.6 & 93.7 \\
\quad BEiT-B/16 & 66.7 & 73.7 & 71.0 & 89.6 \\
\quad Swin-B & 77.1 & 81.9 & 79.7 & 93.3 \\
\quad ConvNeXt-B & 74.7 & 80.6 & 76.6 & 92.0 \\
\midrule
\multicolumn{5}{l}{\textit{VLMs --- Zero-Shot}} \\
\midrule
\quad CLIP-L/14 & 5.2 & 3.8 & 3.4 & 24.4 \\
\quad OpenCLIP-L/14 & 4.6 & 3.9 & 3.7 & 29.9 \\
\quad SigLIP & 9.1 & 12.4 & 5.1 & 24.1 \\
\quad BioCLIP & 11.3 & 9.6 & 7.4 & 19.8 \\
\quad \textbf{BioCLIP2} & \textbf{23.5} & \textbf{21.6} & \textbf{17.1} & \textbf{41.2} \\
\midrule
\multicolumn{5}{l}{\textit{MLLMs --- Zero-Shot}} \\
\midrule
\quad LLaVA-NeXT & 2.6 & 0.0 & 0.0 & 0.0 \\
\quad LLaVA-OneVision & 3.3 & 2.1 & 1.1 & 5.4 \\
\quad MiniCPM-V & 3.6 & 3.3 & 2.8 & 22.3 \\
\quad \textbf{Qwen3-VL} & \textbf{5.1} & \textbf{3.8} & \textbf{4.1} & \textbf{35.4} \\
\quad InternVL3 & 4.0 & 3.5 & 2.4 & 15.8 \\
\midrule
\multicolumn{5}{l}{\textit{VLMs --- Fine-Tuned}} \\
\midrule
\quad CLIP-L/14 & 76.3 & 65.0 & 67.6 & 89.3 \\
\quad OpenCLIP-L/14 & 76.8 & 72.9 & 73.8 & 91.8 \\
\quad \textbf{BioCLIP2} & \textbf{79.2} & \textbf{73.8} & \textbf{75.5} & \textbf{92.2} \\
\midrule
\multicolumn{5}{l}{\textit{MLLMs --- Fine-Tuned}} \\
\midrule
\quad MiniCPM-V & 63.4 & 74.9 & 65.3 & 86.6 \\
\quad InternVL3 & 53.4 & 64.7 & 55.5 & 81.2 \\
\quad \textbf{Qwen3-VL} & \textbf{73.0} & \textbf{82.2} & \textbf{75.5} & \textbf{92.1} \\
\bottomrule
\end{tabular}%
}
\end{table}

\FloatBarrier
\subsection{Cross-Model Failure Analysis}
\label{sup_cross_model_failures}
The reported comparison of BioCLIP2 and Qwen3-VL on the same 23,043 ReefNet test samples records 20,348 cases where both are correct (88.3\%) and 940 where both are wrong (4.1\%). It separately reports 2,222 prediction disagreements (9.6\%). Disagreement can include different incorrect predictions from both models, so these three counts are not presented as mutually exclusive categories or summed into a partition of the test set.

High BioCLIP2 confidence does not eliminate errors: in 273 cases, BioCLIP2 is wrong with confidence above 0.90 while Qwen3-VL is correct; in 280 cases, both predict the same incorrect genus with BioCLIP2 confidence above 0.90. Reported joint-failure rates are 30.8\% for \textit{Gardineroseris} and 28.0\% for \textit{Acanthastrea}, compared with 0.5--1.7\% for the four head genera. Reported confusion patterns include \textit{Gardineroseris} predicted as \textit{Porites} and \textit{Acanthastrea} as \textit{Favites}. Table~\ref{tab:cross_model_failure_examples} gives the six label-level examples reported with this analysis.

\begin{table}[htbp]
\centering\small
\begin{tabular}{llrl}
\toprule
Ground truth & BioCLIP2 & Confidence & Qwen3-VL\\
\midrule
\multicolumn{4}{l}{\textit{BioCLIP2 confident but wrong; Qwen3-VL correct (273 cases)}}\\
Cyphastrea & Plesiastrea & 1.00 & Cyphastrea\\
Porites & Leptoria & 0.99 & Porites\\
Pavona & Montipora & 0.98 & Pavona\\
\midrule
\multicolumn{4}{l}{\textit{Both predict the same wrong genus; BioCLIP2 confident (280 cases)}}\\
Favites & Goniastrea & 1.00 & Goniastrea\\
Leptastrea & Gardineroseris & 1.00 & Gardineroseris\\
Stylophora & Acropora & 0.98 & Acropora\\
\bottomrule
\end{tabular}
\caption{Reported examples of complementary and shared classification failures. Confidence values belong to BioCLIP2; the 273 and 280 counts describe the corresponding groups, with three examples shown from each.}
\label{tab:cross_model_failure_examples}
\end{table}

These examples illustrate both complementary predictions and shared errors, without establishing calibrated uncertainty. A genus name generated by Qwen3-VL does not itself supply a class probability; label probabilities require separate scoring. This analysis reports the available aggregate summaries and label-level examples rather than full confusion matrices for the two models.

\section{Additional Qualitative Examples}

\begin{figure}[htbp]
    \centering
    \includegraphics[width=\textwidth]{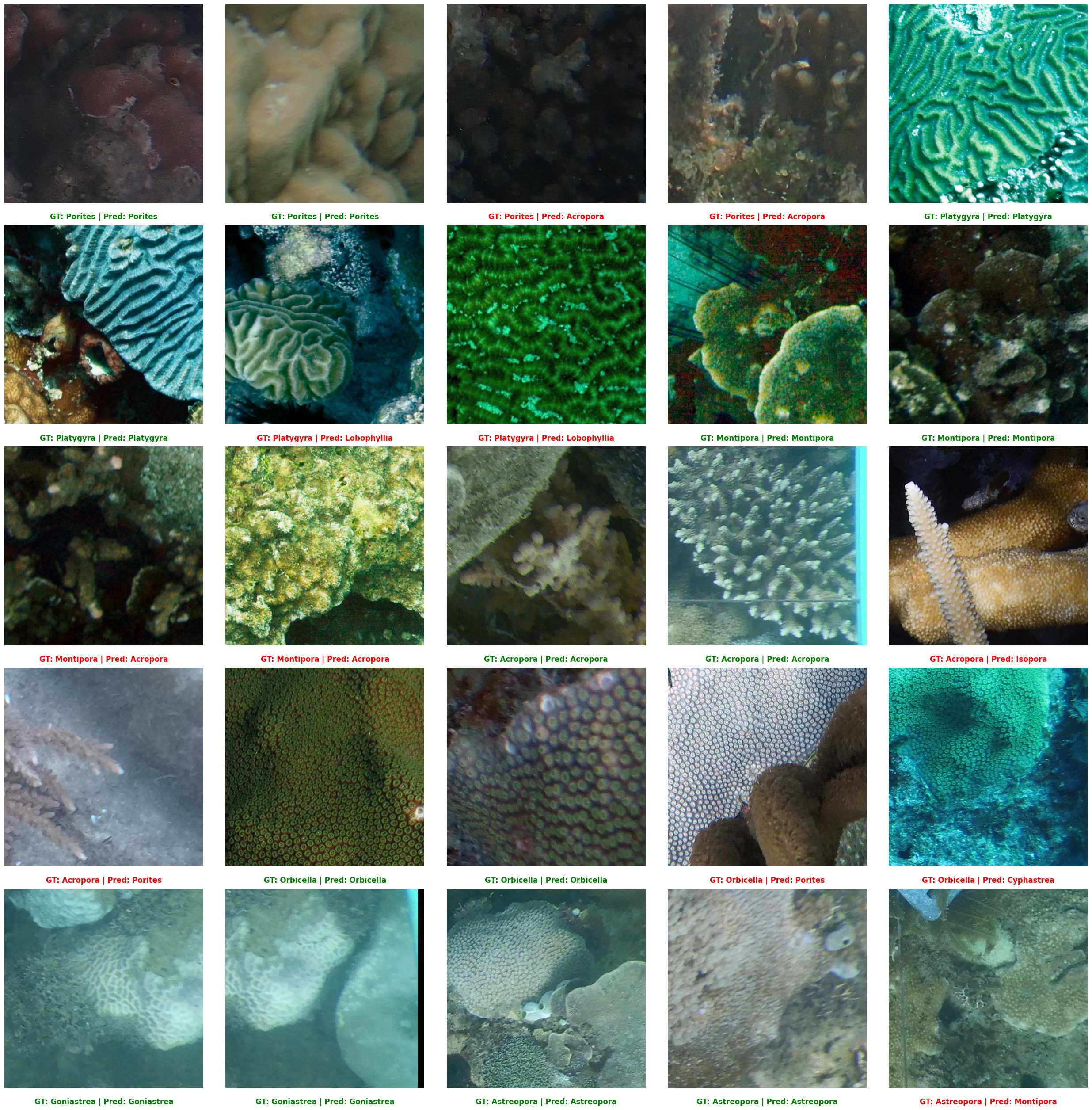}
    \caption{Historical ReefNet/CoralNet qualitative examples from the 2025 cross-source split~00 (34,040 test crops), using an MAE-pretrained ViT. These are not from the current within-source benchmark. GT: ground truth; Pred: prediction. Green/red prediction text indicates correct/incorrect classifications.}
    \label{fig_qualitative_global}
\end{figure}

\begin{figure}[htbp]
    \centering
    \includegraphics[width=\textwidth]{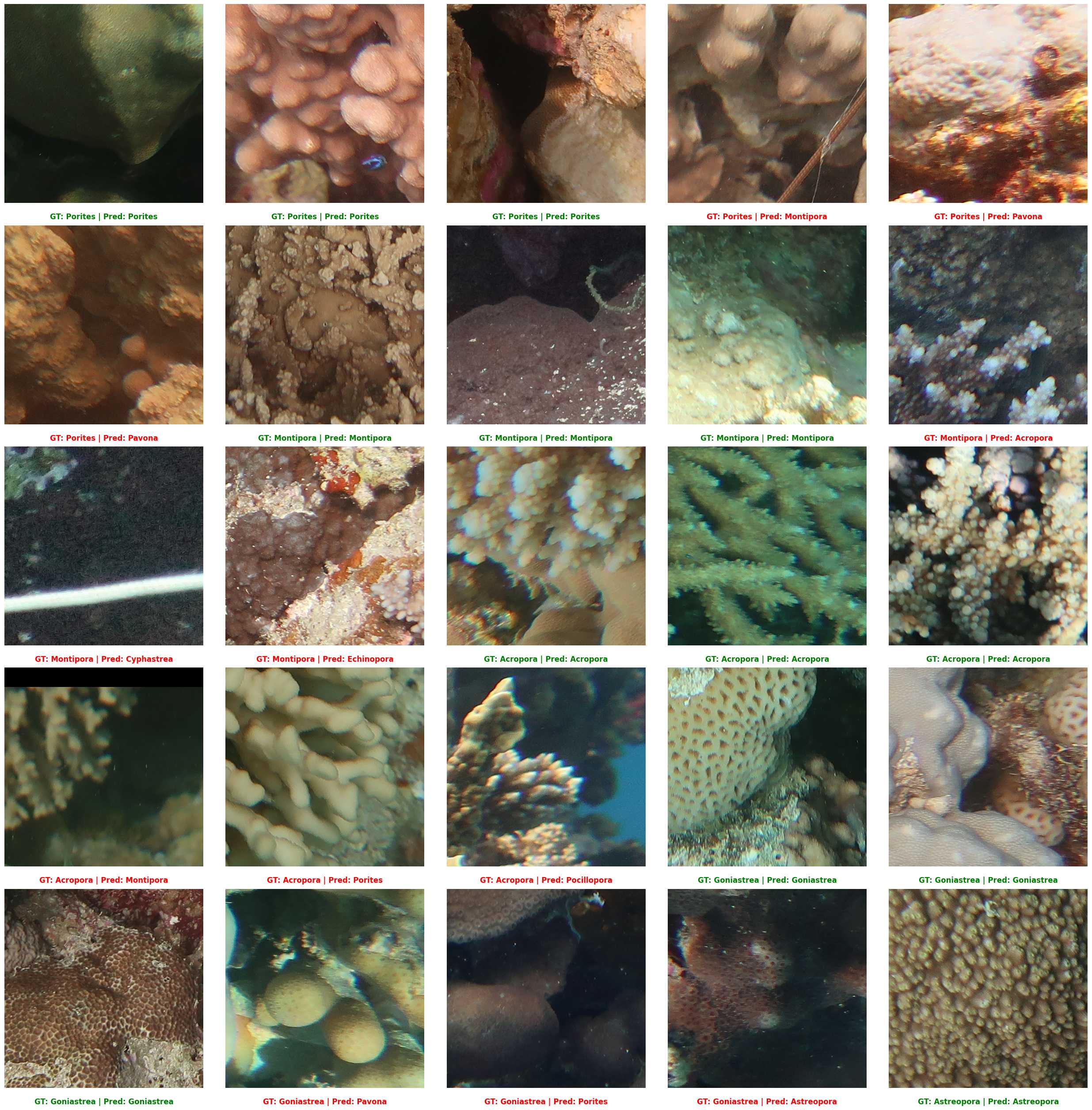}
    \caption{Historical Al-Wajh qualitative examples from the 2025 MAE-pretrained ViT experiment (4,606 test crops, $512\times512$ pixels), distinct from the current 4,624-sample benchmark. GT: ground truth; Pred: prediction. Green/red prediction text indicates correct/incorrect classifications.}
    \label{fig_qualitative_redsea}
\end{figure}

Figures~\ref{fig_qualitative_global} and~\ref{fig_qualitative_redsea} retain illustrative predictions from historical 2025 experiments with an MAE-pretrained ViT. The ReefNet/CoralNet examples use the earlier cross-source split~00 (34,040 test crops); the Al-Wajh examples use an earlier set of 4,606 test crops at $512\times512$ resolution. These are not predictions from the current ViT-B/16 benchmark runs on the 23,043-sample ReefNet and 4,624-sample Al-Wajh tests, and are not evidence for the current tables' numerical results.

Each grid displays ground-truth (GT) and predicted (Pred) labels, with correct predictions in green and errors in red; no confidence scores are shown. In Figure~\ref{fig_qualitative_global}, examples include \textit{Porites} predicted as \textit{Acropora}, \textit{Platygyra} as \textit{Lobophyllia}, and \textit{Orbicella} as \textit{Cyphastrea}. In Figure~\ref{fig_qualitative_redsea}, examples include \textit{Porites} predicted as \textit{Montipora}, \textit{Montipora} as \textit{Echinopora}, and \textit{Acropora} as \textit{Pocillopora}. These selected examples illustrate successful and unsuccessful classifications; they do not estimate class-wise accuracy or confusion frequencies.

\FloatBarrier
\newpage

\section{Text Descriptions and Reasoning}
\label{sup_text_descriptions}
The description condition augments retained taxon names with short morphological descriptions of taxa, rather than descriptions of the individual test image. The main zero-shot table compares direct, answer-only classification using names versus names plus descriptions on the same two test sets. The saved GPT-5.4 prompts use 39 candidates on ReefNet and 13 on Al-Wajh in both text conditions. Its name-only and description comparisons are therefore matched within each test, but its Al-Wajh candidate set differs from the standard 39-label protocol.

Descriptions increase Al-Wajh macro recall for InternVL-3.5 (9.80 to 10.70), MiniCPM-V 4.5 (9.50 to 12.50), Qwen3-VL (15.20 to 16.80), and GPT-5.4 (26.39 to 32.22). On ReefNet, three improve, while InternVL-3.5 decreases from 4.00 to 3.20. Thus the effects depend on both model and evaluation setting. The description-conditioned GPT-5.4 result remains below BioCLIP2's name-only result on each test. Unavailable description results for embedding VLMs are not inferred from name-only performance.
\paragraph{Documentation scope and provenance.}
The exact description file and prompts reproduced below document GPT-5.4. The 39 texts were selected from the 86-row experimental file \texttt{coral\_classes\_agentic.csv}; every benchmark label has exactly one entry. Texts range from 29 to 55 words (mean 36.2). The generation procedure and source attribution for individual descriptions were not recorded, so these texts are reproduced as experimental inputs, not as verified quotations from taxonomy references. Their wording is preserved without editorial correction.

The reported InternVL-3.5, MiniCPM-V 4.5, and Qwen3-VL-8B description scores are retained, but their exact description files and inference prompts were not recovered. The GPT artifacts therefore do not establish that those models used identical descriptions, templates, or parsing. This limits reproducibility of the open-model description comparisons.

\begin{table}[ht]
\centering
\begin{tabular}{llrr}
\toprule
GPT-5.4 response condition & Reference text & ReefNet & Al-Wajh\\
\midrule
Direct classification & Names & 14.18 & 26.39\\
Reasoning before label & Names & 12.33 & 22.91\\
Direct classification & Names + descriptions & 14.92 & 32.22\\
Reasoning before label & Names + descriptions & Not run & 30.39\\
\bottomrule
\end{tabular}
\caption{GPT-5.4 reasoning ablation: macro recall (\%). GPT prompts use 39 candidates on ReefNet and 13 on Al-Wajh. The within-source reasoning-plus-description condition was not run.}
\label{tab:reasoning}
\end{table}

Eliciting reasoning before the label reduces name-only performance on both tests. Adding descriptions increases Al-Wajh performance in the reasoning condition from 22.91 to 30.39, but this remains below direct classification with descriptions (32.22). This experiment therefore does not show an advantage from reasoning over the best matched direct condition. It remains a classification experiment, not a visual question answering benchmark: image-grounded questions about colony form, health, size, or context would require new expert-authored question--answer annotations.

Generated labels and verbal explanations are not calibrated class probabilities. Training-seed variability, confidence calibration, selective accuracy, and prediction-set coverage address different questions. Because same-image crops can overlap (Appendix~\ref{sup_crop_analysis}), predictive-uncertainty evaluation must account for dependence and define calibration/evaluation groups explicitly before claiming coverage guarantees.

\subsection{Description Prompt Templates}
\label{sup_description_prompts}
Each prompt is sent as the text part of the single user message before the image, with no system message. The placeholder \texttt{\{genus\_descriptions\}} expands to one paragraph per candidate, formatted as \texttt{Label: description}, separated by a blank line and ordered alphabetically. The 39 exact descriptions are given in Table~\ref{tab:api_descriptions}. For Al-Wajh only the 13 candidates listed in Appendix~\ref{sup_api_models} are included. Parsing and decoding are specified there as well. The full saved prompts and the description CSV are included in the manuscript source under \texttt{artifacts/api\_text\_inputs/}; their inclusion here does not imply publication on the dataset hosting service.

\paragraph{Direct classification with descriptions.}
This is the exact ReefNet template. The Al-Wajh template differs only by replacing ``39 candidate coral genera'' with ``13 candidate coral genera'' and expanding the placeholder with the corresponding 13 entries.
\begin{Verbatim}
You are a marine biologist specializing in coral taxonomy.

Task:
You are given:
1. An image of a coral colony.
2. A list of 39 candidate coral genera.
3. A short description for each genus, summarizing its distinguishing morphological characteristics.

Your task is to identify which genus the coral colony belongs to.

Instructions:
- Carefully compare the visual characteristics of the coral in the image with the descriptions of all candidate genera.
- Use the genus descriptions as the primary reference when distinguishing between visually similar genera.
- Consider characteristics such as colony growth form, branching pattern, corallite arrangement, surface texture, and any other diagnostic morphological features described.
- Select the single genus whose description best matches the coral in the image.

Rules:
- You must select exactly ONE genus.
- The answer MUST be one of the provided genera.
- Do NOT provide any explanation.
- Do NOT describe the coral.
- Output ONLY the genus name.

Candidate Genera and Descriptions:
{genus_descriptions}
\end{Verbatim}

\paragraph{Reasoning with descriptions.}
For the executed Al-Wajh reasoning-plus-description condition, use the direct template above with 13 candidates and replace its entire Rules block with the following exact text. Keep the candidate-description block unchanged. The full template is also included in the source files. A ReefNet reasoning-plus-description template exists, but that condition was not run and no result is inferred from its existence.
\begin{Verbatim}
Rules:
- First reason explicitly about which candidate descriptions match, and which you
  are ruling out and why, grounding each point in a feature you can see.
- Then select exactly ONE genus.
- The answer MUST be one of the provided genera.
- Respond with ONLY a single JSON object, with no markdown fences and no text
  before or after it, in exactly this form:
  {"reasoning": "<your concise reasoning>", "genus": "<one genus from the list>"}
- The "genus" value must be the genus name alone, spelled exactly as it appears
  in the list below.
\end{Verbatim}

\subsection{Exact Description Texts}
The following descriptions are preserved as used in GPT-5.4 inputs, including historical taxonomic wording. Fungiidae is a family-level benchmark label. These texts characterize the candidate taxon in general and are not image-specific annotations.
\begingroup
\small
\setlength{\tabcolsep}{4pt}
\begin{longtable}{@{}>{\raggedright\arraybackslash}p{0.20\linewidth}>{\raggedright\arraybackslash}p{0.76\linewidth}@{}}
\caption{Exact 39-label description texts used in GPT-5.4 prompts. Al-Wajh prompts include only the 13 candidates listed in Appendix~\ref{sup_api_models}.}\label{tab:api_descriptions}\\
\toprule
Label & Description \\
\midrule
\endfirsthead
\multicolumn{2}{l}{\tablename~\thetable{}: Exact description texts (continued)}\\
\toprule
Label & Description \\
\midrule
\endhead
\midrule
\multicolumn{2}{r}{Continued on next page}\\
\endfoot
\bottomrule
\endlastfoot
\textit{Acanthastrea} & The Acanthastrea coral is characterized by its large, fleshy polyps, which are circular to irregular in outline, often showcasing a mottled palette of dull browns, greens, and occasional vivid hues. Its colony is encrusting and broad, resembling a flat, stone-like surface, accentuated by spine-like structures and gentle, folded textures that convey a spiny, rich complexity. \\
\textit{Acropora} & Acropora is a vibrant, table-like coral with porous skeletal structures, exhibiting bright blue, green, yellow, or brown hues. It features flattened branches with ridges and unique tubular corallites at tips, growing over 2 meters wide, forming dynamic underwater habitats. \\
\textit{Agaricia} & Agaricia corals are massive and plate-like, featuring tan to yellow-brown colors with deep valleys and irregular fronds. They form complex horizontal plates, characterized by clear septocostae and a substantial, textured surface. \\
\textit{Astreopora} & Astreopora is a massive, hemispherical coral with rough, spine-covered surfaces in cream, beige, or brown, displaying crater-like corallites. These distinct, irregularly spaced structures reveal vibrant polyp tissue in hues of white, bright green, or yellow. \\
\textit{Colpophyllia} & Large, domed colonies resembling a human brain, with meandering brown ridges and valleys in whitish, green, or tan. Can grow up to 2 meters, robust, with retracted polyps by day, color variations enhancing its distinctive appearance. \\
\textit{Cyphastrea} & The Cyphastrea coral genus features massive, encrusting to columnar colonies with smooth or hillocky surfaces. Color varies from mottled grey to vibrant hues, with rounded corallites exhibiting 10-12 primary septa, creating a visually dense and intricate structure. \\
\textit{Dendrogyra} & Dendrogyra, or pillar coral, showcases tall, cylindrical columns resembling fingers, often in shades of beige and brown. Its polyp-topped structure has a furry texture, while the rounded base features conical skeletal processes, creating a striking, fan-like silhouette. \\
\textit{Diploastrea} & Imagine an enormous, dome-shaped coral, reaching up to 2 meters tall, with a smooth, even surface. Its color varies from uniform cream to gentle greenish tones. Thick, towering corallites form low cones, creating stability and resilience. \\
\textit{Diploria} & Diploria labyrinthiformis exhibits a hemispherical, brain-like shape, ranging in colors from brown to yellow and grey. Its deep, interconnected valleys, measuring up to 10 mm, cradle retractable polyps, lending a textured, varied surface to this robust coral. \\
\textit{Dipsastraea} & Dipsastraea corals exhibit large, dome-shaped colonies in beige or greenish yellow, with tightly fused, steeply plunging corallite walls. They feature retracted polyps and a noticeable groove at the upper edge, enhancing their intricate structure. \\
\textit{Echinophyllia} & The Echinophyllia genus showcases vase-shaped colonies, 8-10 cm wide, with circular polyps that exhibit brown or dark green colors. Its fleshy texture and distinct bulbous corallite structures enhance its intricate, pitted surface. \\
\textit{Echinopora} & The Echinopora genus features diverse corals, displaying vivid hues from amber to deep brown and pale greens. With plate-like shapes, prominent spines, and intricate textures of spiky dentations, they typically span 6 inches to 2 feet in size. \\
\textit{Favia} & Favia coral showcases massive, rounded colonies in a breathtaking array of colors, from pale yellows to vibrant reds and greens, featuring conical, slightly irregular corallites and fleshy polyps that hide in distinct patterns, creating a textured visual feast of nature. \\
\textit{Favites} & The Favites coral genus displays massive, dome-shaped colonies in various shades of brown, green, yellow, and purple, featuring polygonal and circular corallites. Its surface showcases intricate patterns, with fleshy, retracted polyps, projecting a robust, textured presence. \\
\textit{Fungiidae} & Fungiidae corals, also known as mushroom or plate corals, feature solitary, disc-like shapes with strikingly colorful mouths. Their textures range from smooth to coarse, with some species exhibiting long, soft tentacles, resembling anemones, and a vibrant palette. \\
\textit{Galaxea} & Galaxea coral features encrusting, massive colonies with oval to circular polyps that measure 1-7 cm wide. They exhibit vibrant colors such as green, brown, and contrasting white-tipped tentacles, creating a striking textural appearance. \\
\textit{Gardineroseris} & The Gardineroseris genus displays massive, encrusting colonies in hues of brown, beige, or greenish-yellow. Its polygonal corallites, measuring 4-8 mm, feature acute ridges, transparent short tentacles, and a central pillar-like columella, creating intricate textures. \\
\textit{Goniastrea} & Goniastrea corals form massive, spherical to dome-shaped colonies in beige or yellowish hues, with deep valleys and tall corallite walls. Their well-developed paliform lobes enhance their brain-like texture, while retracted polyps reveal intricate, polygonal structures by night. \\
\textit{Hydnophora} & Hydnophora displays stunning sprawling colonies, often composed of submassive encrusting shapes. Their color palette includes soothing cream, dull green, and ochre, adorned with charming conical mounds and short, furry tentacles, evoking a rich depth and texture. \\
\textit{Isopora} & The Isopora coral genus exhibits encrusting plates or thick ridges in pale cream or brown. Its broad, characteristic branches are smooth and rounded, often displaying slight tints of purple or pink, with fine spinules enhancing its texture. \\
\textit{Leptastrea} & Leptastrea corals feature a shaggy, encrusting form, showcasing vibrant colors like emerald green and cream beige. Their surface is marked by irregular grooves, while tightly packed, soft tentacles extend, creating a fuzzy texture, providing a rich visual experience. \\
\textit{Leptoria} & Leptoria corals present massive, ridged colonies, smooth surfaces in cream, brown, or green, with sinuous valleys and uniformly spaced septa. Their plate-like columellae lack central points, creating an intricate, textured underwater landscape. \\
\textit{Lobophyllia} & The Lobophyllia coral genus exhibits massive dome-shaped colonies with large, spiny, and vibrant polyps. Colors include striking greens, browns, and purples with notable darker outlines, forming a visually intricate tapestry of life beneath the ocean's surface. \\
\textit{Meandrina} & Meandrina, known as maze or brain coral, features submassive colonies with irregular, smooth surfaces and deep valleys. Its grey or pale brown color envelops large, rounded verrucae, resembling a labyrinth, inviting gentle touches across its textured body. \\
\textit{Merulina} & Merulina coral presents varied laminar, foliose, or branching structures in earthy tones of brown, blue, or green, characterized by thick ridges and fan-like valleys, featuring long, infrequently visible tentacles within. \\
\textit{Montastraea} & The Great Star Coral, Montastraea cavernosa, exhibits massive, dome-shaped colonies, reaching several meters. Its surface, adorned with large polyps, showcases shades of pale to deep brown, punctuated by vibrant fluorescent green and reddish hues, embodying the reef's intricate beauty. \\
\textit{Montipora} & Montipora exhibits a variety of growth forms, including stout branches and encrusting surfaces, displaying colors of beige, pinkish-brown, or green, with small, spiny polyps clustered at irregular intervals, giving it a textured and diverse appearance. \\
\textit{Mycedium} & Mycedium corals exhibit large, vase-shaped forms with contorted leaf-like plates, displaying bright colors around polyp mouths. Their surface features distinct, regular ridged patterns, with corallites 5-15 mm wide and arranged in neat, concentric rows. \\
\textit{Orbicella} & The Orbicella coral genus features massive, cylindrical colonies up to 3 meters wide, displaying shades of yellow or beige. Its rounded corallites, 4-5 mm wide, are surrounded by thick walls, creating a distinct lobed appearance in reef habitats. \\
\textit{Pachyseris} & The Elephant Skin Coral exhibits a greyish-beige to brown, corrugated surface with contorted ridges, resembling rugged hills. Its frond-like, vase-shaped colonies house transparent polyps nestled in valleys, creating an intricate, textured landscape across vast underwater expanses. \\
\textit{Pavona} & Pavona corals exhibit massive, meandering forms with rough surfaces of raised ridges, typically in pale brown or greenish-brown hues. Their polyp tentacles extend vibrantly at night, revealing white mouths, creating a lively, flower-like appearance. \\
\textit{Platygyra} & Platygyra coral exhibits massive, meandroid formations resembling brain structures, with thick, ragged brown walls contrasting against green or grey valleys. Its dome or plate shapes, retracted polyps, and intricately patterned surfaces create a visually striking marine presence. \\
\textit{Plesiastrea} & Plesiastrea, or bubble coral, showcases large, hemispherical colonies up to 80 cm wide, featuring smooth or lobate tissue vesicles, meandering corallites, and hues of yellow, cream, or green, presenting a vibrant, textured underwater landscape. \\
\textit{Pocillopora} & Pocillopora appears as a dense, cauliflower-like structure with robust, cylindrical branches adorned with textured verrucae. Its colors range from pale brown to vibrant pink, and the rounded ends are often soft to the touch, resembling a living sculpture. \\
\textit{Porites} & Porites corals appear as robust, irregular colonies of brown or greenish-brown, featuring finger-like lobes and well-formed corallites with pitted surfaces. Their textures vary from smooth to rough, enhancing their intricate shapes in diverse reef habitats. \\
\textit{Pseudodiploria} & Pseudodiploria displays massive, dome-shaped colonies with knobby, textured surfaces in beige, yellow, or greyish-brown tones, featuring prominent ridges and valleys, while housing small worms that enhance its unique structure. \\
\textit{Seriatopora} & The Seriatopora coral genus showcases delicate, tapering branches that form intricate, bush-like structures. Its polyp-covered surfaces can exhibit a palette of colors ranging from deep brown to soft pink, with tips often pale or white, enriching the underwater landscape. \\
\textit{Siderastrea} & The massive starlet coral presents as rounded boulder-shaped structures, boasting a smooth, dimpled surface of a light reddish-brown hue, with shallow, symmetrical corallites measuring 3-5 mm, emanating an intricate pattern beneath the sea's surface. \\
\textit{Stylophora} & The Stylophora genus, or Cat's Paw corals, features branching colonies with smooth, rounded tips. Colors range from brown and beige to pink and purplish-brown, with paler tips. Their small polyps extend during the day, showcasing vivid, knob-like tentacles. \\
\end{longtable}
\endgroup

\section{Limitations}
\label{sup_limitations}
\paragraph{Geography and label space.}
ReefNet covers 26 ecoregions but is geographically imbalanced. Al-Wajh is one independently collected regional stress test, not a systematic region-disjoint study or evidence of global robustness. Its 13-genus macro average and ReefNet's 39-label average are not directly comparable generalization gaps. The benchmark is closed-set identification: images without retained hard-coral points are not negatives, and detection or open-set recognition would require separate protocols and verified labels.

\paragraph{Taxonomy and audit uncertainty.}
Taxonomy evolves; WoRMS AphiaIDs support traceability~\cite{worms}. Fungiidae is deliberately retained at family level because photo-quadrat imagery cannot reliably resolve all constituent genera. The 92\% agreement refers to retained patches in a non-proportional stratified audit, not exhaustive label accuracy. Residual disagreement and uncertainty remain, and verification depends on scarce expert time.

\paragraph{Point-based supervision.}
A crop can contain several taxa even though its target is the center point. Sparse annotations do not provide exhaustive masks or image-grounded explanations. Same-image crops overlap and are statistically dependent, although image grouping keeps them in the same split. Full-scene segmentation and image-grounded question answering require additional annotations; ReefNet complements coarse coral segmentation resources~\cite{coralscop,coralvos}.

\paragraph{Experimental scope and reproducibility.}
Model families differ in pretraining, capacity, and compute; the comparison is not compute-normalized. Most scores are single-run estimates, and the three-seed ViT and BioCLIP2 studies do not establish variability for other models. The Qwen3-VL seed study uses a class-capped subset and does not establish full-data variability. The augmentation study covers one architecture and one augmented run. Neither a generated label nor a reasoning trace establishes calibrated uncertainty or reliable abstention. API request settings and exact GPT text inputs are documented, but immutable provider snapshots and the open-model description-run inputs are unavailable. The 13-candidate Claude/GPT Al-Wajh prompts also limit comparisons with models using the standard 39-label protocol. The public data release is available, while the complete executable benchmark workflow and versioned description artifact remain release gaps (Appendix~\ref{sup_resources}).

\paragraph{Potential impact.}
Scalable coral recognition can support ecological monitoring, but errors concentrated in rare taxa or underrepresented regions may bias abundance estimates and conservation decisions. Benchmark recall alone is insufficient to validate an operational monitoring system. Deployment requires local validation and appropriate expert review; the present experiments do not establish ecological decision accuracy.

\section{Contributor Attribution and Ethical Data Use.}
To ensure transparency, traceability, and proper credit to original data providers, we compiled a comprehensive list of all 85 CoralNet sources that passed the pre-final curation stages of ReefNet, along with their corresponding contributors and institutional affiliations. This attribution acknowledges the global community of researchers and practitioners who have made their data publicly available through CoralNet, often as part of long-term ecological monitoring efforts. The information was curated through a combination of CoralNet metadata, institutional websites, and direct outreach to contributors. In line with principles of responsible data stewardship, we have preserved all original attribution metadata and encourage future users of ReefNet to do the same. Note that while the attribution covers all 85 sources that passed pre-final filtering, the final ReefNet dataset consists of 76 sources after the ecological and taxonomic quality filters described in Section~\ref{sup_data_collection}; the classification benchmark in the main paper further restricts to the 68 high-confidence sources that met expert-agreement criteria.

\section{Resource Availability}
\label{sup_resources}
The dataset is hosted at \url{https://huggingface.co/datasets/ReefNet/ReefNet-1.0}. The public listing checked on 9 October 2026 includes \texttt{All\_ReefNet\_annotations.csv}, source and annotation metadata workbooks, ReefNet and Al-Wajh label-mapping workbooks, Al-Wajh annotations and imagery, and \texttt{Overview\_Sources\_for\_Image\_Download.csv}. The dataset card declares a CC BY-NC-SA 4.0 license; users must also respect the applicable terms and attribution for original source assets. The contributor list below covers the 85 sources retained at the pre-final curation stage, rather than the final 68-source benchmark.

The benchmark entry point is \url{https://github.com/ReefNet-Project/ReefNet_DS_BenchMark}. The inspected repository revision (\href{https://github.com/ReefNet-Project/ReefNet_DS_BenchMark/tree/63307fa3fccf7025556ca6fc572340996170a9df}{\texttt{63307fa}}) contains a README and license describing data access; it does not yet supply the complete executable benchmark workflow. The README points to \url{https://github.com/ReefNet-Project/coralnet_crawler} for CoralNet image retrieval. The annotation/image download overview preserves links to the source providers.

The inspected dataset listing does not contain a versioned 39-label description file. Appendix~\ref{sup_text_descriptions} now reproduces the exact GPT-5.4 descriptions and templates; the manuscript source includes their CSV and saved prompts. This local inclusion does not establish publication in the public dataset or completeness of the open-model description inputs. Complete experiment configurations, checkpoint identifiers, scoring scripts, and a table-to-output reproduction map are still required to reproduce all reported experiments. Trained-checkpoint availability has not been established by these dataset and repository listings. These release gaps should not be inferred to be resolved from the existence of the resource landing pages.

\newpage
\begin{footnotesize}
\centering
\begin{longtable}[p]{@{}p{\dimexpr0.25\textwidth-4\tabcolsep/3\relax} p{\dimexpr0.5\textwidth-4\tabcolsep/3\relax} p{\dimexpr0.25\textwidth-4\tabcolsep/3\relax}@{}}
\caption{Contributors and affiliations for the 85 CoralNet sources retained at the pre-final curation stage (76 sources in the curated release; 68 in the classification benchmark). This list is based on publicly available information from the CoralNet platform, supplemented by our in-depth online research into each source and outreach to their original contributors.}
\label{tab_contributors} \\

\textbf{CoralNet source} & \textbf{Listed contributors} & \textbf{Listed affiliations} \\
\midrule
\endfirsthead

\textbf{CoralNet source} & \textbf{Listed contributors} & \textbf{Listed affiliations} \\
\midrule
\endhead

\midrule
\multicolumn{3}{r}{\textit{Continued on the next page}} \\
\midrule
\endfoot

\bottomrule
\endlastfoot

2002	&	Zuhairah Dindar	&	University of New South Wales	\\
2016	&	Zuhairah Dindar	&	University of New South Wales	\\
2Maldives\_\allowbreak{}2Katie	&	Katie Lubarsky, Hugh Runyan	&	University of California San Diego	\\
403 He'eia Restoration	&	Cynthia Hunter, Dorian Brunzelle, Kaitlyn Jacobs, Alana Minato, Cody Powers, Connor Antonis, Devon Stapleton, Dylan Rich, Jeany Robledo, Jacob Nygaard, Jordan Pounds-Crihfield, Jacquelyn Simpson, Kevin Christensen, Kaitlin Hooper, Madeline Payne, Renee Wold, Shayna Arakaki, Zachary Clark	&	Hawaii Institute of Marine Biology, University of Hawaii at Manoa	\\
403 He'eia Restoration '21	&	Cynthia Hunter, Dorian Brunzelle, Kaitlyn Jacobs, Keisha Bahr, Brooklyn Bennett, Marisa Bhao-Intr, Brittany Kernodle, Corey Ling, Dan Zhuo, Desiree Shaw, Johann Vollrath, Kenzie Vierra, Lena Marinkovich, Lauryn Pisciotto, Mellisa Gajardo, Madeleine Perez, Sophia Hanscom, Samantha Thomas, Toranosuke Degawa, Zada Boyce-Quentin	&	Hawaii Institute of Marine Biology, University of Hawaii at Manoa	\\
Ain Sukhna Sea Urchin and Coral Cover	&	Omar Attum	&	Indiana University Southeast	\\
Ant	&	Hugh Runyan	&	Scripps Institution of Oceanography, University of California San Diego	\\
Antigua\_\allowbreak{}weird	&	Hugh Runyan	&	Scripps Institution of Oceanography, University of California San Diego	\\
APN Project	&	Roslizawati Ab Lah, Kirsten Benkendorff, Zoe White	&	University of Malaysia Terengganu, Southern Cross University	\\
BHP GSR Fiji	&	John Stratford, Jason Lynch	&	University College London, Zoological Society London	\\
BIO 403	&	Morgan Guadagnoli, Ana Velasquez, Eliza Beckwith, Haley Weis, Isabella Davila, Jamie Mazurski, Rachel Bagnas, Terra Stevens	&	University of Hawaii at Manoa	\\
BIO403 Green Group	&	Morgan Guadagnoli, Ana Velasquez, Eliza Beckwith, Haley Weis, Terra Stevens	&	University of Hawaii at Manoa	\\
BTS\_\allowbreak{}DEF	&	Henrique dos Santos, Igor Cruz, Joao Ferreira, Ian Vinicio	&	Universidade Federal da Bahia	\\
Cape Maeda2021	&	Tomofumi Nagata	&	Okinawa Environment Science Center	\\
Capstone 2023	&	Emily Ogawa, David Hyrenbach, Kristina Bechthold, Leslie Rosa, Ivy Haxo	&	Hawaii Pacific University	\\
carib	&	Hugh Runyan, Nathaniel Hanna Holloway	&	Scripps Institution of Oceanography, University of California San Diego	\\
carib\_\allowbreak{}bitbybit	&	Hugh Runyan	&	Scripps Institution of Oceanography, University of California San Diego	\\
Chao Jing Park Surveys &	Emma Chen, Shinya Shikina, Kaixiang Yang, Chu Yu Ling, Joey Hsia, Yu-Chund Chuan, Tzu-Cheng Lin, Cheng Yin Chu, Wen Teng Huang, Yuenyi Leung	&	National Taiwan Ocean University	\\
CoralNet	&	Dong Li	&	Zhejiang University	\\
CoralNet\_\allowbreak{}Assistance\_\allowbreak{}Test\_\allowbreak{}0921	&	Hugh Runyan, Ceiba Becker, Esmaralda Alcantar, Nicole Pedersen	&	Scripps Institution of Oceanography, University of California San Diego	\\
CREP-REA HAWAII \cite{NOAA_NCRMP_HI_2018}	&	Annette DesRochers, Andrew Gray, Brett Schumacher, Bernardo Vargas Angel, Courtney Couch, Ivor Williams, Jonathan Charendoff, Morgan Winston Pomeroy, Paula Misa, Tom Oliver, Troy Kanemura, Ari Halperin, Chelsie Counsell, Colt Davis, Isabelle Basden, Jon Ehrenberg, Kerry Reardon, Mia Lamirand, Roseanna Lee	&	National Oceanographic and Atmospheric Administration, Pacific Islands Fisheries Science Center	\\
CREP-REA MARIANAS \cite{NOAA_NCRMP_2018}	&	Annette DesRochers, Andrew Gray, Bernardo Vargas Angel, Courtney Couch, Jonathan Charendoff, Morgan Winston Pomeroy, Paula Misa, Tom Oliver, Troy Kanemura, Kaylyn McCoy, Kevin Lino, Marie Ferguson, Mia Lamirand, Nalani Kito-Ho, Winter Jimenez, Brett Schumacher	&	National Oceanographic and Atmospheric Administration, Pacific Islands Fisheries Science Center	\\
CREP-REA NWHI 2017 Tiny Photos \cite{NOAA_NCRMP_HI_2018} 	&	Andrew Gray, Bernardo Vargas Angel, Courtney Couch, Jon Ehrenberg	&	National Oceanographic and Atmospheric Administration, Pacific Islands Fisheries Science Center	\\
CREP-REA SAMOA/PRIA	\cite{NOAA_NCRMP_SAMOA_2018} &	Annette DesRochers, Andrew Gray, Brett Schumacher, Bernardo Vargas Angel, Courtney Couch, Ivor Williams, Jonathan Charendoff, Morgan Winston Pomeroy, Paula Misa, Tom Oliver, Troy Kanemura, Isabelle Basden, Mia Lamirand, Andrew Shantz, Brittany Huntington, Corinne Amir, Hatsue Bailey, Marie Ferguson, Mollie Asbury, Nalani Kito-Ho, Winter Jiminez, Helen Ford, Nicole Kamalu	&	National Oceanographic and Atmospheric Administration, Pacific Islands Fisheries Science Center	\\
curacao	&	Hugh Runyan	&	Scripps Institution of Oceanography, University of California San Diego	\\
ESD\_\allowbreak{}REA HAWAII\_\allowbreak{}v2	 \cite{NOAA_NCRMP_HI_2018}  &	Annette DesRochers, Bernardo Vargas Angel, Courtney Couch, Jonathan Charendoff, Morgan Winston Pomeroy, Tom Oliver, Isabelle Basden, Jon Ehrenberg, Hatsue Bailey, John Morris, Mia Larimand, Paula Misa	&	National Oceanographic and Atmospheric Administration, Pacific Islands Fisheries Science Center	\\
ESD\_\allowbreak{}REA\_\allowbreak{}MARIANA\_\allowbreak{}v2 \cite{NOAA_NCRMP_2018}	&	Andrew Gray, Bernardo Vargas Angel, Courtney Couch, Jonathan Charendoff, Kaylyn McCoy, Tom Oliver, Ari Halperin, Mia Lamirand, Hatsue Bailey, Nicolas Osborn, Jon Ehrenberg	&	National Oceanographic and Atmospheric Administration, Pacific Islands Fisheries Science Center	\\
ESD\_\allowbreak{}REA\_\allowbreak{}SAMOA\_\allowbreak{}PRIA\_\allowbreak{}v2 \cite{NOAA_NCRMP_SAMOA_2018}	&	Andrew Gray, Bernardo Vargas Angel, Courtney Couch, Jonathan Charendoff, Morgan Winston Pomeroy, Paula Misa, Ari Halperin, Isabelle Basden, Mia Lamirand, Hatsue Bailey, Nicolas Osborn, Tom Oliver	&	National Oceanographic and Atmospheric Administration, Pacific Islands Fisheries Science Center	\\
French Polynesia SEA Reef Surveys	&	Elliott Bates	&	International Master of Science in Marine Biological Resourses, Sea Education Association	\\
Hawaii\_\allowbreak{}Leleiwi	&	Russel Sparks, Devon Aguiar	&	Department of Land and Natural Resources - Aquatics	\\
island wide	&	Daniela Escontrela, Elena Turner	&	University of Hawaii	\\
Kahekili	&	Bernardo Vargas Angel, Ivor Williams, Andrew Gray,Tye Kindinger, Mia Lamirand,	&	National Oceanographic and Atmospheric Administration, Pacific Islands Fisheries Science Center	\\
Kimbe Bay study	&	Alice Williams, Kitty Watts	&	University of Bristol	\\
Kulbul	&	Ben Murphy, Caitlin Younis, Hannah Kish, Azri Saparwan, Justin Bovery-Spencer, Tarquin Singleton	&	GBR Biology	\\
mal100	&	Hugh Runyan	&	Scripps Institution of Oceanography, University of California San Diego	\\
Maldives\_\allowbreak{}2021	&	Katie Lubarsky, Hugh Runyan, Anupama Sethuraman, Ceiba Becker, Jamie Pettengell	&	Scripps Institution of Oceanography, University of California San Diego	\\
Maldives\_\allowbreak{}2022	&	Hugh Runyan	&	Scripps Institution of Oceanography, University of California San Diego	\\
Maldives\_\allowbreak{}bitbybit3	&	Hugh Runyan	&	Scripps Institution of Oceanography, University of California San Diego	\\
maldives\_\allowbreak{}bbb5	&	Hugh Runyan	&	Scripps Institution of Oceanography, University of California San Diego	\\
maldives\_\allowbreak{}check2020	&	Hugh Runyan	&	Scripps Institution of Oceanography, University of California San Diego	\\
maldives\_\allowbreak{}check2021	&	Hugh Runyan	&	Scripps Institution of Oceanography, University of California San Diego	\\
Maldives\_\allowbreak{}Katie	&	Hugh Runyan, Katie Lubarsky	&	Scripps Institution of Oceanography, University of California San Diego	\\
maldives\_\allowbreak{}theirs2	&	Hugh Runyan	&	Scripps Institution of Oceanography, University of California San Diego	\\
Maui\_\allowbreak{}Honolua	&	Russel Sparks	&	Department of Land and Natural Resources - Aquatics	\\
Maui\_\allowbreak{}Kanahena Bay	&	Russel Sparks	&	Department of Land and Natural Resources - Aquatics	\\
Maui\_\allowbreak{}Mahinahina	&	Russel Sparks	&	Department of Land and Natural Resources - Aquatics	\\
Maui\_\allowbreak{}Molokini	&	Russel Sparks	&	Department of Land and Natural Resources - Aquatics	\\
Maui\_\allowbreak{}Olowalu	&	Russel Sparks, Tatiana Martinez	&	Department of Land and Natural Resources - Aquatics	\\
Maui\_\allowbreak{}Papa'ula	&	Russel Sparks, Tatiana Martinez	&	Department of Land and Natural Resources - Aquatics	\\
Maui\_\allowbreak{}Puamana	&	Russel Sparks	&	Department of Land and Natural Resources - Aquatics	\\
Maunalua Bay	&	Paula Moehlenkamp	&	Univeristy of Hawaii	\\
Maunalua Bay Nearshore (official)	&	Pamela Weiant, Alexandria Barkman	&	Malama Maunalua	\\
micro\_\allowbreak{}bbb4	&	Hugh Runyan	&	Scripps Institution of Oceanography, University of California San Diego	\\
Micro\_\allowbreak{}challenge	&	Hugh Runyan, Katie Lubarsky	&	Scripps Institution of Oceanography, University of California San Diego	\\
micro\_\allowbreak{}combined	&	Hugh Runyan, Katie Lubarsky, Chris Sullivan, Ahmyia Cacapit, Charles Hambley, Isa Bersamin, Sarah Romero	&	Scripps Institution of Oceanography, University of California San Diego	\\
Micronesia\_\allowbreak{}2022	&	Hugh Runyan, Katie Lubarsky, Chris Sullivan	&	Scripps Institution of Oceanography, University of California San Diego	\\
Micronesia\_\allowbreak{}bitbybit2	&	Hugh Runyan	&	Scripps Institution of Oceanography, University of California San Diego	\\
MLC2008	&	Dong Li	&	Zhejiang University	\\
Moreton Bay Transects	&	Joshua Wirth, Gal Eyal	&	University of Queensland	\\
N. Caribbean ACA	&	Alexandra Ordonez Alvarez, Brianna Bambic, Myles Phillips, Bernadette Charpentier	&	National Geographic, Queensland University, Wildlife Conservation Society, University of Ottowa	\\
NOAA ESD Coral Bleaching Classifier ~\cite{Ehrenberg2022}	&	Courtney Couch, Jonathan Charendoff, Morgan Winston Pomeroy, Tom Oliver, Jon Ehrenberg	&	National Oceanographic and Atmospheric Administration, Pacific Islands Fisheries Science Center	\\
North Shore Disease Surveys	&	Julianna Renzi, Maddie Cunningham	&	University of California Santa Barbara	\\
Okinawa Coral Reef Conservation2017-18	&	Tomofumi Nagata, Eiji Yamakawa	&	Okinawa Environment Science Center	\\
Okinawa Coral Reef Conservation2019-20	&	Tomofumi Nagata	&	Okinawa Environment Science Center	\\
Penida benthic surveys	&	Pascal Sebastian, Rinaldi Gotama	&	Indo Ocean Project	\\
Pohnpei	&	Hugh Runyan	&	Scripps Institution of Oceanography, University of California San Diego	\\
REEFolution Kenya	&	Ewout G. Knoester, Anniek Vos, Bulisa Masiga, Jowan van Lente, Luc Visser, Mercy Zawadi Katana, Omar F. Yusuf	&	Wageningen University and Research, REEFolution Trust	\\
Saudi Red Sea Seasonal Differences: Offshore-Midshelf sheltered crest	&	Clara Nuber, Matt Tietbohl, Karla Gonzalez	&	King Abdullah University of Science and Technology	\\
Saudi Red Sea Seasonal Differences: Offshore-Midshelf sheltered slope	&	Clara Nuber, Matt Tietbohl, Karla Gonzalez	&	King Abdullah University of Science and Technology	\\
SFS Mangrove DR	&	Max Vierling, Samantha Krausse, Toni Trinh	&	School for Field Studies	\\
Sint Eustatius	&	Myrsini Lymperaki	&	Univeristy of Amsterdam	\\
Southern Line Islands	&	Nicole Pedersen, Samantha Clements	&	Scripps Institution of Oceanography, University of California San Diego	\\
STINAPA GCRMN	&	Caren Eckrich, Tessa Haanskorf, Angelica Verschragen	&	STIchting NAtionale PArken Bonaire, Wageningen University and Research, University of Amsterdam	\\
TIBS Directed Research Mangroves	&	Emma Greenberg, Jenna Shea, Rachel Schneider	&	School for Field Studies	\\
Tonga	&	Patrick Smallhorn-West, Lucy Southworth	&	James Cook University	\\
Tonga\_\allowbreak{}2022-08 \& Samoa\_\allowbreak{}2022-09 \& Samoa\_\allowbreak{}2019-12 \& Samoa\_\allowbreak{}2017-12\cite{NOAA_NCRMP_SAMOA_2018} & Chris Sullivan, Katie Lubarsky, Gloria Mariño-Briceño, Hannah Gower, Kylie Yogi, Phi Lang & Scripps Institution of Oceanography, University of California San Diego \\

USSGuardianWreck	&	Catherine Kim, Ben Neal, Dominic Bryant	&	Univeristy of Queensland	\\
Valentina Rocha	&	Valentina Rocha, Emily Esplandiu	&	University of Miami	\\
Vamizi	&	Marques da Silva Isabel, Erwan Sol, Felix Domadoma	&	Center for Research and Environmental Conservation - Lurio University	\\
WAPA Coral Inventory 2.0	&	David Burdick, Colin Lock, Melissa Vaccarino	&	National Park Service	\\
WAPA Interns	&	Ashton Williams, Marisa Agarwal, Andrew O'Connor, Christina Kilkeary, Emma Vaughn, Erin Mullins, Katherine Tangney, Malvika Shrimali, Michelle Diminuco, Motusaga Vaeoso, Natalie Scott, Nicholas Burgos, Philippe Astier, Ryan Stanley, Sarah Yokota, Serena Butler, Ashley Swafford, Xavier Quinata	&	National Park Service	\\
WAPA RFI	&	Anneke Padmos, Julia Padilla, Ronja Steinbach, Tim Clark, Terence Dela Cruz, Eliza Frances Manglona	&	National Park Service	\\
WAPA RFM	&	Anneke Padmos, Julia Padilla, Ronja Steinbach, Tim Clark, Terence Dela Cruz	&	National Park Service	\\
WSU West Hawaii	&	Brian Tissot, Molly Bogeberg	&	Washington State University	\\
ZONE 3 new	&	Jamila Hassan, Sulemani Mohamed	&	Wildlife Conservation Society	\\
\end{longtable}
\end{footnotesize}

\clearpage
\section*{NeurIPS Paper Checklist}

\begin{enumerate}

\item {\bf Claims}
    \item[] Question: Do the main claims made in the abstract and introduction accurately reflect the paper's contributions and scope?
    \item[] Answer: \answerYes{}
    \item[] Justification: The abstract and introduction describe the dataset construction, benchmark settings, and empirical findings that are substantiated by the results and supplementary material. See the abstract, Secs.~1--4, and Supplementary Sec.~3.

\item {\bf Limitations}
    \item[] Question: Does the paper discuss the limitations of the work performed by the authors?
    \item[] Answer: \answerYes{}
    \item[] Justification: The paper discusses geographic imbalance, audit uncertainty, point-based supervision, experimental variability, hosted models, incomplete release artifacts, and potential ecological impact. See Appendix~\ref{sup_limitations}.

\item {\bf Theory assumptions and proofs}
    \item[] Question: For each theoretical result, does the paper provide the full set of assumptions and a complete (and correct) proof?
    \item[] Answer: \answerNA{}
    \item[] Justification: The paper does not present theoretical results, theorems, or proofs; it is a dataset and benchmark paper.

	    \item {\bf Experimental result reproducibility}
	    \item[] Question: Does the paper fully disclose all the information needed to reproduce the main experimental results of the paper to the extent that it affects the main claims and/or conclusions of the paper (regardless of whether the code and data are provided or not)?
	    \item[] Answer: \answerNo{}
	    \item[] Justification: The paper documents the splits, macro-recall protocol, principal training settings, and added results, but exact checkpoint/API identifiers, complete request settings, and executable configurations remain incomplete. See Appendices~\ref{sec:exp_detail} and~\ref{sup_resources}.

\item {\bf Open access to data and code}
    \item[] Question: Does the paper provide open access to the data and code, with sufficient instructions to faithfully reproduce the main experimental results, as described in supplemental material?
    \item[] Answer: \answerNo{}
    \item[] Justification: Annotations, metadata, and Al-Wajh imagery are public. The inspected benchmark repository documents data access but does not yet provide the complete executable workflow, and a versioned description artifact is absent from the dataset listing. See Appendix~\ref{sup_resources}.

\item {\bf Experimental setting/details}
    \item[] Question: Does the paper specify all the training and test details (e.g., data splits, hyperparameters, how they were chosen, type of optimizer) necessary to understand the results?
    \item[] Answer: \answerYes{}
    \item[] Justification: The paper reports split construction, within-source and cross-source evaluation, shot regimes, optimizers, learning rates, epochs, batch sizes, prompting, and model-selection protocol. See Sec.~3 and Supplementary Sec.~3.

\item {\bf Experiment statistical significance}
    \item[] Question: Does the paper report error bars suitably and correctly defined or other appropriate information about the statistical significance of the experiments?
    \item[] Answer: \answerNo{}
    \item[] Justification: Three runs each for full-data ViT-B/16 and BioCLIP2, plus a separate class-capped Qwen3-VL study, are reported with individual results and sample standard deviations (Appendix~\ref{sup_training_analysis}). Most other scores and the augmentation ablation remain single-run estimates; comprehensive error bars and significance tests are not available.

\item {\bf Experiments compute resources}
    \item[] Question: For each experiment, does the paper provide sufficient information on the computer resources (type of compute workers, memory, time of execution) needed to reproduce the experiments?
    \item[] Answer: \answerNo{}
    \item[] Justification: Principal optimization settings are reported, but consistent per-run accelerator counts, effective batches, trainable parameters, wall times, and GPU-hours remain incomplete. Model-family protocols are explicitly not compute-matched. See Appendix~\ref{sup_training_analysis}.

\item {\bf Code of ethics}
    \item[] Question: Does the research conducted in the paper conform, in every respect, with the NeurIPS Code of Ethics \url{https://neurips.cc/public/EthicsGuidelines}?
    \item[] Answer: \answerYes{}
    \item[] Justification: To the best of our knowledge, the work conforms to the NeurIPS Code of Ethics. The paper uses publicly available coral reef imagery and curated expert annotations for scientific benchmarking and conservation-oriented research.

\item {\bf Broader impacts}
    \item[] Question: Does the paper discuss both potential positive societal impacts and negative societal impacts of the work performed?
    \item[] Answer: \answerYes{}
    \item[] Justification: Appendix~\ref{sup_limitations} discusses potential monitoring benefits and how errors on rare taxa or underrepresented regions could bias ecological estimates and conservation decisions.

\item {\bf Safeguards}
    \item[] Question: Does the paper describe safeguards that have been put in place for responsible release of data or models that have a high risk for misuse (e.g., pre-trained language models, image generators, or scraped datasets)?
    \item[] Answer: \answerNA{}
    \item[] Justification: The released assets are coral reef imagery, taxonomy mappings, and classification models for scientific benchmarking, which do not present the high-risk misuse profile contemplated by this checklist item.

\item {\bf Licenses for existing assets}
    \item[] Question: Are the creators or original owners of assets (e.g., code, data, models), used in the paper, properly credited and are the license and terms of use explicitly mentioned and properly respected?
    \item[] Answer: \answerNo{}
    \item[] Justification: The paper credits the original datasets, models, and platforms through citations, but it does not yet enumerate the licenses and terms of use for all external assets in the manuscript. This should be added to the release documentation and, if needed, to the supplement.

\item {\bf New assets}
    \item[] Question: Are new assets introduced in the paper well documented and is the documentation provided alongside the assets?
    \item[] Answer: \answerYes{}
    \item[] Justification: The paper and supplement document the dataset construction pipeline, taxonomy mapping, benchmark splits, verification procedure, and release plan for the new annotations, metadata, models, and Al-Wajh images. See Secs.~2--4 and Supplementary Secs.~2, 3, and 9.

\item {\bf Crowdsourcing and research with human subjects}
    \item[] Question: For crowdsourcing experiments and research with human subjects, does the paper include the full text of instructions given to participants and screenshots, if applicable, as well as details about compensation (if any)?
    \item[] Answer: \answerNA{}
    \item[] Justification: The paper does not report crowdsourcing experiments or research studies involving human subjects as participants.

\item {\bf Institutional review board (IRB) approvals or equivalent for research with human subjects}
    \item[] Question: Does the paper describe potential risks incurred by study participants, whether such risks were disclosed to the subjects, and whether Institutional Review Board (IRB) approvals (or an equivalent approval/review based on the requirements of your country or institution) were obtained?
    \item[] Answer: \answerNA{}
    \item[] Justification: The paper does not involve human-subjects research of the kind covered by this checklist item.

\item {\bf Declaration of LLM usage}
    \item[] Question: Does the paper describe the usage of LLMs if it is an important, original, or non-standard component of the core methods in this research? Note that if the LLM is used only for writing, editing, or formatting purposes and does \emph{not} impact the core methodology, scientific rigor, or originality of the research, declaration is not required.
    \item[] Answer: \answerYes{}
    \item[] Justification: Multimodal large language models are a core part of the benchmark evaluation, and the paper describes how they are prompted and fine-tuned. See Sec.~4 and Supplementary Sec.~3.

\end{enumerate}

\end{document}